\documentclass{article}
\usepackage[nonatbib, preprint]{neurips_2021}
\usepackage{amsfonts}
\usepackage[numbers]{natbib}
\usepackage{amsmath}
\usepackage{amssymb}
\usepackage[normalem]{ulem}
\usepackage{amsthm}
\usepackage{bbm}
\usepackage{adjustbox}
\usepackage{xcolor}
\usepackage{microtype}
\usepackage{graphicx}
\usepackage{booktabs} 
\usepackage{algorithm}
\usepackage{hyperref}
\usepackage{algpseudocode}
\usepackage{mathtools}
\mathtoolsset{showonlyrefs}
\usepackage{wrapfig}
\usepackage[subtle]{savetrees}
\usepackage{subcaption}
\usepackage{multirow}
\usepackage{placeins}
\newcommand{\colorred}[1]{{\color{black} #1}}

\makeatletter
\def\BState{\State\hskip-\ALG@thistlm}
\makeatother

\theoremstyle{definition}
\newtheorem{proposition}{Proposition}

\newtheorem{theorem}{Theorem}

\newtheorem{assumption}{Assumption}

\newcommand{\tSc}{\tilde{\Sc}}
\newcommand{\tAc}{\tilde{\Ac}}

\DeclareRobustCommand{\Rb}{\mathbb{R}}

\DeclareRobustCommand{\Ac}{\mathcal{A}}
\DeclareRobustCommand{\Dc}{\mathcal{D}}
\DeclareRobustCommand{\Sc}{\mathcal{S}}
\DeclareRobustCommand{\Xc}{\mathcal{X}}

\DeclareRobustCommand{\Rb}{\mathbb R}

\DeclareRobustCommand{\norm}[1]{\left\lVert #1 \right\rVert}

\newcommand{\distas}[1]{\mathbin{\overset{#1}{\kern\z@\sim}}}%
\newsavebox{\mybox}\newsavebox{\mysim}

\newcommand{\distras}[1]{%
  \savebox{\mybox}{\hbox{\kern3pt$\scriptstyle#1$\kern3pt}}%
  \savebox{\mysim}{\hbox{$\sim$}}%
  \mathbin{\overset{#1}{\kern\z@\resizebox{\wd\mybox}{\ht\mysim}{$\sim$}}}%
}

\newcommand{\Ex}{\mathbb{E}}

\title{PerSim: Data-efficient Offline Reinforcement\\ Learning  with Heterogeneous Agents\\ via Personalized Simulators}

\author{%
  Anish Agarwal\\
  MIT\\
  \texttt{anish90@mit.edu}\\
  \And
  Abdullah Alomar \\ 
  MIT\\
  \texttt{aalomar@mit.edu}\\
\And
  Varkey Alumootil \\
    MIT\\
  \texttt{varkey@mit.edu}\\
  \And
  Devavrat Shah \\ 
    MIT\\
  \texttt{devavrat@mit.edu}\\
  \And
  Dennis Shen \\   
    MIT\\
  \texttt{deshen@mit.edu}\\
  \And
  Zhi Xu \\   
  MIT\\
  \texttt{zhixu@mit.edu}\\
  \And
  Cindy Yang \\
  MIT \\
  \texttt{cxy99@mit.edu}
}

\begin{document}

\maketitle

\begin{abstract}
We consider offline reinforcement learning (RL) with heterogeneous agents under severe data scarcity, i.e., we only observe a single historical trajectory for every agent under an unknown, potentially sub-optimal policy. 
We find that the performance of state-of-the-art offline and model-based RL methods degrade significantly given such limited data availability, even for commonly perceived ``solved” benchmark settings such as ``MountainCar'' and ``CartPole''.
To address this challenge, we propose PerSim, a model-based offline RL approach which first learns a personalized simulator for each agent by collectively using the historical trajectories across all agents, prior to learning a policy. 
We do so by positing that the transition dynamics across agents can be represented as a latent function of latent factors associated with agents, states, and actions; subsequently, we theoretically establish that this function is well-approximated by a   ``low-rank''  decomposition of separable agent, state, and action latent functions.
This representation suggests a simple, regularized neural network architecture to effectively learn the transition dynamics per agent, even with scarce, offline data.
We perform extensive experiments across several benchmark environments and RL methods. 
The consistent improvement of our approach, measured in terms of both state dynamics prediction and eventual reward, confirms the efficacy of our framework in leveraging limited historical data to simultaneously learn personalized policies across agents. 

\end{abstract}

\section{Introduction}\label{sec:introduction}
Reinforcement learning (RL) coupled with expressive deep neural networks has now become a generic yet powerful solution for learning complex decision-making policies for an agent of interest; it provides the key algorithmic foundation underpinning recent successes such as game solving~\citep{mnih2015human,silver2017mastering,silver2017masteringchess} and robotics~\citep{levine2016end,kalashnikov2018qt}. 
However, many state-of-the-art RL methods are data hungry and require the ability to query samples at will, which is infeasible for numerous settings such as healthcare, autonomous driving, and socio-economic systems.
As a result, there has been a rapidly growing literature on ``offline RL''~\citep{levine2020offline,kumar2019stabilizing,liu2020provably,fujimoto2019off}, which focuses on leveraging existing datasets to learn decision-making policies. 

\noindent Within offline RL, we consider a regime of {\em severe data scarcity}: 
there are multiple agents and for each agent, we only observe a single historical trajectory generated under an unknown, potentially sub-optimal policy; 
further, the agents are {\em heterogeneous}, i.e., each agent has unique state transition dynamics. 
Importantly, the characteristics of the agents that make their transition dynamics heterogeneous are {\em latent}. 
Using such limited offline data to {\em simultaneously} learn a good ``personalized'' policy for each agent is a challenging setting that has received limited attention.
Below, we use a prevalent example from healthcare to motivate and argue that tackling such a challenge is an important and necessary step towards building personalized decision-making engines via RL in a variety of important applications. 

\begin{figure*}[t!]
\begin{subfigure}[t]{0.33\textwidth}
        \centering
    \includegraphics[width = 0.9\textwidth]{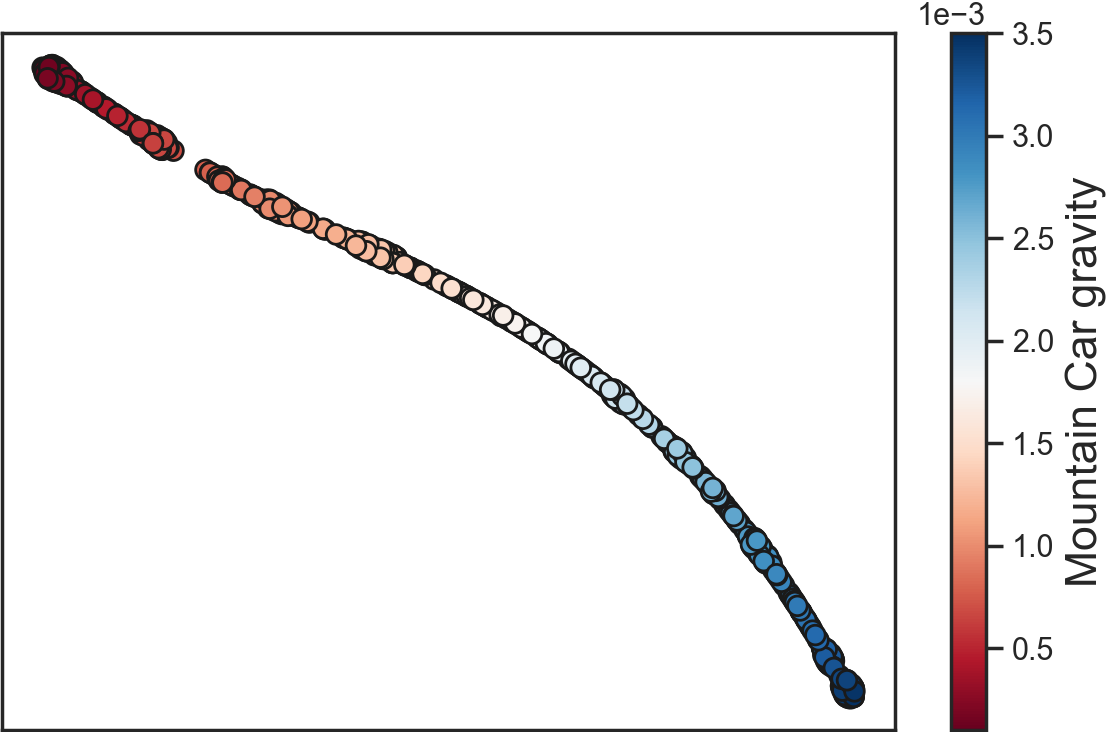}
    \caption{MountainCar}
\end{subfigure}
\begin{subfigure}[t]{0.33\textwidth}
        \centering
    \includegraphics[width = 0.9\textwidth]{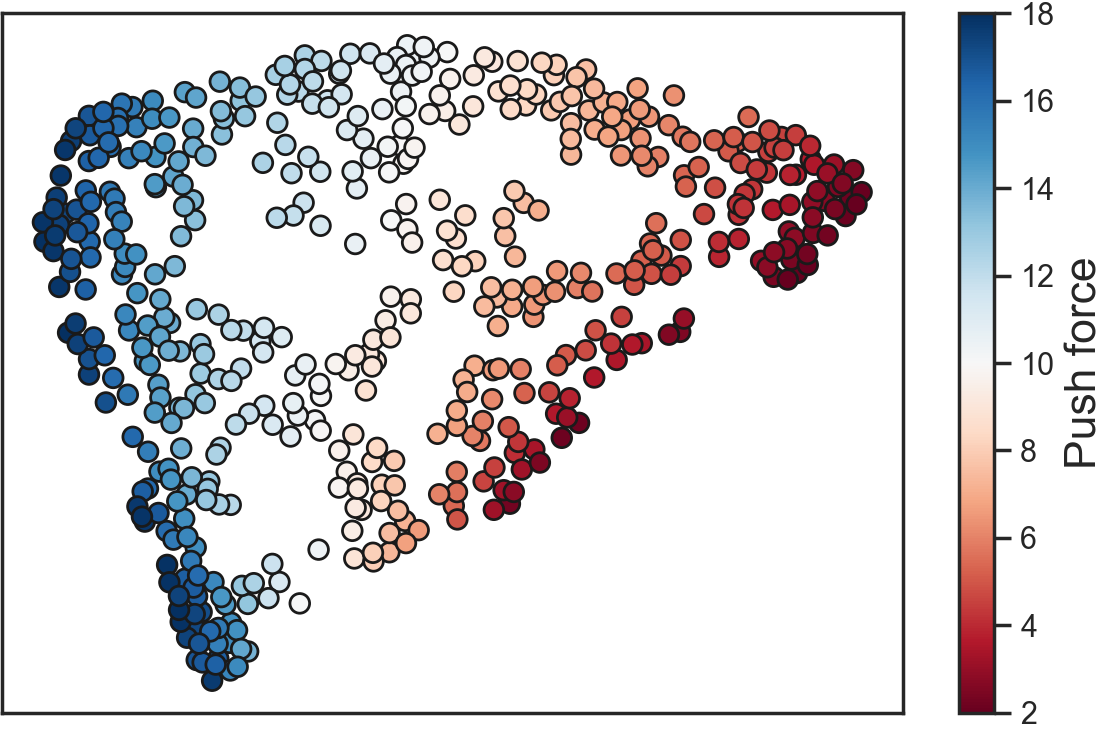}
    \caption{CartPole}
\end{subfigure}
\begin{subfigure}[t]{0.33\textwidth}
        \centering
    \includegraphics[width = 0.9\textwidth]{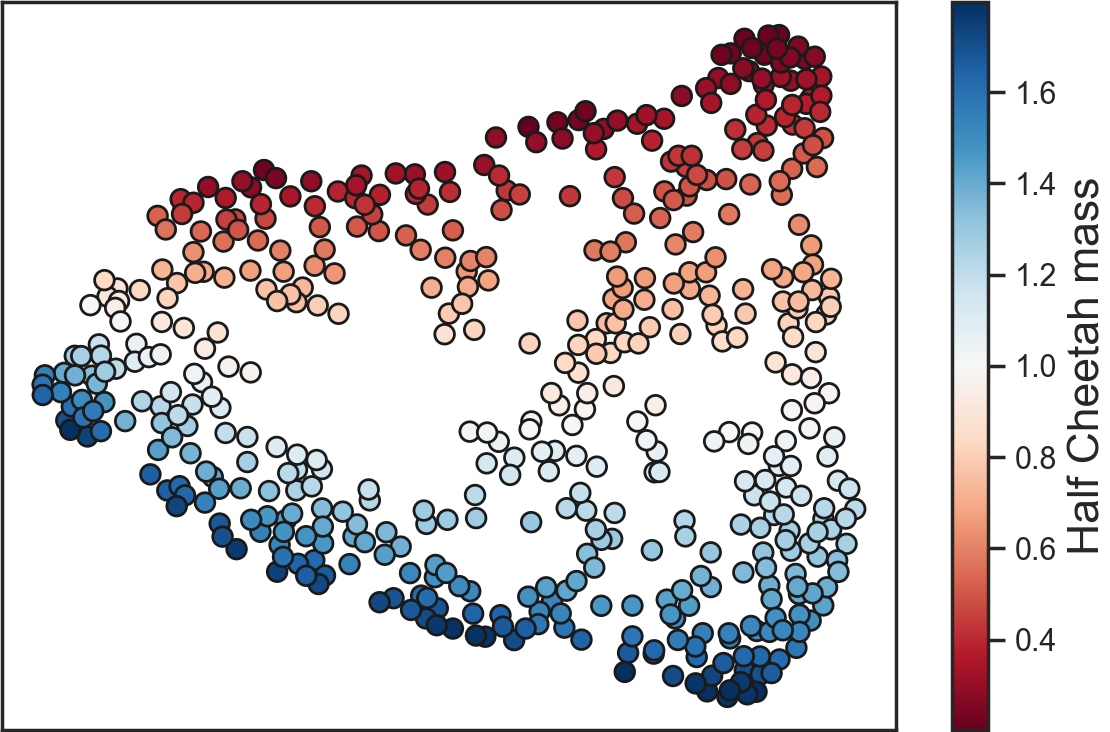}
    \caption{HalfCheetah}
\end{subfigure}
\vspace{-3mm}
\caption{\footnotesize{
t-SNE visualization of the learned latent factors for the 500 heterogeneous agents. Colors indicate the value of the modified parameters in each environment (e.g., gravity in MountainCar).
There is an informative low-dimensional manifold induced by the agent-specific latent factors, and there is a natural direction on the manifold along which the parameters that characterize the heterogeneity vary continuously and smoothly.
}}
\label{fig:LF_main_paper}
\vspace{-4mm}
\end{figure*}

\noindent {\em A Motivating Example.}
Consider a pre-existing clinical dataset of patients (agents). 
Our goal is to design a personalized treatment plan (policy) for each patient moving forward. 
Notable challenges include the following: 
First, each patient only provides a single trajectory of their medical history.
Second, each patient is heterogeneous in that they may have a varied response for a given treatment under similar medical conditions; 
further, the underlying reason for this heterogeneous response across patients is likely unknown and thus not recorded in the dataset.
Third, in the absence of an accurate personalized ``forecasting'' model for a patient's medical outcome given a treatment plan, the treatment assigned is likely to be sub-optimal. 
This is particularly true for complicated medical conditions like T-Cell Lymphoma.
We aim to address the challenges laid out above (offline scarce data, heterogeneity, and sub-optimal policies) so as to develop a personalized ``forecasting'' model for each patient  under any given treatment plan. 
Doing so will then naturally enable ideal personalized treatment policies for each patient.


\noindent  {\em Key question.}
Tackling a scenario like the one described above in a principled manner is the focus of this work. 
In particular, we seek to answer the following question: 
%
\begin{center}
    {\em``Can we leverage scarce, offline data collected across heterogeneous agents under 
    unknown, sub-optimal policies to learn a personalized  policy for each agent?''} 
\end{center}
%

\noindent 
{\bf Our Contributions.} 
As our main contribution, we answer this question in the affirmative by developing a structured framework to tackle this challenging yet meaningful setting. 
Next, we summarize the main methodological, theoretical, algorithmic, and experimental contributions in our proposed framework. 

\noindent 
{\em Methodological---personalized simulators.}
We propose a novel methodological framework, PerSim, to learn a policy given the data availability described above. 
Taking inspiration from the model-based RL literature, our approach is to first build a personalized simulator for each agent. 
Specifically, we use the offline data collectively available across heterogeneous agents to learn the unique transition dynamics for each agent. 
We do this \emph{without} requiring access to the covariates or features that drive the heterogeneity amongst the agents.
Having constructed a personalized simulator, we then learn a personalized decision-making policy separately for each agent by simply using online model predictive control (MPC)~\citep{garcia1989model,camacho2013model}.

%
\noindent 
{\em Theoretical---learning across agents.} 
As alluded to earlier, the challenge in building a personalized simulator for each agent is that we only have access to a single offline trajectory for any given agent. 
Hence, each agent likely explores a very small subset of the entire state-action space. 
However, by viewing the trajectories across the multitude of agents collectively, we potentially have access to a relatively larger and more diverse offline dataset that covers a much richer subset of the state-action space.
Still, any approach that augments the data of an agent in this manner must address the possibly large heterogeneity amongst the agents, which is challenging as we do not observe the characteristics that make agents heterogeneous.
Inspired by the literature on collaborative filtering for recommendation systems, we posit that the transition dynamics across agents can be represented as a latent function of latent factors associated with agents, states, and actions. 
In doing so, we establish
that this function is well-approximated by a ``low-rank''  decomposition of separable agent, state, and action latent functions (Theorem \ref{thm:representation}).
Hence, for any finite sampling of the state and action spaces, accurate model learning for each agent with offline data---generated from any policy---can be reduced to estimating a low-rank tensor corresponding to agents, states, and actions.
As such, low-rank tensors can provide a useful algorithmic lens to enable model learning with offline data in RL and we hope this work leads to further research studying the relationship between these seemingly disparate fields.
%
%

\noindent 
{\em Algorithmic---regularizing via a latent factor model.} 
As a consequence of our low-rank representation result, we propose a natural neural network architecture that respects the constraints induced by the factorization across agents, states, and actions (Section \ref{sec:algorithm}). 
It is this principled structure, which accounts for agent heterogeneity and regularizes the model learning, that ensures the success of our approach despite access to only scarce and heterogeneous data. 
Further, we propose a natural extension of our framework which generalizes to unseen agents, i.e., agents that are not observed in the offline data. 

\begin{wrapfigure}{r}{0.6\linewidth}
\centering
\includegraphics[width =0.6\textwidth]{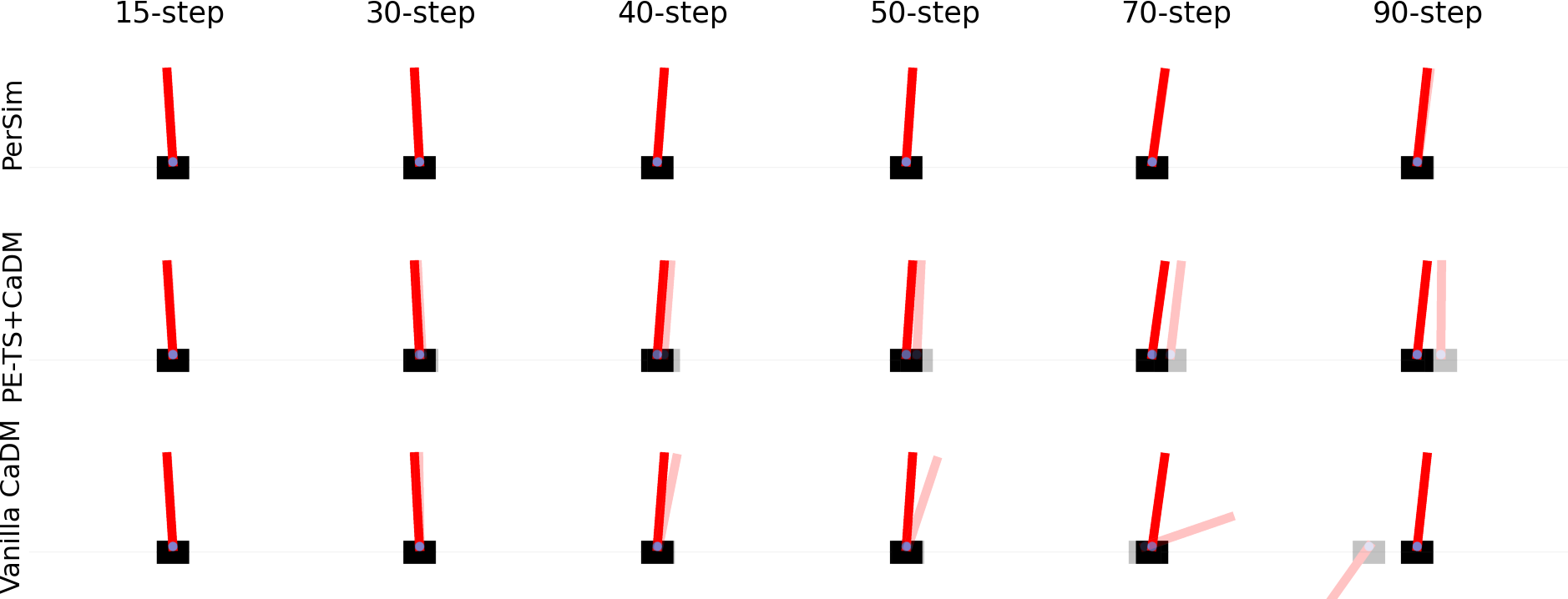}
\caption{\footnotesize{Visualization of prediction accuracy of the various learned models for CartPole. Actual and predicted states are denoted by the opaque and translucent objects, respectively.}}
\label{fig:visulization_main_paper}
\vspace{-2mm}
\end{wrapfigure}

\colorred{
\noindent 
{\em Experimental---extensive benchmarking.}
Using standard environments from OpenAI Gym, 
we extensively benchmark PerSim against four state-of-the-art methods: 
a model-based online RL method (CaDM) \cite{lee2020context}, two model-free offline RL method (BCQ and CQL) \cite{fujimoto2019off, kumar2020conservative}, and a model-based offline RL method (MOReL)~\cite{kidambi2020morel}. 
%
%
Below, we highlight six conclusions we reach from our experiments. 
(i) Despite access to only a single trajectory from each agent (and no access to the covariates that drive agent heterogeneity), PerSim  produces accurate personalized simulators for each agent.
(ii) All benchmarking algorithms perform sub-optimally for the data availability we consider, even on simple baseline environments such as MountainCar and Cartpole, which are traditionally considered to be ``solved''.
(iii) PerSim is able to robustly extrapolate outside the policy used to generate the offline dataset, even if the policy is highly sub-optimal (e.g., actions are sampled uniformly at random). 
(iv) To corroborate our latent factor representation, we find that across all environments, the learned agent-specific latent factors correspond very closely with the latent source of heterogeneity amongst agents; we re-emphasize this is despite PerSim not getting access to the agent covariates.
(v) We find that augmenting the training data of an offline model-free method (e.g., BCQ) with PerSim-generated synthetic trajectories results in a significantly better average reward. }
(vi) As an ablation study, if we decrease the number of observed trajectories, PerSim consistently achieves a higher reward than the other baselines across most agents, indicating its robustness to data scarcity. 
For a visual depiction of the conclusions, see Figure~\ref{fig:LF_main_paper} for  the learned latent agent factors and Figure~\ref{fig:visulization_main_paper} for the relative prediction accuracy of the learned model using PerSim versus ~\cite{lee2020context}. 

\vspace{1mm}

\noindent\textbf{Related Work. }
Due to space constraints, we present a short overview of the related literature.
A more detailed review is provided in Appendix \ref{appendix:related_work}. 

\noindent 
There are two sub-fields within RL that are of particular relevance: (i) model-based online RL and (ii) model-free offline RL.
(i) In the model-based RL literature, the transition dynamics (simulator) is learnt and subsequently utilized for policy learning. 
%
These methods have been found to have far better data efficiency compared to their model-free counterparts~\citep{wang2019benchmarking,chua2018deep,clavera2018model,kurutach2018model,kaiser2019model,hafner2019learning}.
However, the current model-based literature mostly focuses on the setting where one can adaptively sample trajectories in an online manner during model-learning.
A few recent works~\citep{lee2020context,nagabandi2018learning} have considered agent heterogeneity.
We compare with \cite{lee2020context} given its strong performance in handling heterogeneous agents.
(ii) In the model-free offline RL literature, one uses a pre-recorded dataset to directly learn a policy, i.e., without first learning a model.
Thus far, the vast majority of offline RL methods are model-free and designed for settings that allow access to numerous trajectories from a single agent, i.e., no agent heterogeneity~\cite{fujimoto2019off,kumar2019stabilizing,laroche2019safe,liu2020provably,wu2019behavior,agarwal2020optimistic,kumar2020conservative}.
\colorred{
To study how much offline methods suffer if agent heterogeneity is introduced, we compare with both \cite{fujimoto2019off} and \cite{kumar2020conservative} given their strong performance with offline data. 
We choose these baselines as they come from a well-established literature but their abilities to simultaneously handle agent heterogeneity and sparse offline data has yet to be studied.}

\colorred{
\noindent 
Model-based offline RL is still a relatively nascent field.
Two recent works~\citep{kidambi2020morel,yu2020mopo} have shown that, in certain settings, model-based offline methods can outperform their model-free counterparts on benchmark environments.
In \cite{yu2020mopo}, authors exhibit this using existing online model-based methods with minimal changes.
However, both works restrict attention to the setting where there is only one agent (or environment) and a large number of observations from that agent are available. 
Hence, we study how these methods perform when given only sparse data from heterogeneous agents, and we find their performance does suffer.
%
Extending these model-based offline RL methods to work with sparse data from heterogeneous agents, possibly by building upon the latent low-rank {functional} representation we propose, remains an interesting future work. 
}

\vspace{-2mm}
\section{Problem Statement}\label{sec:problem_statement}
\vspace{-2mm}
We consider the standard RL framework with $N$ heterogeneous agents. 
%
We index agents with $n \in [N]$\footnote{For any positive integer $N$, let $[N] = \{1,\dots, N\}$.}.
Formally, we describe our problem as a Markov Decision Process (MDP) defined by the tuple $(\Sc, \Ac, P_n, R_n, \gamma_n, \mu_n)$. 
Here, $\Sc$ and $\Ac$ denote the state and action spaces, respectively, which are common across agents. 
For every agent $n$, $P_n(s^\prime | s, a)$ is the unknown transition kernel, $R_n(s,a)$ is the immediate reward received, $\gamma_n \in (0,1)$ is the discounting factor, and $\mu_n$ is the initial state distribution. 

\noindent {\bf Observations.}
We consider an offline RL setting where we observe a single trajectory of length $T$ for each of the $N$ heterogeneous agents. 
Formally, for each agent $n$ and time step $t$, let $s_n(t) \in \Sc$, $a_n(t) \in \Ac$, and $r_n(t) \in \Rb$ denote the observed state, action, and reward. 
We denote our observations as $\Dc = \{(s_n(t), a_n(t), r_n(t)): n \in [N], t \in [T]\}$. 


\noindent {\bf Goals.} We state our two primary goals below. 
%

%
\noindent {\em ``Personalized'' trajectory prediction. }
For a given agent $n$ and state-action pair $(s, a) \in \Sc \times \Ac$, we would like to estimate 
$
 \Ex[s'_n | (s_n, a_n) = (s, a)],
$
i.e., given the observations $\Dc$, we would like to build a ``personalized'' simulator (i.e., a model of the transition dynamic) for each agent $n$. 
%

%
\noindent {\em ``Personalized'' model-based policy learning. }
To test the efficacy of the personalized simulator, we would like to subsequently use it to learn a good decision-making policy for agent $n$, denoted as $\pi_n: \Sc \to \Ac$, which takes as input a given state and produces a corresponding action.

\noindent
\colorred{
{\em A causal inference lens.} 
We note that this problem can be thought of as a counterfactual prediction problem.
Our goal is to build a personalized simulator that answers the following question:
``what would have happened had the agent took a sequence of actions other than the one we observe in the dataset?''.
Broadly speaking, our aim in PerSim is to answer such questions by observing how other agents behave under varied sequences of actions.
More generally, we hope PerSim serves to further link causal inference with offline RL, which is fundamentally regarded as a counterfactual inference problem ~\cite{levine2020offline}.
}

\vspace{-3mm}

\section{Latent Low-rank Factor Representation}\label{sec:low_rank_representation}
%

To address the goals, we introduce a latent factor model for the transition dynamics. 
Leveraging latent factors have been successful in recommendation systems for overcoming heterogeneity of users. 
Such models have also been shown to provide a ``universal'' representation for multi-dimensional exchangeable arrays \cite{aldous, hoover}.  
Indeed, our latent model holds for known environments such as MountainCar.

\noindent 
Assume $\Sc \subseteq \Rb^D$, i.e., the state is $D$-dimensional.
Let $s_{nd}$ refer to the $d$-th coordinate of $s_{n}$.
We posit the transition dynamics (in expectation) obey the following model: for every agent $n$ and state-action pair $(s,a)$,
\begin{align} \label{eq:model} 
    \Ex[s_{nd}' | (s_{n}, a_{n}) = (s, a)] = f_d(\theta_n, \rho_s, \omega_a),
\end{align}
where
$s_{nd}'$ denotes the $d$-th state coordinate after taking action $a$.
Here, $\theta_n \in \Rb^{d_1}$, $\rho_s \in \Rb^{d_2}$, $\omega_a \in \Rb^{d_3}$ for some $d_1, d_2, d_3 \ge 1$ are latent feature vectors capturing relevant information specific to the agent, state, and action; 
$f_d: \Rb^{d_1} \times \Rb^{d_2} \times \Rb^{d_3} \rightarrow \Rb$ is a latent function capturing the model relationship between these latent feature vectors. 
%
We assume $f_d$ is $L$-Lipschitz and the latent features are bounded.
\begin{assumption}\label{assumption:f_properties}
Suppose $\theta_n \in [0,1]^{d_1}, \rho_s \in [0,1]^{d_2}, \omega_a \in [0,1]^{d_3}$, and $f_d$ is $L$-Lipschitz with respect to its arguments, i.e., $| f_d(\theta_{n'}, \rho_{s'}, \omega_{a'}) - f_d(\theta_n, \rho_s, \omega_a) | \le L (\| \theta_{n'} - \theta_n \|_2 + \|\rho_{s'} - \rho_s \|_2 + \| \omega_{a'} - \omega_a \|_2$). 
\end{assumption}
\vspace{-2mm}
\noindent 
For notational convenience, let $\tilde{f}_d:[N] \times \Sc \times \Ac \rightarrow \Rb$ be such that $\tilde{f}_d(n, s, a) = \Ex[s^\prime_{nd} | (s_{n}, a_{n}) = (s, a)]$.
\begin{theorem}\label{thm:representation}
Suppose Assumption \ref{assumption:f_properties} holds and without loss of generality, let $d_1, d_3 \le d_2$.
Then for all $d \in [D]$ and any $\delta > 0$, there exists $h_d:[N] \times \Sc \times \Ac \rightarrow \Rb$, such that 
$h_d(n, s, a) = \sum^r_{\ell=1} u_{\ell}(n) v_{\ell}(s, d) w_{\ell}(a)$ with $r \le C \delta^{-(d_1 + d_3)}$ and $\|h_d - \tilde{f}_d \|_\infty \le 2L\delta$, where $C$ is an absolute constant.
\end{theorem} 

\noindent 
Theorem~\ref{thm:representation} suggests that under the model in \eqref{eq:model}, the transition dynamics are well approximated by a low-rank order-three functional tensor representation. 
In fact, as we show below, for a classical non-linear dynamical system, it is 
exact. 

\noindent 
{\em An example.} 
%
We show that the MountainCar transition dynamics~\cite{brockman2016openai} exactly satisfies this low-rank tensor representation.
In MountainCar, the state $s_n = [s_{n1}, s_{n2}]$ consists of car (agent) $n$'s position and velocity, i.e., $\Sc \subseteq \Rb^2$; 
the action $a_n$ is a scalar that represents the applied acceleration, i.e., $\Ac \subseteq \Rb$. 
For car $n$, parameterized by gravity $g_n$, the (deterministic) transition dynamics given action $a_n$ are
\begin{align*}
s^\prime_{n1} = 
\  s_{n1} +s_{n2}   -\frac{g_n\cos(3s_{n1})}{2} +\frac{a_n}{2}, \quad 
 s^\prime_{n2} = s_{n2}  - g_n\cos(3s_{n1}) + a_n.
\end{align*}
\begin{proposition}\label{proposition:example}
%
%
%
In MountainCar, $r = 3$.
\end{proposition}


{
\noindent{\bf Model Learning \& Tensor Estimation.} 
}
Consider any finite sampling of the states $\tSc \subset \Sc$ and actions $\tAc \subset \Ac$. 
%
Let $\Xc = [X_{nsad}] \in \Rb^{N \times |\tSc| \times |\tAc| \times D}$ be the order-four tensor, where $X_{nsad} = f_d(\theta_n, \rho_s, \omega_a)$. 
%
Hence, to learn the model of transition dynamics for all the agents over
$\tSc$, $\tAc$, it is sufficient to estimate the tensor, $\Xc$, from observed data.
The offline data collected for a given policy induces a corresponding observation pattern of this tensor. 
Whether the complete tensor is recoverable, is determined by this induced sparsity pattern and the rank of $\Xc$. 
Notably, Theorem \ref{thm:representation} suggests that $\Xc$ is low-rank under mild regularity conditions.
Therefore, the question of model identification, i.e., completing the tensor $\Xc$, boils down to conditions on the offline data in terms of the observation pattern that it induces in the tensor. 
In the existing tensor estimation literature, there are few sparsity patterns for which the underlying tensor can be provably recovered, provided $\Xc$ has a low-rank structure.
They include: 
(i) each entry of the tensor is observed independently at random with sufficiently high-probability~\cite{tensor1, tensor2, tensor3};
(ii) the entries that are observed are {\em block-structured}~\cite{SI, shah2020sample}.
However, the most general set of conditions on the sparsity pattern under which the underlying tensor can be faithfully estimated, i.e., the model is identified for our setup, remains an important and active area of research. 

\vspace{-2mm}
\section{PerSim Algorithm}  \label{sec:algorithm}
%

\begin{wrapfigure}{r}{0.4\textwidth}
        \centering
    \includegraphics[width = 0.4\textwidth]{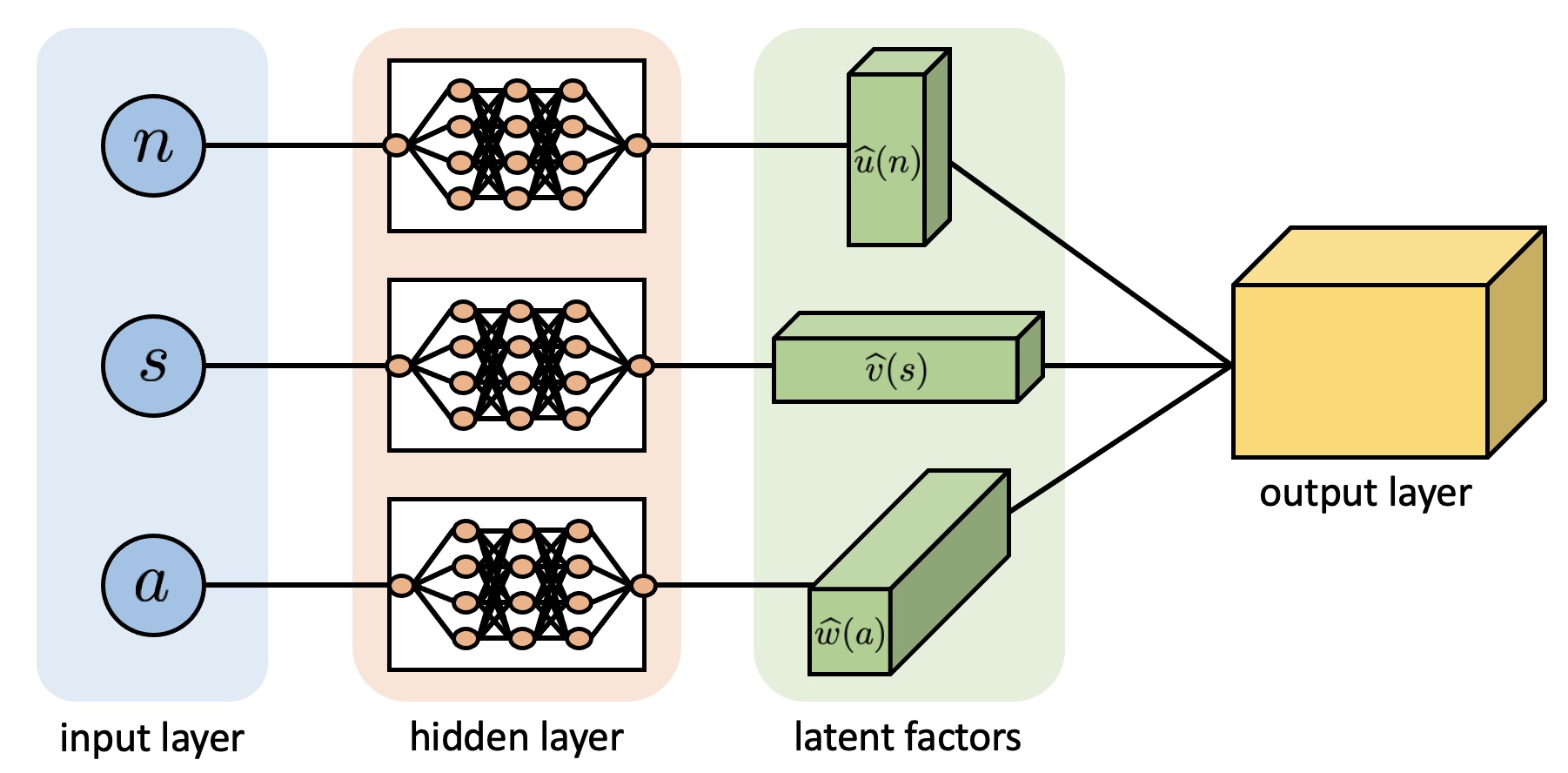}
    \caption{\footnotesize{Neural Network Architecture}}
    \vspace{-3mm}
    \label{fig:nn}
\end{wrapfigure}

We now detail our proposed algorithm which is composed of two steps: (i) build a personalized simulator for each agent $n$ given offline observations $\Dc$, which is comprised of a single trajectory per agent; (ii) learn a personalized decision-making policy using MPC.
\noindent \textbf{Step 1. Learning Personalized Simulators. }
Theorem \ref{thm:representation} suggests that the transition dynamics can be represented as a low-rank tensor function with 
latent functions associated with the agents, states, and actions. 
This guides the design of a simple, regularized neural network architecture: we use three separate neural networks to learn the agent, state, and action latent functions, i.e., we remove any edges between these three neural networks. 
See Figure \ref{alg:nn} for a visual depiction of our proposed architecture.
Specifically, to estimate the next state for a given agent $n$ and  state-action pair $(s, a) \in \Sc \times \Ac$, we  learn the following functions: 

\begin{enumerate}
    \item An agent encoder $g_u: [N] \to \Rb^{r}$, parameterized by $\psi$, which estimates the latent function associated with an agent, i.e., $\widehat{u}(n) = g_u(n;\psi)$.
    \item A state encoder $g_v: \Sc  \rightarrow \Rb^{D\times r}$ parameterized by $\phi$, which estimates $D$ latent functions, where each vector is associated with the corresponding state coordinate, i.e.,
    $
    \widehat{v}(s) := (\widehat{v}(s, 1), \ \dots, \ \widehat{v}(s, D))^T = g_v(s;\phi).
    $
    \item An action encoder $g_w: \Ac  \rightarrow \Rb^r$   parameterized by $\theta$, which estimates the latent function associated with the action, i.e., $\widehat{w}(a) = g_w(a;\theta)$.
\end{enumerate}
Then, our estimate of the  expected $d$-th coordinate of the next state is given by
$
    \widehat{\Ex}[ s_{nd}' | (s_{n}, a_{n}) = (s, a)]=
    \sum_{\ell=1}^r \widehat{u}_{\ell}(n) \widehat{v}_{\ell}(s,d) \widehat{w}_{\ell}(a). 
$
We optimize our agent, state, and action encoders by minimizing the squared loss:
$
    \mathcal{L}(s,a,n,s';\psi, \phi,\theta ) =  \sum_{d=1}^D \Big(s_{nd}' -\sum_{\ell=1}^r \widehat{u}_{\ell}(n) \widehat{v}_{\ell}(s,d) \widehat{w}_{\ell}(a)\Big)^2. 
$

\noindent \textbf{Step 2. Learning a Decision-making Policy.}
We use MPC to select the next action, as common practice in the literature \cite{lee2020context}. 
Since the offline data may not span the entire state-action space, planning using a learned simulator without any regularization may result in ``model exploitation'' \cite{levine2020offline}.
Thus, when choosing the action via MPC, we choose the first action from the sequence of actions with the best \textit{average} reward across an ensemble of $M$ simulators. 
%
%
\noindent 
In our experiments, we find that the simple technique of using the state difference instead of the raw state improves performance. 
For a detailed description of the algorithm, including the pseudo-code, please refer to Appendix~\ref{appendix:sec:implementation}. 
%


\vspace{-3mm}
\section{Experiments}\label{sec:experiments}
\vspace{-2mm}
In this section, through a systematic collection of experiments on a variety of benchmark environments, we demonstrate that PerSim consistently outperforms state-of-the-art model-based and offline model-free RL algorithms, in terms of prediction error and reward, for the data regime we consider.
\vspace{-2mm}
\subsection{Setup and Benchmarks}\label{sec:setup}
\vspace{-2mm}
We evaluate PerSim on three benchmark environments from OpenAI Gym~\cite{brockman2016openai}: MountainCar, CartPole, and HalfCheetah.
A detailed description of each environment can be found in Appendix~\ref{appendix:sec:setup}.

\noindent {\bf Heterogeneous Agents.}
We introduce agent heterogeneity across the various environments by modifying the covariates that characterize the transition dynamics of an agent. 
This is in line with what has been done in the literature~\cite{lee2020context} to study algorithmic robustness to agent heterogeneity.
{
The range of parameters we consider for each of the three environments is given in Table~\ref{table:env}  of Appendix \ref{appendix:sec:setup}}; 
for example, we create heterogeneous agents in MountainCar by varying the gravity parameter of a car (agent) within the interval $[0.01,0.035]$.
%

%
\noindent 
For each environment, we uniformly sample 500 covariates (i.e., heterogeneous agents) from the parameter ranges displayed in Table \ref{table:env} of Appendix \ref{appendix:sec:setup}.
We then select five of the 500 agents to be our ``test'' agents, i.e., these are the agents for which we report the prediction error and eventual reward for the various RL algorithms.
Our rationale for selecting these five agents is as follows:
one of the five is the default covariate parameter in an environment; the other four are selected so as to cover the ``extremes'' of the parameter range.
Due to space constraints, we show results for three test agents in this section; the results for the remaining two agents can be found in Appendix \ref{appendix:sec:results}.
We note that the conclusions we draw from our experiments continue to hold over all test agents we evaluate on. 

\noindent {\bf Offline Data from Sub-optimal Policies.}
%
%
To study how robust the various RL algorithms are to the ``optimality'' of the sampling policy used to generate the historical trajectories, we create four offline datasets of 500 trajectories (one per agent) for each environment as follows:  

\noindent \textit{(i) Pure policy}. 
    For each agent, actions are sampled according to a fixed policy that has been trained to achieve ``good'' performance. 
    Specifically, for each environment, we first train a policy in an online fashion for a few training agents (see Table~\ref{table:env} for details).
    We pick the training agents to be uniformly spread throughout the training range to ensure reasonable performance for all agents. 
   See Appendix~\ref{appendix:sec:setup} for details about the average reward achieved across all agents using this procedure. 
    For MountainCar and CartPole, we use DQN~\citep{mnih2015human} to train the sampling policy;
    for HalfCheetah, we use TD3~\citep{fujimoto2018addressing}.
    The policies are trained to achieve rewards of approximately -200, 120, and 3000 for MountainCar, CartPole, and HalfCheetah, respectively.  
    Then for each of the 500 agents, we sample one trajectory using the policy trained on the training agent with the closest parameter value. 
\textit{(ii) Random}. Actions are selected uniformly at random.
\textit{(iii/iv) Pure-$\varepsilon$-20/Pure-$\varepsilon$-40}.
%
For Pure-$\varepsilon$-20/40, actions are selected uniformly at random with probability 0.2/0.4, respectively, and selected via the pure policy otherwise. 
%


    
    
%

\noindent 
The ``pure policy'' dataset has relatively optimal transition dynamics for each agent compared to the other three policies.
This is likely the ideal scenario when we only have access to limited offline data.
%
%
However, such an ideal sampling procedure is hardly met in practice.
Real-world data often contains at least some amount of ``trial and error'' in terms of how actions are chosen; we model this by sampling a fraction of the trajectory at random.

\colorred{
\noindent {\bf Benchmarking Algorithms.}
We compare with four state-of-the-art RL algorithms; (i) one from the online model-based literature; (ii) two from the offline model-free literature; (iii) and one offline model-based approach.
%
%
In Appendix~\ref{appendix:sec:implementation}, we give implementation details.
}

\noindent 
\textit{Vanilla CaDM + PE-TS CaDM}~\citep{lee2020context}. 
As aforementioned, we choose CaDM as a baseline given its superior performance against other popular model-based (and meta-learning) methods in handling heterogeneous agents. 
CaDM tackles heterogeneity by learning a context vector using the recent trajectory of a given agent, with a common context  encoder across all agents.
%
In our evaluation, we compare against two CaDM variants 
discussed in~\citep{lee2020context} with respect to both model prediction error and eventual reward. 

\noindent 
\textit{BCQ-P + BCQ-A}~\citep{fujimoto2019off}.
BCQ is an offline model-free RL method that has been shown to exhibit excellent performance in the standard offline setting. 
In light of this, we consider two BCQ baselines:
(i) BCQ-Population (or BCQ-P), where a {\em single} policy is trained using data from all available (heterogeneous) agents, i.e., all 500 observed trajectories; 
(ii) BCQ-Agent (or BCQ-A), where a separate policy is learned for each of the test agents using just the single observed trajectory associated with that test agent.  
We compare PerSim against both BCQ variants with respect to the eventual reward in order to study the  effect that data scarcity and agent heterogeneity can have on standard offline RL methods.

\colorred{
\noindent 
\textit{CQL-P + CQL-A}~\citep{kumar2020conservative}.
CQL is considered one of the state-of-the-art offline model-free RL methods.
Similar to BCQ, we consider two CQL baselines: CQL-P and CQL-A, which are defined analogously to BCQ-P and BCQ-A.

\noindent 
\textit{MOReL-P + MOReL-A}~\citep{kidambi2020morel}.
MOReL is a recent work that proposes a model-based approach for offline RL. 
MOReL does not consider a setting where data is collected from heterogeneous agents, unlike CaDM and PerSim. 
However, we include MOReL as a baseline to provide a more comprehensive evaluation. 
Similar to BCQ and CQL, we consider two MOReL baselines: MOReL-P and MOReL-A, which are defined analogously to BCQ-P and BCQ-A.
}
\vspace{-3mm}
\subsection{Model Learning: Prediction Error} \label{sec:prediction}
\vspace{-2mm}
The core of PerSim is to build a personalized simulator for each agent. 
This is also the case for two of the methods we compare against, Vanilla CaDM and PE-TS CaDM.
Thus, for each of these algorithms, we first evaluate the accuracy of the learned transition dynamics for each agent, focusing on long-horizon model prediction.
Specifically, given an initial state and an unseen sequence of 50 actions, the task is to predict the next 50-step state trajectory for the test agents.
The sequence of 50 actions are chosen according to an unseen test policy that is different than the policies used to sample the training dataset.  Precisely , the test  policies are fitted via DQN for MountainCar and CartPole and via TD3 for HalfCheetah for the agent with the default covariate parameters. 
The test policies were such that they achieved rewards of -150, 150, and 4000 for MountainCar, CartPole and HalfCheetah, respectively.

\noindent 
For each environment and for PerSim, Vanilla CaDM, and PE-TS CaDM, we train four separate models using each of the four offline datasets described earlier.
We repeat the process 200 times (totaling 200 trajectories for each test agent) and report the \textit{mean} root-mean-squared error (RMSE) over all 50-steps and over all 200 trajectories. 
In addition to RMSE, we provide the \textit{median} $R^2$ (in parentheses within the tables) to facilitate a better comparison in terms of relative error. 
The results are summarized in Table \ref{table:prediction_error} for MountainCar, CartPole, and HalfCheetah. 
Further experimental results regarding prediction error can be found in {Appendix \ref{appendix:sec:prediction_results}}.

\begin{table}
\vspace{-5mm}
\caption{Prediction error.}
\label{table:prediction_error}
\tabcolsep=0.05cm
\fontsize{7.0pt}{10.0pt}\selectfont
\centering
    \addtolength{\leftskip} {-2cm}
    \addtolength{\rightskip}{-2cm}
\begin{tabular}{@{}p{0.5cm}p{1.55cm}p{1.35cm}p{1.35cm}p{1.55cm}p{1.35cm}p{1.35cm}p{1.55cm}p{1.35cm}p{1.35cm}p{1.35cm}@{}}
\toprule
& &\multicolumn{3}{c}{MountainCar}&\multicolumn{3}{c}{CartPole}  &\multicolumn{3}{c}{HalfCheetah} 
\\  \cmidrule[0.8pt](r){3-5} \cmidrule[0.8pt](r){6-8}
\cmidrule[0.8pt](r){9-11}
{Data}                     &   Method     & {0.01 }   & {0.025}               & {0.035}           & {(2/0.5) } & {(10/0.85)}   & {(10/0.15)}     &  {(0.3/1.7)} & {(1.7/0.3)}   & {(0.3/0.3)}  \\ \midrule
\multirow{3}{*}{\rotatebox[origin=c]{90}{{{Pure}} }}      
                           & PerSim   &  {0.025 (0.97)} &  {0.014 (0.94)} &  {0.039 (0.80)}  & {0.01 (1.00)} &   {0.01 (1.00)} &   {0.035 (1.00)} & {1.194 (0.92)} &  4.064 (0.67) & {4.070 (0.81)} \\
                           & Vanilla CaDM & 0.149 (0.74)&0.177 (-18.4)&0.238 (-30.1)   & 0.403 (0.71)&0.039 (0.98)&2.531 (-0.53) & 3.902 (0.47) & 3.851(0.49) & 6.308 (0.38)  \\      
                           & PE-TS CaDM   & 0.326 (-1.91)&0.154 (-1.11)&0.148 (-0.18)     &0.031 (0.99) &0.06 (1.00)&0.319 (0.65)  & 3.147 (0.61) & {3.080 (0.68)} & 5.270 (0.57)   \\\midrule

\multirow{3}{*}{\rotatebox[origin=c]{90}{{{Random}} }} 
& PerSim    & {0.04 (1.00)} & {0.01 (1.00)} &  {0.01 (1.00) } &  {0.014 (0.98)} &  {0.06 (1.00)} &  {0.152 (0.88)}  &  {1.194 (0.92)} &  4.064 {(0.67)} & {4.070 (0.81)}  \\
& Vanilla CaDM &  0.256 (0.43)&0.134 (0.71)&0.217 (-1.77)  &0.172 (0.10)&0.098 (0.64)&3.307 (-0.73)  & 4.030 (0.44) & 4.121 (0.43) & 4.446 (0.67)        \\
& PE-TS CaDM   & 0.242 (0.31)&0.101 (0.87)&0.075 (0.97)      &0.564 (-1.36)&0.216 (0.45)&3.764 (-1.10) & 2.735 (0.73) & {2.756} {(0.69)} & 4.141 (0.77)   \\\midrule

\multirow{3}{*}{\rotatebox[origin=c]{90}{{{Pure-$\varepsilon$-20}} }} 
& PerSim     &  {0.04 (1.00)} & {0.02 (1.00)} &  {0.04 (1.00)}  &  {0.0 (1.00)} & {0.01 (1.00)} &  {0.048 (0.98)}  & {1.172  (0.92)} &  4.283 {(0.66)} &  {3.832 (0.84)}  \\
& Vanilla CaDM &  0.227 (0.31)&0.101 (-0.37)&0.157 (-2.25)  &0.270 (-0.98)&0.058 (0.94)&2.193 (-0.43) &3.613 (0.53) & 3.455 (0.54) & 6.046 (0.48)     \\
& PE-TS CaDM   &  0.350 (-2.57)&0.139 (0.75)&0.130 (0.89) &0.639 (-1.21)&0.148 (0.38)&2.680 (-0.56) &2.913 (0.70) & {2.959} (0.65) & 4.818 (0.65)    \\\midrule

\multirow{3}{*}{\rotatebox[origin=c]{90}{{{Pure-$\varepsilon$-40}} }}  
& PerSim   &  {0.06 (1.00)}&{0.06 (0.99)} &  {0.03 (1.00)} &   {0.0 (1.00)} &  {0.0 (1.00)} &   {0.018 (1.00)} &  {1.016 (0.94)} &  4.021 {(0.66)} &  {3.742 (0.85)}  \\
& Vanilla CaDM & 0.199 (0.54)&0.119 (0.38)&0.187 (-9.00) &0.233 (-0.88)&0.035 (0.99)&3.286 (-0.62) & 3.685 (0.49) & 3.612 (0.48) & 6.000 (0.39)      \\
& PE-TS CaDM   & 0.639 (-2.27)&0.192 (-6.06)&0.157 (0.55) &0.051 (0.98)&0.010 (1.00)&0.411 (0.55)  &3.021 (0.67) & {3.025 }(0.61) & 5.075 (0.62) \\ \cmidrule[1.5pt]{1-11}

\end{tabular}
\vspace{-3mm}
\end{table}

%
\noindent {\bf PerSim Accurately Learns Personalized Simulators.} 
Consistently across environments, PerSim  outperforms both Vanilla CaDM and PE-TS CaDM for most test agents.
RMSE for PerSim is notably lower by orders of magnitude in MountainCar and CartPole. 
%
Indeed, in these two environments, $R^2$ for PerSim is nearly one (i.e., the maximum achievable) across most agents, while for the two CaDM variants it is notably lower. In a number of experiments, the two CaDM variants have negative $R^2$ values, i.e., their predictions are worse than simply predicting the average true state across the test trajectory.
For HalfCheetah, 
which has a relatively high-dimensional state space (it is 18-dimensional), PerSim 
continues to deliver reasonably good predictions for each agent despite the challenging data availability.
Though PerSim still outperforms CaDM in HalfCheetah, we note CaDM's performance is more comparable in this environment.

\noindent 
For a more visual representation of PerSim's prediction accuracy, refer back to Figure~\ref{fig:visulization_main_paper} in Section \ref{sec:introduction}, which shows the predicted and actual state for different horizon lengths in CartPole; there, we see that PerSim consistently produces state predictions that closely match the actual state observed from the true environment. 
Additional visualizations across environments are provided in Appendix~\ref{appendix:sec:prediction_results}

\noindent {\bf Rethinking Model-based RL for Our Setting.}
{
{Altogether, 
these results indicate that existing model-based RL methods cannot effectively learn a model of the transition dynamics simultaneously across {\em all} heterogeneous agents (e.g., CaDM sometimes has negative $R^2$ values) if only given access to sparse offline data.
It has been exhibited~\citep{janner2019trust,yu2020mopo,levine2020offline} that certain model-based methods 
that were originally targeted for the online setting can potentially still deliver reasonable performance with 
offline data and minimal algorithmic change.
Our experiments offer evidence that model-based RL methods, even those which are optimized to work with heterogeneous agents, do not provide a uniformly good ``plug-in'' solution for the particular data availability we consider.
In contrast, our latent factor approach consistently learns a reasonably good model of the dynamics across all agents.
%
}}


\noindent {\bf Latent Factors Capture Agent Heterogeneity.}
The poor performance of standard model-based RL methods for the data availability setting we consider emphasizes the need for new principled approaches.
Ours is one such approach, where we posit a low-rank latent factor representation of the agents, states, and actions.
To further validate our approach, we visualize the learned latent agent factors associated with the 500 heterogeneous agents in each environment in Figure~\ref{fig:LF_main_paper} of Section \ref{sec:introduction}.
Pleasingly, across environments, the latent factors correspond closely with the heterogeneity across agents.

\vspace{-4mm}
\subsection{Policy Performance: Average Reward} \label{subsec:experiment_reward}
\vspace{-2mm}
In this section, we evaluate the average reward achieved by PerSim compared to several state-of-the-art model-based and model-free offline RL methods.
For the PerSim and the two CaDM variants, we follow \cite{lee2020context} in utilizing MPC to make policy decisions on top of the learned model. 
We include an additional baseline, ``True env + MPC'', where we apply MPC on top of the actual ground-truth environment; this allows us to quantify the difference in reward when using MPC with the actual environment versus the learned simulators. 
\colorred{For the model-free methods (e.g., BCQ and CQL variants), we can directly apply the policy resulted from the learned $Q$-value.
Likewise, we directly apply the policy trained by MOReL. We use the trained policy for this model-based method instead of utilizing MPC to faithfully follow the method described in \cite{kidambi2020morel}.}

\noindent 
We evaluate the performance of each method using the average reward over 20 episodes for  model-based methods and the average over 100 episodes for model-free methods. 
We repeat each experiment 
five times and report the mean and standard deviation.
Table \ref{table:avg_rew} presents the results for MountainCar, CartPole, and HalfCheetah. 
\noindent 
Further experimental results regarding reward are given in  Appendix \ref{appendix:sec:average_reward}.
\begin{table}
\vspace{-4mm}
\caption{Average reward.}
\label{table:avg_rew}
\tabcolsep=0.05cm
\fontsize{7.0pt}{10.5pt}\selectfont
\centering
    \addtolength{\leftskip} {-2cm}
    \addtolength{\rightskip}{-2cm}
\begin{tabular}{@{}p{0.5cm}p{1.6cm}p{1.4cm}p{1.4cm}p{1.6cm}p{1.4cm}p{1.4cm}p{1.6cm}p{1.4cm}p{1.4cm}p{1.4cm}@{}}
\toprule
& &\multicolumn{3}{c}{MountainCar}&\multicolumn{3}{c}{CartPole}  &\multicolumn{3}{c}{HalfCheetah} 
\\  \cmidrule[0.8pt](r){3-5} \cmidrule[0.8pt](r){6-8}
\cmidrule[0.8pt](r){9-11}
{Data}                     &   Method     & {0.01 }   & {0.025}               & {0.035}           & {(2/0.5) } & {(10/0.85)}   & {(10/0.15)}     &  {(0.3/1.7)} & {(1.7/0.3)}   & {(0.3/0.3)}  \\ \midrule
\multirow{9}{*}{\rotatebox[origin=c]{90}{{{Pure}} }}      
& PerSim   &  -56.80$\pm$\scriptsize{1.83} & {-189.4}$\pm$\scriptsize{6.44} & {-210.6}$\pm$\scriptsize{4.27}  &{199.7}$\pm$\scriptsize{0.58} & {193.8}$\pm$\scriptsize{4.28} & {192.0}$\pm$\scriptsize{2.28} & 
~{1984}~$\pm$\scriptsize{763} &	~997.0$\pm$\scriptsize{403} &	~{714.7}$\pm$\scriptsize{314} \\
& Vanilla CaDM & -106.3$\pm$\scriptsize{44.1} &  -432.3$\pm$\scriptsize{117}  &  -471.8$\pm$\scriptsize{43.0}  
    & 168.0$\pm$\scriptsize{19.7} &    190.8$\pm$\scriptsize{6.80}  &   58.10$\pm$\scriptsize{10.7} & 
    ~50.31$\pm$\scriptsize{71.7}  & -134.0$\pm$\scriptsize{81.1}  & ~11.39$\pm$\scriptsize{171}  \\      
& PE-TS CaDM   & -74.23$\pm$\scriptsize{16.5}  & -492.3$\pm$\scriptsize{13.3} &  -500.0$\pm$\scriptsize{0.0}    
&92.30$\pm$\scriptsize{44.8}         &  193.6$\pm$\scriptsize{8.30}            &  127.5$\pm$\scriptsize{9.50}  & 
~481.1$\pm$\scriptsize{252} & ~503.7$\pm$\scriptsize{181} & ~553.0$\pm$\scriptsize{127} \\
& BCQ-P &  -67.60$\pm$\scriptsize{22.3}  & -267.8$\pm$\scriptsize{202} & -295.1$\pm$\scriptsize{180} &166.2$\pm$\scriptsize{39.3}  &  181.2$\pm$\scriptsize{13.5}   &   182.8$\pm$\scriptsize{15.0} &{~549.8}$\pm$\scriptsize{322} & 	{~2006~}$\pm$\scriptsize{153} & -65.18$\pm$\scriptsize{92.8} \\
& BCQ-A &  {-44.79}$\pm$\scriptsize{0.08}  & 	-380.7$\pm$\scriptsize{170}  & -500.0$\pm$\scriptsize{0.0} & 65.40$\pm$\scriptsize{67.5} & 79.20$\pm$\scriptsize{69.5}    &   132.1$\pm$\scriptsize{85.0} &-262.7$\pm$\scriptsize{96.6} & -139.0$\pm$\scriptsize{236} & ~165.6$\pm$\scriptsize{83.1} \\
&CQL-P& -176.1$\pm$\scriptsize{45.2}  &-316.4$\pm$\scriptsize{26.4} &-362.9$\pm$\scriptsize{17.9} & 154.4$\pm$\scriptsize{17.7} &190.9$\pm$\scriptsize{0.7} &170.2$\pm$\scriptsize{0.6}  & -353.5$\pm$\scriptsize{78.4} &-453.6$\pm$\scriptsize{71.9} &-476.7$\pm$\scriptsize{129} \\
&CQL-A&  -44.60$\pm$\scriptsize{0.0} &-500.0$\pm$\scriptsize{0.0} &-499.3$\pm$\scriptsize{0.7} &  122.8$\pm$\scriptsize{63.4} &179.4$\pm$\scriptsize{20.6} &193.6$\pm$\scriptsize{5.4} & -65.00$\pm$\scriptsize{105} &-257.9$\pm$\scriptsize{35.9} &-279.6$\pm$\scriptsize{34.4}\\
&MOReL-P& -46.00$\pm$\scriptsize{1.1} &-500.0$\pm$\scriptsize{0.0} &-500.0$\pm$\scriptsize{0.0} & 35.80$\pm$\scriptsize{1.4}  &96.60$\pm$\scriptsize{24.1} &66.40$\pm$\scriptsize{24.5} & -1297~$\pm$\scriptsize{519} &-1256~$\pm$\scriptsize{627} &-1470~$\pm$\scriptsize{727}  \\
&MOReL-A &-373.0$\pm$\scriptsize{33.5}	&-500.0$\pm$\scriptsize{0.0}	&-500.0$\pm$\scriptsize{0.0} & 33.70$\pm$\scriptsize{3.8}  &27.50$\pm$\scriptsize{0.1} &10.10$\pm$\scriptsize{0.7} & -726.2$\pm$\scriptsize{4.9} &-666.1$\pm$\scriptsize{42.6} &-841.7$\pm$\scriptsize{39.5} 
\\\midrule
\multirow{9}{*}{\rotatebox[origin=c]{90}{{{Random}} }} 
& PerSim    & {-57.70}$\pm$\scriptsize{5.63} & {-186.6}$\pm$\scriptsize{4.25} & {-210.1}$\pm$\scriptsize{4.48} &  
                {197.7}$\pm$\scriptsize{7.82} & 193.0$\pm$\scriptsize{6.60} & {185.7}$\pm $ \scriptsize{3.49}  &  
                    ~{2124}~$\pm$\scriptsize{518}&	~{2060}~$\pm$\scriptsize{900}&	~472.0$\pm$\scriptsize{56.9}  \\
& Vanilla CaDM &  -62.57$\pm$\scriptsize{5.11}  &  -479.3$\pm$\scriptsize{21.7}  &  -497.5$\pm$\scriptsize{4.39}  &
                150.5$\pm$\scriptsize{15.7} &     175.6$\pm$\scriptsize{5.80}  &    65.10$\pm$\scriptsize{16.3}  & 
                ~288.4$\pm$\scriptsize{32.4}  & ~362.9$\pm$\scriptsize{55.4}  & ~351.8$\pm$\scriptsize{34.9}        \\
& PE-TS CaDM   & -82.00$\pm$\scriptsize{3.47}  & -500.0$\pm$\scriptsize{0.0} &  -500.0$\pm$\scriptsize{0.0}      &
                88.60$\pm$\scriptsize{18.5}    &     {196.1}$\pm$\scriptsize{3.00}    &     171.0$\pm$\scriptsize{21.7} &
                ~754.6$\pm$\scriptsize{242} & ~744.5$\pm$\scriptsize{281} & {~767.4}$\pm$\scriptsize{214}   \\
& BCQ-P &       -500.0$\pm$\scriptsize{0.0}   & -500.0$\pm$\scriptsize{0.0}   &   -500.0$\pm$\scriptsize{0.0} &
                44.80$\pm$\scriptsize{34.0}   &    57.91$\pm$\scriptsize{53.0}    &    36.40$\pm$\scriptsize{36.0} &
                -1.460$\pm$\scriptsize{0.16} & -1.750$\pm$\scriptsize{0.22} & -1.690$\pm$\scriptsize{0.19} \\
& BCQ-A &       -500.0$\pm$\scriptsize{0.0} &   -500.0$\pm$\scriptsize{0.0} &  -500.0$\pm$\scriptsize{0.0} &
                43.90$\pm$\scriptsize{16.4}   &    21.10$\pm$\scriptsize{5.81}  &    39.50$\pm$\scriptsize{12.1} & -498.9$\pm$\scriptsize{108} & -113.3$\pm$\scriptsize{13.0} & -159.5$\pm$\scriptsize{51.7}\\
&CQL-P&  -500.0$\pm$\scriptsize{0.0}  &-500.0$\pm$\scriptsize{0.0} &-500.0$\pm$\scriptsize{0.0} &  39.90$\pm$\scriptsize{29.9}  &77.90$\pm$\scriptsize{26.2} &148.7$\pm$\scriptsize{17.1} &-481.5$\pm$\scriptsize{25.9} &-442.2$\pm$\scriptsize{56.4} &-672.4$\pm$\scriptsize{17.6} \\ 
&CQL-A&  -500.0$\pm$\scriptsize{0.0}  &-500.0$\pm$\scriptsize{0.0} &-500.0$\pm$\scriptsize{0.0} &  67.40$\pm$\scriptsize{49.1} &30.60$\pm$\scriptsize{26.4} &7.000$\pm$\scriptsize{1.9} & -0.700$\pm$\scriptsize{0.4} &-2.800$\pm$\scriptsize{0.8} &-0.600$\pm$\scriptsize{0.3}  \\
&MOReL-P& -45.00$\pm$\scriptsize{0.3}  &-500.0$\pm$\scriptsize{0.0} &-500.0$\pm$\scriptsize{0.0} & 50.20$\pm$\scriptsize{7.9} &68.90$\pm$\scriptsize{17.4} &40.40$\pm$\scriptsize{12.4} & -102.3$\pm$\scriptsize{45.7} &-188.7$\pm$\scriptsize{37.1} &-181.0$\pm$\scriptsize{67.7} \\
&MOReL-A &-495.0$\pm$\scriptsize{3.9}&-500.0$\pm$\scriptsize{0.0}	&-500.0$\pm$\scriptsize{0.0} & 35.80$\pm$\scriptsize{0.4} &27.50$\pm$\scriptsize{0.7} &10.60$\pm$\scriptsize{0.2} & -430.8$\pm$\scriptsize{195} &-673.7$\pm$\scriptsize{39.3} &-365.5$\pm$\scriptsize{97.4}
\\\midrule

\multirow{9}{*}{\rotatebox[origin=c]{90}{{{Pure-$\varepsilon$-20}} }} 
& PerSim     &  {-54.20}$\pm$\scriptsize{0.56} & {-191.2}$\pm$\scriptsize{6.70} & {-199.7}$\pm$\scriptsize{3.99}  &
                {199.8}$\pm$\scriptsize{0.24} & {199.1}$\pm$\scriptsize{1.30} & {197.8}$\pm$\scriptsize{1.68}  &
                ~{3186}~$\pm$\scriptsize{604}& ~1032$\pm$\scriptsize{232}& ~{1121}$\pm$\scriptsize{243}  \\
& Vanilla CaDM &  -56.73$\pm$\scriptsize{4.20} &  -463.2$\pm$\scriptsize{57.5}  &  -478.9$\pm$\scriptsize{35.8}  &
                171.1$\pm$\scriptsize{38.1} &     193.4$\pm$\scriptsize{2.10}  &    64.20$\pm$\scriptsize{10.0} &
                   ~412.0$\pm$\scriptsize{152}  & ~31.92$\pm$\scriptsize{109}  & ~460.2$\pm$\scriptsize{159}     \\
& PE-TS CaDM   &  -107.6$\pm$\scriptsize{36.3} & -500.0$\pm$\scriptsize{0.0} &  -500.0$\pm$\scriptsize{0.0} &
                    98.30$\pm$\scriptsize{42.9} &      198.6$\pm$\scriptsize{0.40}  &    141.1$\pm$\scriptsize{12.0} &
                    {~1082~}$\pm$\scriptsize{126} & ~{1125}~$\pm$\scriptsize{132} & {~1067~}$\pm$\scriptsize{64.3}   \\ 
& BCQ-P &    -71.21$\pm$\scriptsize{24.4}  &      -286.6$\pm$\scriptsize{196}   &     -328.3$\pm$\scriptsize{158} &
            98.90$\pm$\scriptsize{30.2}      &   162.1$\pm$\scriptsize{15.5}   &     86.10$\pm$\scriptsize{72.1} &
            ~254.6$\pm$\scriptsize{352} & ~406.7$\pm$\scriptsize{71.1} & ~385.9$\pm$\scriptsize{57.1} \\
& BCQ-A &   -364.5$\pm$\scriptsize{180}  & -260.6$\pm$\scriptsize{51.4} &   -204.5$\pm$\scriptsize{68.9} &
            67.30$\pm$\scriptsize{62.2}      &65.60$\pm$\scriptsize{51.8}        & 140.0$\pm$\scriptsize{80.6} &
            ~376.8$\pm$\scriptsize{102} & ~84.66$\pm$\scriptsize{53.3} & ~230.1$\pm$\scriptsize{10.0} \\
&CQL-P& -79.40$\pm$\scriptsize{16.1} &-357.9$\pm$\scriptsize{12.5} &-407.6$\pm$\scriptsize{15.3}  & 163.7$\pm$\scriptsize{13.6}&198.1$\pm$\scriptsize{2.6} &190.4$\pm$\scriptsize{6.6}  & ~838.7$\pm$\scriptsize{24.5} &~3155~$\pm$\scriptsize{125} &~539.9$\pm$\scriptsize{313}   \\  
&CQL-A& -44.70$\pm$\scriptsize{0.1} &-500.0$\pm$\scriptsize{0.0} &-500.0$\pm$\scriptsize{0.0}  &  42.40$\pm$\scriptsize{11.9} &199.0$\pm$\scriptsize{1.9} &199.8$\pm$\scriptsize{0.2} & -15.50$\pm$\scriptsize{9.0} &-73.00$\pm$\scriptsize{26.3} &-108.0$\pm$\scriptsize{64.6} \\
&MOReL-P&  -83.50$\pm$\scriptsize{15.6}  &-357.0$\pm$\scriptsize{13.8} &-407.4$\pm$\scriptsize{17.1}  & 166.7$\pm$\scriptsize{13.6}  &197.7$\pm$\scriptsize{2.7} &189.4$\pm$\scriptsize{7.1} & ~0.600$\pm$\scriptsize{209} &-171.2$\pm$\scriptsize{125} &-219.9$\pm$\scriptsize{76.7}  \\
&MOReL-A & -44.60$\pm$\scriptsize{0.1} &-500.0$\pm$\scriptsize{0.0} &-500.0$\pm$\scriptsize{0.0}  & 41.40$\pm$\scriptsize{13.1} &198.8$\pm$\scriptsize{2.1} &199.9$\pm$\scriptsize{0.2}  & -781.0$\pm$\scriptsize{37.9} &-613.1$\pm$\scriptsize{49.9} &-847.1$\pm$\scriptsize{64.7} 
\\\midrule

\multirow{9}{*}{\rotatebox[origin=c]{90}{{{Pure-$\varepsilon$-40}} }}  
& PerSim   &  -54.60$\pm$\scriptsize{0.55} & {-189.7}$\pm$\scriptsize{7.14} & {-200.3}$\pm$\scriptsize{2.26} &
            {199.9}$\pm$\scriptsize{0.18} & {198.0}$\pm$\scriptsize{1.21} & {197.4}$\pm$\scriptsize{1.72} &
                ~{2590}~$\pm$\scriptsize{813}&	~1016~$\pm$\scriptsize{283}&	~{1365}~$\pm$\scriptsize{582}  \\
& Vanilla CaDM & -55.23$\pm$\scriptsize{0.76} &  -481.7$\pm$\scriptsize{25.3}  &  -496.2$\pm$\scriptsize{4.31}  &
           160.6$\pm$\scriptsize{46.6} &    191.9$\pm$\scriptsize{6.40}&   79.60$\pm$\scriptsize{31.6} &
                ~465.6$\pm$\scriptsize{49.2}   & ~452.7$\pm$\scriptsize{130}  & ~720.0$\pm$\scriptsize{74.9}      \\
& PE-TS CaDM   & -102.3$\pm$\scriptsize{20.3}  & -500.0$\pm$\scriptsize{0.0} &  -500.0$\pm$\scriptsize{0.0} &
                91.90$\pm$\scriptsize{67.6}   & 197.0$\pm$\scriptsize{1.40}   &   143.5$\pm$\scriptsize{17.5}  &
                   {~1500~}$\pm$\scriptsize{246} & ~{1218}~$\pm$\scriptsize{221} & {~1339~}$\pm$\scriptsize{54.8} \\
& BCQ-P &     {-50.01}$\pm$\scriptsize{7.50}                  &   -373.6$\pm$\scriptsize{180}            &     -352.0$\pm$\scriptsize{211} &
            28.90$\pm$\scriptsize{6.80} &    31.80$\pm$\scriptsize{25.9}             &       18.50$\pm$\scriptsize{11.1} &
            ~78.25$\pm$\scriptsize{200}   & ~173.8$\pm$\scriptsize{189}& ~417.1$\pm$\scriptsize{155} \\
& BCQ-A &   -94.87$\pm$\scriptsize{0.88}            &    -358.7$\pm$\scriptsize{201}                &     -486.5$\pm$\scriptsize{20.6} &
            34.60$\pm$\scriptsize{1.55}              &    47.71$\pm$\scriptsize{48.7}                &     23.20$\pm$\scriptsize{9.44} &
            ~269.2$\pm$\scriptsize{60.7}  & -181.5$\pm$\scriptsize{57.4}& ~193.0$\pm$\scriptsize{31.8} \\
&CQL-P&-61.40$\pm$\scriptsize{2.1} &-366.9$\pm$\scriptsize{21.3} &-429.1$\pm$\scriptsize{18.8} & 182.2$\pm$\scriptsize{18.0}&
198.3$\pm$\scriptsize{2.9} &191.9$\pm$\scriptsize{8.7} & ~808.5$\pm$\scriptsize{240} &~1662~$\pm$\scriptsize{221} &-156.3$\pm$\scriptsize{119}  \\ 
&CQL-A& -44.70$\pm$\scriptsize{0.0}  &-500.0$\pm$\scriptsize{0.0} &-490.1$\pm$\scriptsize{9.6} &   20.70$\pm$\scriptsize{0.7} &134.2$\pm$\scriptsize{10.2} &9.700$\pm$\scriptsize{8.1} & -6.200$\pm$\scriptsize{2.9} &-386.0$\pm$\scriptsize{42.4} &~37.10$\pm$\scriptsize{154} \\
&MOReL-P& -61.30$\pm$\scriptsize{2.3}  &-373.2$\pm$\scriptsize{19.3} &-428.5$\pm$\scriptsize{21.0} & 178.9$\pm$\scriptsize{18.7} &197.9$\pm$\scriptsize{3.1} &190.6$\pm$\scriptsize{9.3} & ~8.500$\pm$\scriptsize{61.6} &-114.2$\pm$\scriptsize{72.2} &-195.9$\pm$\scriptsize{77.3}  \\
&MOReL-A & -44.70$\pm$\scriptsize{0.0} &-500.0$\pm$\scriptsize{0.0} &-492.1$\pm$\scriptsize{9.3} & 20.70$\pm$\scriptsize{0.8}  &135.4$\pm$\scriptsize{11.2} &10.80$\pm$\scriptsize{8.8}     & -325.9$\pm$\scriptsize{17.1} &-644.5$\pm$\scriptsize{18.8} &-798.8$\pm$\scriptsize{130}     
\\ \cmidrule[1.5pt]{1-11}

&True env+MPC& {-53.95$\pm$\scriptsize{4.10}}   & {-182.9$\pm$\scriptsize{22.9}} & {-197.5$\pm$\scriptsize{20.7}} &
                {200.0$\pm$\scriptsize{0.0}}  & {198.4$\pm$\scriptsize{7.20}}& {200.0$\pm$\scriptsize{0.0}} &
                {~7459~$\pm$\scriptsize{171}}  & {42893$\pm$\scriptsize{6959}}  & {66675$\pm$\scriptsize{9364}}
                \\\cmidrule[1.5pt]{1-11}
\end{tabular}
\vspace{-4mm}
\end{table}

{
\noindent 
As a high-level summary, in most 
experiments, {\em PerSim either achieves the best reward or close to it.}
\colorred{
This not only reaffirms our prediction results in Section~\ref{sec:prediction}, 
but also is particularly encouraging for PerSim since the policy utilized is simply MPC and not optimized as done in model-free approaches and in MOReL.}}
Furthermore, these results corroborate the appropriateness of our low-rank latent factor representation and our overall methodological framework as a principled solution to this challenging yet meaningful setting within offline RL. 
In what follows, we highlight additional interesting conclusions.

\noindent {\bf ``Solved'' Environments are Not Actually Solved.} 
MountainCar and CartPole are commonly perceived as simple, ``solved'' environments within the RL literature. 
Yet, results in Table \ref{table:avg_rew} demonstrate that the offline setting with 
scarce and heterogeneous data impose unique challenges, and undoubtedly warrants a new methodology. 
{ We find state-of-the-art model-based and model-free methods perform poorly on some of the test agents in MountainCar and CartPole.
In comparison, 
%
PerSim's performance in these environments is close to that of MPC planning using the ground-truth environment across all test agents, thereby confirming the success of learning the personalized simulators in Step 1 of our algorithm. 
\colorred{In certain rare cases (e.g., MountainCar test agent 0.01) where other baselines have a comparable performance to PerSim, we see that these baselines even outperforms True env + MPC. 
This indicates that the bottleneck in these experiments is using MPC for policy planning rather than the learned simulator in PerSim.}

%
}

\noindent {\bf PerSim Robustly Extrapolates with Sub-optimal  Data.}
Crucially, across all environments and offline data generating processes, PerSim remains the most {\em consistent and robust} winner. 
This is a much desired property for offline RL.
In real-world applications, the policy used to generate the offline dataset is likely unknown. 
Thus, for broad applicability of a RL methodological framework, it is vital that it is robust to sub-optimality in how the dataset was generated, e.g., the dataset may contain a significant amount of ``trial and error'' (i.e, randomized actions).
Indeed, this is one of the primary motivations to use RL in such settings in the first place.

\colorred{
\noindent 
As mentioned earlier, the four offline datasets correspond to varying degrees of ``optimality'' of the policy used to sample agent trajectories. 
We highlight that PerSim achieves uniformly good reward, even with random data, i.e., the offline trajectories are produced using totally random actions. 
This showcases that by first learning a personalized simulator for each agent, PerSim is able to robustly {\em extrapolate} outside the policy used to generate the data, however sub-optimal that policy might be.
In contrast, BCQ and CQL are not robust to such sub-optimality in the offline data generation. For example, 
for HalfCheetah, BCQ achieves reasonable performance primarily when trained on ``optimal'' 
offline data (i.e., pure policy). This is because BCQ, by design, is conservative and is regularized to only pick actions that are close to what is seen in the offline data.} 
In summary, PerSim's ability to successfully extrapolate, even with sub-optimal offline data, makes it a suitable candidate for
real-world applications.


%
\colorred{
\vspace{-2mm}
\subsection{Combination with Model-Free Methods: PerSim+BCQ/CQL}
\label{sec:BCQ_persim}
\vspace{-2mm}
We explore whether simulated trajectories produced 
from PerSim can be used to improve the performance of model-free RL methods such as BCQ and CQL.
%
%
In particular, instead of using a single observed trajectory to learn an agent-specific policy, as is done in BCQ-A and CQL-A, we use PerSim to generate $5$ ``synthetic'' trajectories for that agent.
We then use both the synthetic and the observed 
trajectories to train both BCQ and CQL (denoted by PerSim-BCQ-5 and PerSim-CQL-5, respectively). 
Note that using model-based methods to augment the data used by a model-free method has been explored in the literature \cite{model_based_aug1,model_based_aug2,janner2019trust}.

\noindent 
If the learned model is accurate, improvements for BCQ and CQL are naturally anticipated. Table \ref{table:bcq_persim_MC} confirms that this is indeed the case for PerSim. 
%
Across all environments, augmenting the training data with PerSim results in a significantly better average reward for most test agents. 
Specifically, the performance of PerSim-BCQ-5  and PerSim-CQL-5 indicates that a personalized BCQ/CQL policy trained on few PerSim-generated trajectories is superior to: (i) a single BCQ/CQL policy trained using all 500 trajectories (e.g., BCQ-P); and (ii) a personalized BCQ/CQL policy trained using a single observed trajectory (e.g., BCQ-A).  
Refer to Appendix \ref{appendix:bcq_persim} for more details about the experiment. 

}
\begin{table}[h]
\vspace{-4mm}
\caption{BCQ+PerSim}
\label{table:bcq_persim_MC}
\tabcolsep=0.02cm
\fontsize{7.0pt}{8.0pt}\selectfont
\centering
\begin{tabular}{@{}p{1.5cm}p{2.0cm}p{2.0cm}p{2.0cm}p{2.0cm}p{2.0cm}p{2.0cm}@{}}
\cmidrule[1.4pt]{1-7}
{Environment}& {Method} & {0.01 }  & {0.05}   & {0.01} & {0.025}   & {0.035}    \\ \toprule
\multirow{3}{*}{\rotatebox[origin=c]{0}{{{MountainCar}}}} 
&BCQ-P& -500.0$\pm$\scriptsize{0.0}                 &   -500.0$\pm$\scriptsize{0.0}                   &    -500.0$\pm$\scriptsize{0.0}                   & -500.0$\pm$\scriptsize{0.0}                    &   -500.0$\pm$\scriptsize{0.0}                     \\
&BCQ-A&                 -500.0$\pm$\scriptsize{0.0} & -500.0$\pm$\scriptsize{0.0} &           -500.0$\pm$\scriptsize{0.0} &   -500.0$\pm$\scriptsize{0.0} &  -500.0$\pm$\scriptsize{0.0} \\
&PerSim-BCQ-5& \textbf{-68.84}$\pm$\scriptsize{15.2} & \textbf{-82.60}$\pm$\scriptsize{7.00} & \textbf{-209.4}$\pm$\scriptsize{206} & -498.7$\pm$\scriptsize{1.89} & -500.0$\pm$\scriptsize{0.0}   \\
&CQL-P&  -500.0$\pm$\scriptsize{0.0} &-500.0$\pm$\scriptsize{0.0} &-500.0$\pm$\scriptsize{0.0} &-500.0$\pm$\scriptsize{0.0} &-500.0$\pm$\scriptsize{0.0} \\ 
&CQL-A&  -500.0$\pm$\scriptsize{0.0} &-500.0$\pm$\scriptsize{0.0} &-500.0$\pm$\scriptsize{0.0} &-500.0$\pm$\scriptsize{0.0} &-500.0$\pm$\scriptsize{0.0} \\
& PerSim-CQL-5&  \textbf{-44.70}$\pm$\scriptsize{0.1} &\textbf{-49.80}$\pm$\scriptsize{0.0} &\textbf{-63.20}$\pm$\scriptsize{0.1} &-500.0$\pm$\scriptsize{0.0} &-500.0$\pm$\scriptsize{0.0} \\
\cmidrule[1.4pt]{1-7}

{Environment}& {Method}& {(2/0.5) } & {(10.0/0.5) } & {(18.0/0.5)} & {(10/0.85)}   & {(10/0.15)}     \\ \toprule
\multirow{3}{*}{\rotatebox[origin=c]{0}{{{CartPole}} }} 
&BCQ-P&   44.80$\pm$\scriptsize{34.0}    &    58.21$\pm$\scriptsize{58.0}   &    56.92$\pm$\scriptsize{56.0}    &    57.91$\pm$\scriptsize{53.0}    &    36.40$\pm$\scriptsize{36.0}     \\
&BCQ-A&   43.90$\pm$\scriptsize{16.4} &    18.70$\pm$\scriptsize{13.1} &    7.200$\pm$\scriptsize{0.84}   &    21.10$\pm$\scriptsize{5.81}  &    39.50$\pm$\scriptsize{12.1}       \\
& PerSim-BCQ-5&   \textbf{80.04}$\pm$\scriptsize{5.68} & \textbf{95.29}$\pm$\scriptsize{29.5} & \textbf{63.86}$\pm$\scriptsize{17.6} & \textbf{92.47}$\pm$\scriptsize{19.9} & \textbf{57.82}$\pm$\scriptsize{19.8} \\
&CQL-P&  39.90$\pm$\scriptsize{29.9} &72.30$\pm$\scriptsize{41.7} &67.80$\pm$\scriptsize{44.4} &77.90$\pm$\scriptsize{26.2} &148.7$\pm$\scriptsize{17.1} \\ 
&CQL-A&  67.40$\pm$\scriptsize{49.1} &9.300$\pm$\scriptsize{0.1} &16.30$\pm$\scriptsize{9.9} &30.60$\pm$\scriptsize{26.4} &7.000$\pm$\scriptsize{1.9} \\
& PerSim-CQL-5&   \textbf{81.80}$\pm$\scriptsize{3.4} &\textbf{198.3}$\pm$\scriptsize{1.3} &\textbf{200.0}$\pm$\scriptsize{0.0} &\textbf{135.5}$\pm$\scriptsize{11.1} &\textbf{190.0}$\pm$\scriptsize{14.2}\\
\cmidrule[1.4pt]{1-7}

{Environment}& {Method} & {(0.3/1.7)} & {(1.7/0.3)}   & {(0.3/0.3)}  & {(1.7/1.7)}  & {(1.0/1.0)}    \\ \toprule
\multirow{3}{*}{\rotatebox[origin=c]{0}{{{HalfCheetah}} }} 
&BCQ-P& -1.460$\pm$\scriptsize{0.16} & -1.750$\pm$\scriptsize{0.22} & -1.690$\pm$\scriptsize{0.19} & -1.790$\pm$\scriptsize{0.21} & -1.720$\pm$\scriptsize{0.20} \\
& BCQ-A& -498.9$\pm$\scriptsize{108} & -113.3$\pm$\scriptsize{13.0} & -159.5$\pm$\scriptsize{51.7} & -35.73$\pm$\scriptsize{7.22} & -171.9$\pm$\scriptsize{41.5} \\
&PerSim-BCQ-5& ~\textbf{571.8}$\pm$\scriptsize{40.3} & ~\textbf{22.19}$\pm$\scriptsize{16.2} & -10.91$\pm$\scriptsize{47.4} & ~\textbf{64.30}$\pm$\scriptsize{2.60} & ~\textbf{157.0}$\pm$\scriptsize{31.6} \\
&CQL-P& -481.5$\pm$\scriptsize{25.9} &-442.2$\pm$\scriptsize{56.4} &-672.4$\pm$\scriptsize{17.6} &-254.5$\pm$\scriptsize{39.6} &-418.2$\pm$\scriptsize{23.4}  \\ 
&CQL-A& -0.700$\pm$\scriptsize{0.4} &-2.800$\pm$\scriptsize{0.8} &-0.600$\pm$\scriptsize{0.3} &-2.000$\pm$\scriptsize{0.6} &-5.500$\pm$\scriptsize{2.0} \\
&PerSim-CQL-5& \textbf{~1674~}$\pm$\scriptsize{135.2} &-38.50$\pm$\scriptsize{35.9} &-402.2$\pm$\scriptsize{113.0} &\textbf{~76.70}$\pm$\scriptsize{23.9} &\textbf{~94.20}$\pm$\scriptsize{102.8} \\
\cmidrule[1.4pt]{1-7}
\end{tabular}
\vspace{-3mm}
\end{table}

\subsection{Additional Experiments}

\noindent \textbf{Generalizing to Unseen Agents (Appendix \ref{appendix:unseen_agents}).} 
We show how to extend our method to unseen test agents, i.e., agents for which we have no training trajectory. 
Our extension crucially relies on having learned a good agent specific latent factor representation. 
Through a case-study for the MountainCar environment, we verify that using this extension, PerSim can successfully simulate unseen agents.

\noindent \textbf{Robustness to Data Scarcity (Appendix \ref{appendix:num_agents}).}  
We investigate how the number of training agents affects  PerSim's performance. 
We find that as we decrease training agents from $N=250$ to $N=25$, PerSim consistently achieves a higher reward than the other baselines across most agents. 
This ablation study directly addresses the robustness of PerSim to data scarcity, demonstrating that our principled framework is particularly suitable for the extreme data scarcity considered.

\vspace{-2mm}
\section{Conclusion}
\label{sec:conclusion}
\vspace{-2mm}
In this work, we investigate RL in an offline setting, where we observe a single trajectory across heterogeneous 
agents under an unknown, potentially sub-optimal policy. 
%
This is particularly challenging for existing approaches even in ``solved'' environments such as MountainCar and CartPole. 
%
PerSim offers a successful first attempt in simultaneously learning personalized policies across all agents under this data regime; we do so by first positing a principled low-rank latent factor representation, and then using it to build personalized simulators in a data-efficient manner.
%

%



\noindent \textbf{Limitations.}
Effectively leveraging offline datasets from heterogeneous sources (i.e., agents) for sequential decision-making problems will likely accelerate the adoption of RL.
However, there is much to be improved in PerSim. 
For example, in environments like HalfCheetah where the transition dynamics of each agent are harder to learn, the performance of PerSim is not comparable with the online setting, where one gets to arbitrarily sample trajectories for each agent. 
Of course, our considered data regime is fundamentally harder. 
%
Therefore, understanding the extent to which we can improve performance, using our low-rank latent factor approach or a different methodology altogether, remains to be established. 
Additionally, a rigorous statistical analysis for this setting, which studies the effect of the degree of agent heterogeneity, the diversity of the samples collected, etc. remains important future work.
%
We believe there are many fruitful inquiries under this challenging yet meaningful data regime for RL. 
Further, while PerSim is motivated by real-world problems, our empirical evaluation is limited to standard RL benchmarks where data collection and environment manipulation are feasible. 
%
Though PerSim's performance on standard RL benchmarks is encouraging, one cannot yet use PerSim as a out-of-the-box solution for critical real-world problems, without first designing a rigorous validation framework.
%

%

%


%

\section*{Acknowledgements and Funding Disclosure}
We would like to express our thanks to the authors of \cite{lee2020context} Kimin Lee, Younggyo Seo, Seunghyun Lee, Honglak Lee, and Jinwoo Shin for for their insightful comments and feedback.

{
\noindent This work was supported in parts by the MIT-IBM project on time series anomaly detection, NSF TRIPODS Phase II grant towards Foundations of Data Science Institute, KACST project on Towards Foundations of Reinforcement Learning, and scholarship from KACST (for Abdullah Alomar).}

\bibliography{example_paper}
\bibliographystyle{abbrv}




\newpage
\appendix

\begin{center}
\LARGE{\textbf{Supplementary Materials}}
\end{center}

\section{Organization of Supplementary Materials}
The supplementary materials consist of five main sections.

\noindent \textbf{Related Work.} 
In Appendix~\ref{appendix:related_work}, we give a detailed overview of the related literature.

\noindent \textbf{Proofs for Section \ref{sec:low_rank_representation}.} 
In Appendix~\ref{appendix:sec:proof}, we give the proofs of Theorem~\ref{thm:representation} and Proposition~\ref{proposition:example}.

\noindent \textbf{Algorithm and Implementation Details.} 
In Appendix~\ref{appendix:sec:implementation}, we provide further details about the implementation and training procedure for PerSim and the RL methods we benchmark against. 

\noindent \textbf{Detailed Experimental Setup.} 
In Appendix~\ref{appendix:sec:setup}, we detail the setup used to run our experiments.
In Appendix~\ref{appendix:sec:environment}, we describe the OpenAI environments used.
In Appendix \ref{appendix:sec:datasets}, we describe how we generate the offline training datasets for each environment.
%

\noindent \textbf{Additional Experimental Results.} 
In Appendix~\ref{appendix:sec:results}, we provide more details for the experiments we run. 
%
%
%
Specifically, in Appendix~\ref{appendix:sec:prediction_results}, we provide comprehensive results for the long-horizon prediction accuracy of model-based methods across all five test agents.
In Appendix~\ref{appendix:sec:average_reward}, we provide comprehensive results for the achieved reward in the various environments for both the model-based and model-free methods across all five test agents.
In Appendix~\ref{appendix:sec:latent_factors}, we present additional visualizations of the latent agent factors.
In Appendix \ref{appendix:bcq_persim}, we provide more details about the PerSim+BCQ/CQL experiments described in Section \ref{sec:BCQ_persim}.
In Appendix \ref{appendix:unseen_agents}, we describe and evaluate our proposed extension of PerSim to unseen test agents.
In Appendix \ref{appendix:num_agents}, we evaluate PerSim's robustness to  further data scarcity as we reduce the number of training agents.

\section{Related Work}\label{appendix:related_work}
{\bf Model-based Online RL.} 
In model-based RL methods~\citep{wang2019benchmarking,schrittwieser2020mastering,janner2019trust,wang2019exploring,kaiser2019model,luo2018algorithmic,deisenroth2011pilco}, the transition dynamics or simulator is learnt and subsequently utilized for policy learning.
%
Compared to their model-free counterparts, model-based approaches, when successful, have proven to be far more data-efficient in terms of the number of samples required to learn a good policy and have shown to generalize better to unseen (state, action) tuples~\citep{chua2018deep,clavera2018model,kurutach2018model,kaiser2019model,hafner2019learning}.
Recently, such methods have also been shown to effectively deal with agent heterogeneity, e.g., \cite{lee2020context} learns a context vector using the recent trajectory of a given agent, with a common context encoder across all agents.
Several recent works also utilize the meta-learning framework~\citep{finn2017model} to quickly adapt the model for model-based RL~\citep{saemundsson2018meta,nagabandi2018deep,nagabandi2018learning}.
Thus far, the vast majority of the model-based RL literature has focused on the online setting, where transition dynamics are learned by adaptively sampling trajectories.
%
Such online sampling helps these methods efficiently quantify and reduce uncertainty for unseen (state, action)-pairs.
Further, there has been some work showing the success of online model-based RL approaches with offline data, with minimal change in the algorithm \citep{janner2019trust,yu2020mopo}.
This serves as additional motivation to compare with a state-of-the-art model-based RL method such as \citep{lee2020context}, which is designed to address agent heterogeneity.

\noindent {\bf Model-free Offline RL.} 
As stated earlier, the offline RL paradigm~\citep{lange2012batch,levine2020offline} is meant to allow one to leverage large pre-recorded (static) datasets to learn policies.
Such methods are particularly pertinent for situations in which interacting with the environment can be costly and/or unethical, e.g., healthcare, autonomous driving, social/economic systems. 
The vast majority of offline RL methods are model-free~\cite{fujimoto2019off,kumar2019stabilizing,laroche2019safe,liu2020provably,wu2019behavior,agarwal2020optimistic,kumar2020conservative}.
%
Despite their rapidly increasing popularity, traditional offline RL methods suffer from ``distribution shift'', i.e., the policy learnt using such methods perform poorly on (state, action)-pairs that are unseen in the offline dataset~\citep{kumar2019stabilizing,fujimoto2019off,agarwal2019striving,levine2020offline}.
To overcome this challenge, offline RL methods design policies that are ``close'', in an appropriate sense, to the observed behavioural policy in the offline dataset~\citep{kumar2019stabilizing,wu2019behavior,fujimoto2019off}.
They normally do so by directly regularizing the learnt policy (e.g. parameterized via the Q-function) based on the quantified level of uncertainty for a given (state, action)-pair. 
Most offline RL methods tend to be designed for the case where there is no agent heterogeneity.
To study how much offline methods suffer if agent heterogeneity is introduced, we compare with one state-of-the-art offline RL method \citep{fujimoto2019off}.

\noindent {\bf Model-based Offline RL.}
Model-based offline RL is a relatively nascent field.
Two recent excellent works~\citep{kidambi2020morel,yu2020mopo} have shown that in certain settings, first building a model from offline data and then learning a policy outperforms state-of-the-art model-free offline RL methods on benchmark environments.
%
By learning a model of the transition dynamics first, it allows such methods to trade-off the risk of leaving the behavioral distribution with the gains of exploring diverse states.
%
However, the current inventory of model-based offline RL methods still require a large and diverse dataset for each agent of interest---in fact, these methods restrict attention to the setting where there is just one agent of interest and one gets observations just from that one agent. 
Our approach effectively resolves the challenge via developing a principled and generic model representation. 
It is worth mentioning that several recent theoretical works have shown that structured MDPs (e.g., low-rank or linear transition model or value functions) can lead to provably efficient RL algorithms~\citep{yang2019sample,agarwal2020flambe,jin2020provably,shah2020sample}, albeit in different settings.
Extending the current model-based offline RL methods to work with sparse data from heterogeneous agents, possibly by building upon the latent low-rank tensor representation we propose, remains interesting future work.

\section{Theoretical Results}\label{appendix:sec:proof}

\subsection{Proof of Theorem \ref{thm:representation}} 
\begin{proof}
We will construct the function $h_d$ by partitioning the latent parameter spaces associated with agents, states, and actions. 
We then complete the proof by showing that $h_d$ is entry-wise  close to $\tilde{f}_d$.

\noindent {\bf Partitioning the latent spaces to construct $h_d$.} 
Fix some $\delta_1, \delta_3 > 0$. 
Since the latent row parameters $\theta_n$ come from a compact space $[0,1]^{d_1}$, we can construct a finite covering or partitioning $P_1(\delta_1) \subset [0,1]^{d_1}$ such that for any $\theta_n \in [0,1]^{d_1}$, there exists a $\theta_{n'} \in P_1(\delta_1)$ satisfying $\|\theta_n - \theta_{n'} \|_2 \le \delta_1$. 
By the same argument, we can construct a partitioning $P_3(\delta_3) \subset [0,1]^{d_3}$ such that $\| \omega_a - \omega_{a'} \|_2 \le \delta_3$ for any $\omega_a \in [0,1]^{d_3}$ and some $\omega_{a'} \in P_3(\delta_3)$.

\noindent For each $\theta_n$, let $p_1(\theta_n)$ denote the unique element in $P_1(\delta_1)$ that is closest to $\theta_n$. 
Similarly, define $p_3(\omega_a)$ as the corresponding element in $P_3(\delta_3)$ that is closest to $\omega_a$. 
We enumerate the elements of $P_1(\delta_1)$ as $\{\tilde{\theta}_1, \dots, \tilde{\theta}_{|P_1(\delta_1)|}\}$.
Analogously, we enumerate the elements of $P_3(\delta_3)$ as $\{\tilde{\omega}_1, \dots, \tilde{\omega}_{|P_3(\delta_3)|}\}$.
We now define $h_d$ as follows: 
\begin{align}
    h_d(n, s, a) 
    &= 
    \sum^{|P_1(\delta_1)|}_{i = 1} 
    \sum^{|P_3(\delta_3)|}_{j = 1} 
    \mathbbm{1}(p_1(\theta_n) = \tilde{\theta}_i)
    \mathbbm{1}(p_3(\omega_a) = \tilde{\omega}_j)
    f_d(\tilde{\theta}_i, \rho_s, \tilde{\omega}_j). 
\end{align}

\noindent {\bf $\tilde{f}_d$ is well approximated by $h_d$.}
Here, we bound the maximum difference of any entry in $\tilde{f}_d$ from $h_d$. 
Using the Lipschitz property of $f_d$ (Assumption \ref{assumption:f_properties}), we obtain for any $(n,s,a)$, 
\begin{align}
	&|\tilde{f}_d(n, s, a) - h_d(n, s, a)| 
	\\ &= 
	\Big|
	f_d(\theta_n, \rho_s, \omega_a)
	- \sum^{|P_1(\delta_1)|}_{i = 1} 
	\sum^{|P_1(\delta_3)|}_{j = 1} 
	\mathbbm{1}(p_1(\theta_n) = \tilde{\theta}_i)
    \mathbbm{1}(p_3(\omega_a) = \tilde{\omega}_j)
    f_d(\tilde{\theta}_i, \rho_s, \tilde{\omega}_j)
    \Big|
	\\ &= \left| f_d(\theta_n, \rho_s, \omega_a) - f_d(p_1(\theta_n), \rho_s, p_3(\omega_a))\right|
	\\ &\le L  \left( 
	    \| \theta_n - p_1(\theta_n) \|_2 
	    + \| \omega_a - p_3(\omega_a) \|_2 
	    \right) 
	\\ &\le L  (\delta_1 + \delta_3 ). 
\end{align}
This proves that $\tilde{f}_d$ is entry-wise arbitrarily close to $h_d$.

\noindent {\bf Concluding the proof.} 
It remains to write $h_d(n, s, a)$ as
$
\sum^r_{\ell = 1}u_{\ell}(n) v_{\ell}(s, d) w_{\ell}(a)
$
and bound the induced $r$.
To that end, for $\ell = (i, j) \in [|P_1(\delta_1)|] \times [|P_3(\delta_3)|]$, we define 
\begin{align}
    u_\ell(n) &:= 
    \mathbbm{1}(p_1(\theta_n) = \tilde{\theta}_i), \quad
     v_\ell(s, d) := 
    f_d(\tilde{\theta}_i, \rho_s, \tilde{\omega}_j), \quad
     w_\ell(a) := 
    \mathbbm{1}(p_3(\omega_a) = \tilde{\omega}_j). 
\end{align}
This allows us to write $r = |P_1(\delta_1)| \cdot |P_3(\delta_3)|$. 
Since each of the latent spaces is a unit cube of 
different dimensions, i.e., $[0,1]^x$ with $x \in \{d_1, d_3\}$, 
we can simply create partitions $P_1(\delta_1), P_3(\delta_3)$
by creating grid of cubes of size $\delta_1$ and $\delta_3$
respectively. 
In doing so, the number of such cubes will scale as 
$|P_1(\delta_1)| \le C  \delta_1^{-d_1}, |P_3(\delta_3)| \le C  \delta_3^{-d_3}$, where $C$ is an absolute constant. 
As such, $r \le C \delta_1^{-d_1} \delta_3^{-d_3}$.
Setting $\delta = \delta_1 = \delta_3$ completes the proof.
\end{proof}

\subsection{Proof of Proposition  \ref{proposition:example}} 

\begin{proof}
To show that $r = 3$, it suffices to find $u(n) , v(s_n,1), v(s_n,2) , w(a_n) \in \Rb^3$ such that $h_d(n,s_n,a_n) = \sum_{\ell=1}^r u_\ell(n) v_\ell(s_n,d) w_\ell(a_n)$ for any $n \in [N]$, $s_n = [s_{n1},s_{n2}] \in \Sc$, $a_n \in \Ac$, and $d \in \{1, 2\}$. 
In particular, we require that $u(n)$ can only depend on agent $n$, i.e., not on the action or state. 
Analogously, 
%
$v(s_n,1)$ and $v(s_n,2)$ can only depend on the state, and $w(a_n)$ can only depend on the action. 
Towards this, consider the following factors: 
\begin{align*}
&u(n) = \begin{bmatrix}
 {1} & {g_n} & {1}
\end{bmatrix}, \qquad  
w(a_n) = \begin{bmatrix}
{1} & {1} & {a_n}
\end{bmatrix}, 
\\ 
&v(s_n, 1) =  \begin{bmatrix}
{ s_{n1} + s_{n2}} &   {-\frac{\cos(3s_{n1})}{2}}  & {\frac{1}{2}}  \\
\end{bmatrix},  \qquad 
v(s_n, 2) =  \begin{bmatrix}
{s_{n2}}  & {-\cos(3s_{n1})} & {1}
\end{bmatrix}. 
\end{align*}
Recalling $ h_1(n, s_n, a_n) = { 
\  s_{n1} + s_{n2} } { -\frac{g_n\cos(3s_{n1})}{2}} { +\frac{a_n}{2} } $ and $h_2(n,  s_n, a_n) = s_{n2}   {- g_n\cos(3s_{n1})} + { a_n}$ completes the proof.  
\end{proof}

\section{Algorithm and Implementation  Details}\label{appendix:sec:implementation}

\subsection{PerSim} \label{appendix:sec:implementation_our}

\textbf{Step 1 Details: Learning Personalized Simulators.}
As explained in Section \ref{sec:algorithm}, the personalized simulators are effectively trained by learning $g_u$,  $g_v$, and $g_w$, which correspond to the agent, state, and action encoders, respectively. 
Below, we detail the architecture used for each function.
\begin{enumerate}
    \item \textbf{Agent encoder: $g_u$.} We use a single layer that takes in a one-hot encoder of the agent and returns an $r$-dimensional latent factor.
    \item \textbf{State encoder: $g_v$.}  We use a multilayer perceptron (MLP) with 1 hidden layer of 256 ReLU activated nodes for both MountainCar and CartPole, and an MLP with 4 hidden layers each with 256 ReLU activated nodes for HalfCheetah.
    \item \textbf{Action encoder: $g_w$}. In environments with discrete action spaces, i.e., MountainCar and CartPole, we  use a single layer that takes in a one-hot encoder of the action and produce an $r$-dimensional latent factor.  For HalfCheetah, we  use an MLP with 2 hidden layers of 256 ReLU activated nodes. 
\end{enumerate}
We choose the tensor rank $r$ to be $3$, $5$, and $15$ for the MountainCar, CartPole, and HalfCheetah environments, respectively. 
The choice is made via cross validation from the set $\{3,5,10,15,20,30\}$. 
Specifically, $20\%$ of the data points (selected randomly from different trajectories) are set aside for validation in the hyper-parameter selection process. We train our simulators with a learning rate of $0.001$, $300$ epochs, and a batch size of $512$ for HalfCheetah and MountainCar and $64$ for CartPole. 
Please refer to the pseudo-code in Algorithm \ref{alg:nn} for a detailed description of the training procedure.

\begin{algorithm}[h]
\caption{Training Personalized Dynamic Models}\label{alg:nn}
\begin{algorithmic}[1]
\State {\bf Input:} Dataset $\Dc$, Rank $r$, Learning rate $\eta$, Batch size $B$, Number of epochs $K$ 
\State {\bf Output:}  $g_u(\cdot;\psi)$,  $g_v(\cdot;\phi)$,  $g_w(\cdot;\theta)$
\State Initialize $\psi$, $\phi$, and $\theta$
%
\For{each epoch} :
\For{each batch} :
\For{$i = 1$ to B} :
\State Sample $ \{s_{i}, a_{i}, s'_{i}, n_i\} \sim \Dc$
\State Compute $ \Delta s_{i} \gets  s'_{i} - s_{i}$
\State Get agent latent factor $\widehat{u}(n_i) \gets g_u(n_i;\psi) $
\State Get state latent factor $\widehat{v}(s_i) \coloneqq [\widehat{v}(s_i, d)]_{d \in [D]} \gets g_v(s_{i};\phi) $
\State Get action latent factor $\widehat{w}(a_i) \gets g_w(a_{i};\theta) $
\State Get the error estimate  
$\mathcal{L}_i \gets \norm{\Delta s_{i} -\sum_{\ell=1}^r \widehat{u}_{\ell}(n_i) v_{\ell}(s_i) \widehat{w}_{\ell}(a_i) }_2^2$
%
\EndFor
\State Update $\psi \gets \psi - \eta \nabla_\psi \frac{1}{B}\sum_{i=1}^B  \mathcal{L}_i $
\State Update $\phi \gets \phi - \eta \nabla_\phi \frac{1}{B}\sum_{i=1}^B  \mathcal{L}_i $
\State Update $\theta \gets \theta - \eta \nabla_\theta \frac{1}{B}\sum_{i=1}^B  \mathcal{L}_i $
\EndFor
\EndFor
\end{algorithmic}
\end{algorithm}


\noindent \textbf{Step 2 Details: Learning a Decision-making Policy.} 
As outlined in Section \ref{sec:algorithm}, we use MPC to select the best action.
Specifically, we sample $C$ candidate action sequences of length $h$, which we denote as $\{a^{(i)}_1, \dots,  a^{(i)}_h\}_{i = 1}^C$.
The actions are sampled using  cross entropy in environments with continuous action spaces and random shooting  in environments with discrete action space \cite{camacho2013model, botev2013cross}.

\noindent 
Since the offline data may not span the entire state-action space, planning using a learned simulator without any regularization may result in ``model exploitation'' \cite{levine2020offline}.
%
To overcome this issue, we gauge the model uncertainty, as is common in the literature, as follows.
We train an ensemble of $M$ simulators $\{g_u^{(m)}, g_s^{(m)}, g_a^{(m)} \}_{m =1}^M$. 
Then, for $i \in [C]$, we evaluate the average reward of performing the $i$-th action sequence, which we denote by $r^{(i)}$,  using the estimates across the $M$ simulators. Specifically,
\begin{equation*}
r^{(i)} = \frac{1}{M} \sum_{m =1}^M \sum_{t =1}^h  R\Big(\widehat{s}^{(m,i)}_t,a_t^{(i)}\Big),
\end{equation*}
where $\widehat{s}^{(m,i)}_t$ is the predicted trajectory according to the $m$-th simulator and the sequence of actions $\{a^{(i)}_1, \dots,  a^{(i)}_h\}$,
and $R$ is the reward function (which we assume is known, as is done in prior works \cite{lee2020context, kidambi2020morel}). 
Finally, we choose the first element from the sequence of actions with the best \textit{average} reward, i.e., the sequence $\{a^{(i^*)}_1, \dots,  a^{(i^*)}_h\}$, where
$
i^* = \arg\max_{i\in[C]} ~ r^{(i)}.
$

\noindent For MountainCar and CartPole, we use random shooting to sample 1000 candidate actions with a planning horizon of 50.
For HalfCheetah, we use the cross entropy method to sample 200 candidate actions with a planning horizon of 30. 
For all environments, we train $M=5$ simulators.

\subsection{Benchmarking Algorithms} \label{appendix:sec:implementation_bench}

\noindent \textbf{Vanilla CaDM + PE-TS CaDM.} We use the implementation provided by the authors in \cite{lee2020context}.\footnote{https://github.com/younggyoseo/CaDM}
To train on offline data, we modify the sampling procedure in the implementation. 
Specifically, we change it to sample from a replay buffer containing the recorded trajectories. 
Similar to our method, we use MPC with a planning horizon of 30 for  HalfCheetah, and 50 for MountainCar and CartPole. 
We train the forward dynamic model, the backward dynamic model, and the context encoder for 20 iterations each with a maximum of 200 epochs and a learning rate of 0.001.
For PE-TS, as is done in \cite{lee2020context}, we use an ensemble of five dynamics models, and use twenty particles for trajectory sampling.

\noindent \textbf{BCQ-P +BCQ-A.}
We use the implementation provided by the authors in \cite{fujimoto2019off}.\footnote{https://github.com/sfujim/BCQ}
Specifically, we use discrete BCQ for MountainCar and CartPole, and continuous BCQ for HalfCheetah.  
For both BCQ-P and BCQ-A, we train the policy for $5.5 \times 10^5$ iterations.      

\colorred{
\noindent \textbf{CQL-P +CQL-A.}
We use the CQL implementation provided by the d3rlpy library \cite{seno2020d3rlpy}.\footnote{https://github.com/takuseno/d3rlpy}
Specifically, we use discrete CQL for MountainCar and CartPole, and continuous CQL for HalfCheetah.  
For discrete CQL, we set the number of critics to 3, the parameter $\alpha$ to 1, and we use a batch size of 512 and a learning rate of $10^{-4}$.
For continuous CQL, we set the Lagrange threshold to $5$, the  policy learning rate to  $3e-5$, and the critic learning rate to $3e-4$. We choose these parameters according to the guidelines  the authors provide in \cite{kumar2020conservative}.

\noindent \textbf{MOReL-P +MOReL-A.}
We use the available open source implementation of MOReL \cite{fujimoto2019off}.\footnote{https://github.com/SwapnilPande/MOReL}
We further extend the implementation to accommodate for environments with discrete actions. 
We use uncertainty penalty of -50 in MountainCar and CartPole and a penalty of -200 in HalfCheetah. 
For the dynamic model, we use an MLP with two hidden layers.
Each layer has  128 neurons in MountainCar and CartPole, and  1024 neurons in HalfCheetah as done in the original paper. 

}

\section{Detailed Setup }\label{appendix:sec:setup} 

\subsection{Environments} \label{appendix:sec:environment}

\begin{table}[htb]
\caption{Environment parameters used for experiments.}
\label{table:env}
\tabcolsep=0.06cm
\fontsize{8.0pt}{9.0pt}\selectfont
\centering
\begin{tabular}{@{}p{1.6cm}p{4.0cm}p{4.0cm}p{4.0cm}@{}}
\toprule
Environment  & Parameter range                                                                    & Test agents    & Policy training agents      \\ \midrule
MountainCar  & gravity $\in$ [0.0001, 0.0035]                                                         & 
\begin{tabular}[c]{@{}l@{}} \{0.0001, 0.0005, 0.0010,\\  0.0025, 0.0035\} \end{tabular}   &  \begin{tabular}[c]{@{}l@{}} \{0.0003, 0.00075, 0.00175,\\  0.0025, 0.0030\} \end{tabular} \\\midrule
CartPole     & \begin{tabular}[c]{@{}l@{}}length $\in$[0.15, 0.85]\\ force $\in$ [2.0, 18]\end{tabular} & \begin{tabular}[c]{@{}l@{}}\{(2.0,0.5), (10.0,0.5), (18.0, 0.5), \\  (10.0,0.85), (10.0,0.15)\}\end{tabular} & \begin{tabular}[c]{@{}l@{}}\{(6.0,0.5), (14.0,0.5), (10.0,0.5), \\   (10.0,0.675), (10.0, 0.325)\}\end{tabular} \\ \midrule
HalfCheetah  & \begin{tabular}[c]{@{}l@{}} relative mass $\in$  [0.2,1.8]\\ relative damping $\in$ [0.2,1.8]\end{tabular}   & \begin{tabular}[c]{@{}l@{}}\{(0.3,1.7), (1.7,0.3), (0.3, 0.3),\\   (1.7,1.7), (1.0,1.0)\}\end{tabular}    & \begin{tabular}[c]{@{}l@{}}\{(0.6,1.4), (1.4,0.6), (0.6, 0.6), \\  (1.4,1.4)\}\end{tabular}  \\\bottomrule 
\end{tabular}
\end{table}

%
\noindent We perform experiments on three environments from the OpenAI Gym: two classical non-linear control environments, MountainCar and CartPole, and one Mujoco environment, HalfCheetah  \citep{todorov2012mujoco}. 
Next, we describe these three environments in detail.

\noindent \textbf{MountainCar.} In MountainCar, the goal  is to drive a under-powered car to the top of a hill by taking the least number of steps.
\begin{itemize}
    \item \emph{Observation.} We observe $x(t)$, $\Dot{x}(t)$: the position and velocity of the car, respectively.   
    \item \emph{Actions.} There are three possible actions $\{0,1,2\}$:   (0) accelerate to the left; (1) do nothing; (2) accelerate to the right.
    \item {\em Reward}. The reward is defined as 
    $$
    R(t) = \begin{cases}
    1, &x(t)\ge0.5 \\
    -1, &\text{otherwise}.
    \end{cases}
    $$
    \item {\em Environment modification.} We vary the gravity within the range $[0.0001,0.0035]$.  Note that with a weaker gravity, the environment is trivially solved by directly moving to the right. On the other hand, with a stronger gravity, the car must  drive left and right to build up enough momentum. See Table \ref{table:env} for details about the parameter ranges and the test agents. 
\end{itemize}

\vspace{2mm}

\noindent \textbf{CartPole.} 
In CartPole, a pole is attached to a cart moving on a frictionless track. The goal is to prevent the pole from falling over by moving the cart to the left or to the right, and to do so for as long as possible (maximum of 200 steps). 
\begin{itemize}
    \item {\em Observation.} We observe $x(t), \Dot{x}(t), \theta(t), \Dot{\theta}(t)$: the cart's position, its velocity, the pole's angle, and its angular velocity, respectively. 
    \item {\em Actions.} There are two possible actions $\{0,1\}$: (0) push to the right; (1) push to the left.
    \item {\em Reward}.  The reward is 1 for every step taken without termination. The  environment terminates when the pole angle exceeds 12 degrees or when the cart position exceeds 2.4.
    \item {\em Environment modification.} As in \cite{lee2020context}, we vary the length of the pole and push force within the ranges $[0.15,0.85]$ and $[2.0,18.0]$, respectively. See Table \ref{table:env} for details about the parameter ranges and the test agents. 
\end{itemize}

\vspace{2mm}

\noindent \textbf{HalfCheetah.} In HalfCheetah, the goal is to move the cheetah as fast as possible. The cheetah's body consists of 7 links connected via 6 joints.
\begin{itemize}
    \item {\em Observation.} We observe an 18-dimensional vector that includes the angle and angular velocity of all six joints, as well as the 3-D position and orientation of the torso.  Additionally, as is done in previous studies \cite{lee2020context, kidambi2020morel}, we append the center of mass velocity to our state vector to enable computing the reward
from observations. 
    \item {\em Actions.} The action $a(t) \in [-1,1]^6$ represents the torque applied at the six joints. 
    \item {\em Reward}.  The reward is defined as
    $$
    R(t) = v(t) - 0.05 \norm{a(t)}^2,
    $$
    where $v(t)$ is the center of mass velocity at time $t$.
    \item {\em Environment modification.} As in \cite{lee2020context}, we scale the mass of every link and the damping of every joint by factors $m$ and $d$, respectively. Specifically, we vary both $m$ and $d$ within the range $[0.2,1.8]$. See Table \ref{table:env} for details about the parameter ranges and the test agents. 
\end{itemize}

\subsection{Offline Datasets} \label{appendix:sec:datasets}


As stated in Section \ref{sec:experiments}, we generate four offline datasets for each environment with varying  ``optimality'' of the sampling policy. Specifically, we generate 500 trajectories (one per agent) for each environment as per the following sampling procedures:  

\noindent \textbf{(i) Pure}. 
In the Pure procedure, actions are sampled according to a fixed policy (for each agent) that has been trained to achieve reasonably good performance. 
Specifically, for each environment, we first train a policy using online model-free algorithms for the training agents shown in Table~\ref{table:env}.
Specifically, we train these logging policies using DQN~\citep{mnih2015human} for MountainCar and CartPole, and using TD3  for and HalfCheetah.
We train these policies to achieve rewards of approximately -200, 120, 3000, for MountainCar, CartPole, and HalfCheetah respectively.  
Then, to sample a trajectory for each of the 500 agents, we use the policy trained on the training agent with the closest parameter value. 
%

\noindent \textbf{(ii) Random}. Actions are selected uniformly at random.

\noindent \textbf{(iii) Pure-$\varepsilon$-20}.
Actions are selected uniformly at random with probability of 0.2, and selected via the pure policy otherwise. 

\noindent \textbf{(iv) Pure-$\varepsilon$-40}.
Actions are selected uniformly at random with probability of 0.4, and selected via the pure policy otherwise. 

\noindent See Table~\ref{table:datasets_details} for details about the reward observed for the five test agents using these four sampling procedures, and the average reward and trajectory length achieved across all 500 agents.

\begin{table}[ht]
\centering
\caption{Observed reward and trajectory length in the four sampled datasets in each environment. Agent 1 to Agent 5 refer to the five test agents. The average is taken across all 500 agents.}
\vspace{2mm}
\label{table:datasets_details}
\tabcolsep=0.15cm
\fontsize{8.5pt}{9.0pt}\selectfont
\begin{tabular}{@{}p{1.5cm}p{1.5cm}p{1.2cm}p{1.2cm}p{1.2cm}p{1.2cm}p{1.2cm}p{1.2cm}p{2.2cm}@{}}
\toprule
\multirow{2}{*}{Environment}                     & \multirow{2}{*}{Data} & \multicolumn{6}{c}{Observed Reward}                  & {Average }                \\ \cmidrule(lr){3-8} 
                                                 &                       & Agent 1 & Agent 2 & Agent 3 & Agent 4 & Agent 5 & Average & Trajectory Length \\ \midrule 
{\multirow{4}{*}{MountainCar}} 
& Pure                  &    -48.0& -50.0& -57.0& -171.0& -134.0& -112.926    & 113.914          \\
   & Random  & -500.0& -500.0& -500.0& -500.0& -500.0& -496.324& 496.344   \\
    & Pure-eps-2  &-46.0& -54.0& -73.0& -165.0& -140.0& -155.252& 156.214  \\
  & Pure-eps-4 &   -55.0& -64.0& -500.0& -264.0& -208.0& -227.278& 228.18 \\\midrule
\multirow{4}{*}{CartPole}                        
& Pure                  &197& 200& 193& 179& 200& 185.14& 186.14           \\
 & Random &         59& 23& 10& 26& 17& 22.39& 23.39       \\
 & Pure-eps-2 &     180& 199& 30& 179& 199& 157.92& 158.92 \\
 & Pure-eps-4 &     38& 23& 11& 170& 14& 92.80& 93.80       \\ \midrule
\multirow{4}{*}{HalfCheetah}                     
& Pure  &522.40& 1246.32& 251.25& 2646.85& 1011.99& 1894.10& 1000.00       \\
 & Random & -395.58& -65.77& -106.80& -150.01& -323.86& -251.75& 1000.00   \\
 & Pure-eps-2           &399.95& 938.58& 189.17& 1742.85& 1137.45& 1121.52& 1000.00   \\
 & Pure-eps-4            &   508.11& -115.94& 128.64& 1155.23& 216.69& 771.71& 1000.00   \\  \bottomrule
\end{tabular}
\end{table}

\newpage

\section{Additional Experimental Results}
\label{appendix:sec:results}


\subsection{Detailed Prediction Error Results} \label{appendix:sec:prediction_results}

\begin{figure}[h!tb]
\begin{subfigure}[t]{\textwidth}
        \centering
    \includegraphics[width = 0.70 \linewidth]{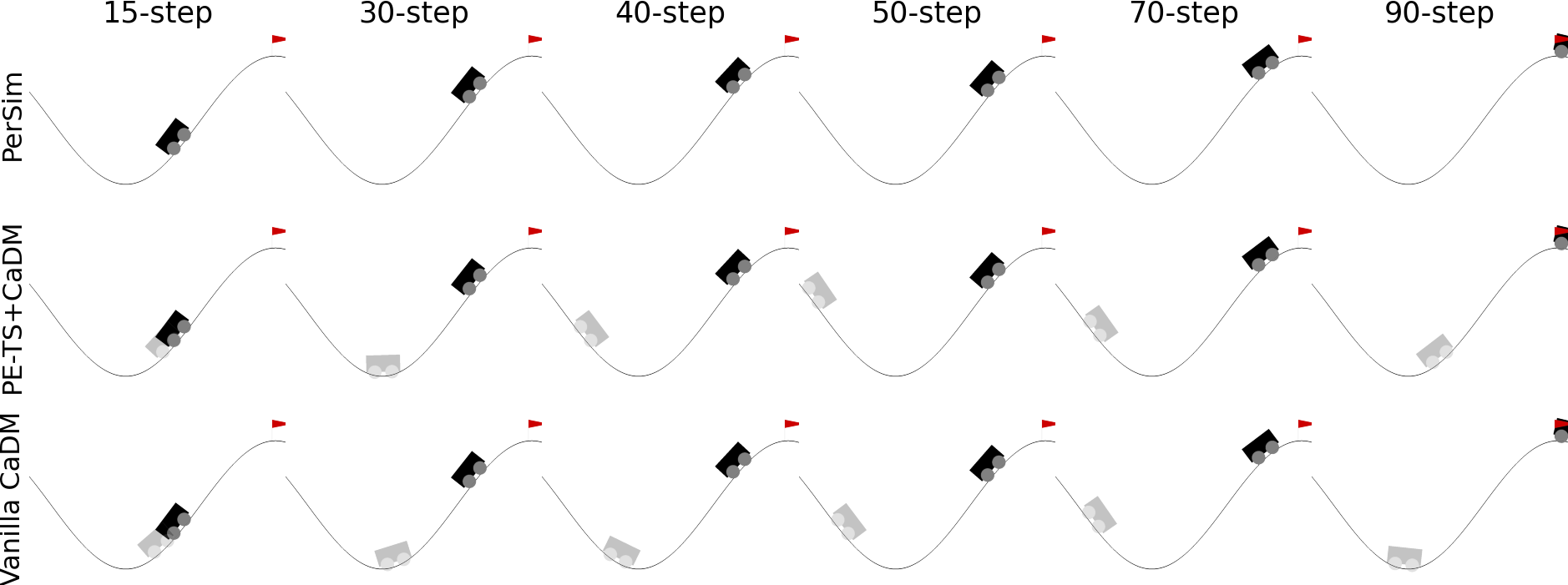}
    \caption{MountainCar with gravity 0.0001}
    \label{fig:MC_vis_1}
\end{subfigure}
\begin{subfigure}[t]{\textwidth}
        \centering
    \includegraphics[width =0.70 \linewidth]{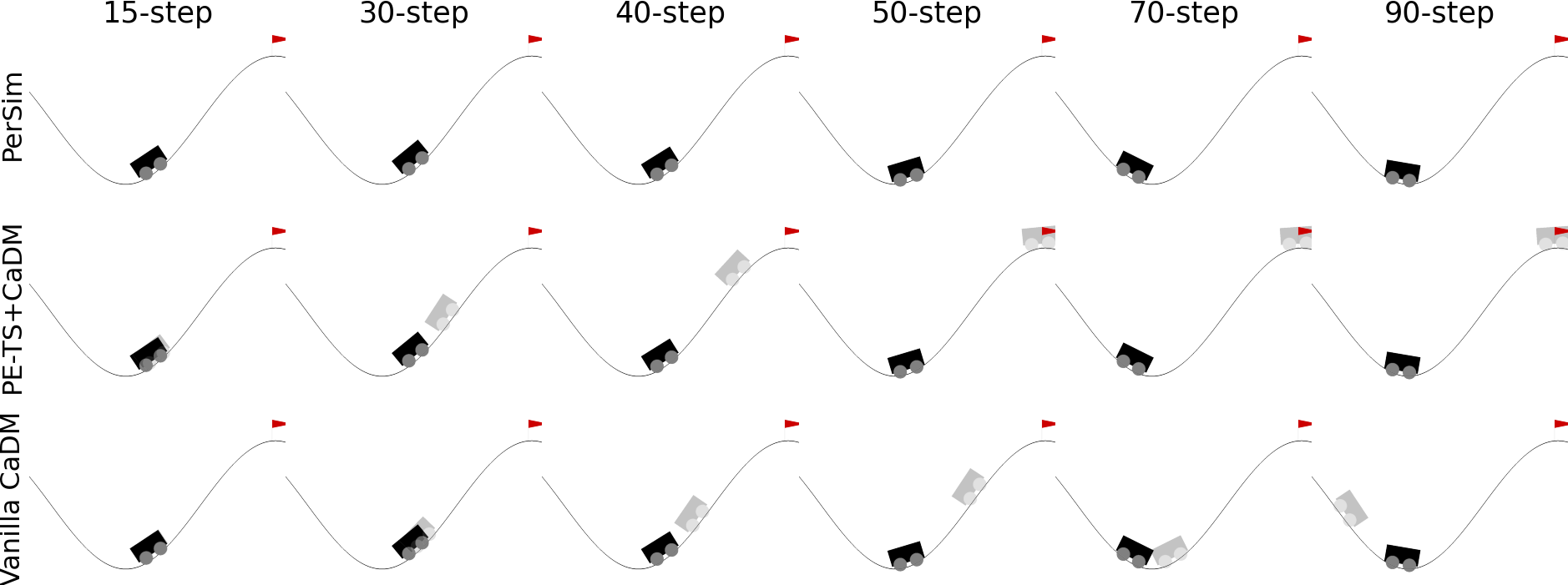}
    \caption{MountainCar with gravity 0.0025}
    \label{fig:MC_vis_2}
\end{subfigure}
\begin{subfigure}[t]{\textwidth}
        \centering
    \includegraphics[width =0.70 \linewidth]{figures/cartPole_unit3_ps.png}
    \caption{CartPole with pole length 0.85}
    \label{fig:CP_vis_1}
\end{subfigure}
\begin{subfigure}[t]{\textwidth}
        \centering
    \includegraphics[width =0.70 \linewidth]{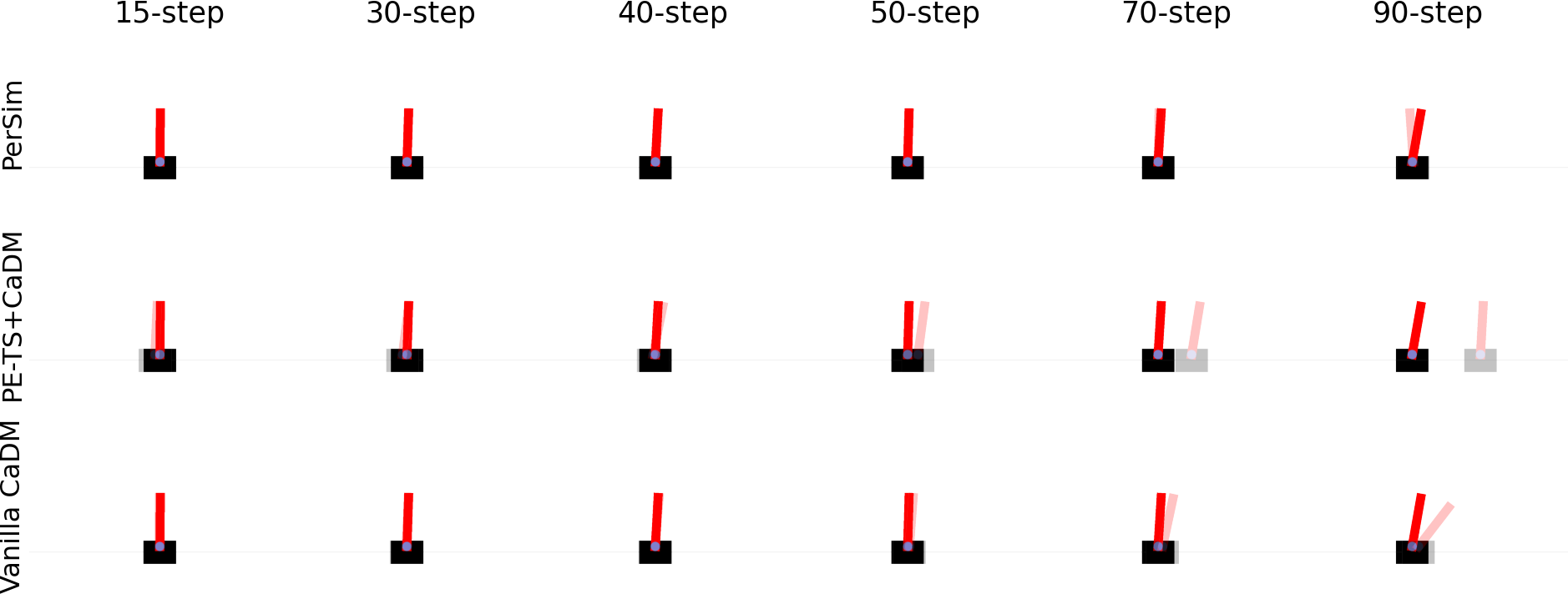}
    \caption{CartPole with pole length 0.5}
    \label{fig:CP_vis_2}
\end{subfigure}
\caption{{Visualization of the prediction accuracy of PerSim for two heterogeneous agents in MountainCar and CartPole, and how it compares with the two CaDM variants. Specifically, given an initial state and a sequence of actions, we predict future states for the next 90 steps. Ground-truth states and predicted states are denoted by the opaque and translucent objects, respectively.}}
    \label{fig:vis_appendix}
\end{figure}

In this section, we provide additional results for the prediction error experiments.
As detailed in Section \ref{sec:experiments}, we evaluate the accuracy of the learned transition dynamics for each of the five test agent, focusing on long-horizon model prediction.
Specifically, we predict the next 50-step state trajectory given an initial state and an unseen sequence of 50 actions.
The sequence of 50 actions are chosen according to an unseen test policy.
Precisely , the test  policies are fitted via DQN for MountainCar and CartPole, and via TD3 for and HalfCheetah, for the agent with the default covariate parameters. 
The test policies were trained to achieve an average rewards of -150, 150, 4000 for the MountainCar, CartPole, and HalfCheetah environments, respectively.
As described in Section \ref{sec:experiments}, we report the mean RMSE and the median $R^2$ across 200 trials. 
The results are summarized in Tables \ref{table:MC_pred_full}, \ref{table:CP_pred_full}, and \ref{table:HC_pred_full} for MountainCar, CartPole, and HalfCheetah, respectively. 
Additionally, Figure \ref{fig:vis_appendix} visualizes the prediction accuracy of PerSim up to 90-steps ahead predictions for two test agents in MountainCar and CartPole.

\begin{table}[h]
\caption{Prediction error: MountainCar}
\label{table:MC_pred_full}
\tabcolsep=0.15cm
\fontsize{8.5pt}{9.0pt}\selectfont
\centering
\begin{tabular}{@{}p{1.0cm}p{1.9cm}p{1.9cm}p{1.9cm}p{1.9cm}p{1.9cm}p{1.9cm}@{}}
\toprule
\multirow{2}{*}{Data}    & \multirow{2}{*}{Method}       & {Agent 1}              & {Agent 2}      & {{Agent 3}} & {Agent 4}               & {Agent 5}               \\
{}                     &        & {0.0001 }              & {0.0005}               & {{0.001 }} & {0.0025}               & {0.0035}               \\ \midrule
\multirow{3}{*}{Pure}      & PerSim   &  {0.025 (0.969)} &  {0.090 (0.973)} &  {0.031 (0.978)} &  {0.014 (0.942)} &  {0.039 (0.803)}  \\
                           & Vanilla CaDM & 0.149 (0.741)&0.126 (0.767)&0.075 (0.913)&0.177 (-18.426)&0.238 (-30.148)         \\
                           & PE-TS CaDM   & 0.326 (-1.912)&0.288 (-3.449)&0.194 (-2.61)&0.154 (-1.114)&0.148 (-0.179)         \\\midrule
\multirow{3}{*}{Random}    & PerSim    & {0.004 (0.999)} &  {0.003 (1.000)} &  {0.001 (1.000)} &  {0.001 (1.000)} &  {0.001 (1.000) }\\
                           & Vanilla CaDM &  0.256 (0.428)&0.203 (0.264)&0.162 (-1.041)&0.134 (0.710)&0.217 (-1.767)        \\
                           & PE-TS CaDM   & 0.242 (0.310)&0.177 (0.725)&0.156 (-0.259)&0.101 (0.868)&0.075 (0.967)        \\\midrule
\multirow{3}{*}{Pure-$\varepsilon$-20} & PerSim     &  {0.004 (1.000)} &  {0.003 (1.000)} &  {0.001 (1.000)} &  {0.002 (0.999)} &  {0.004 (0.998)}\\
                           & Vanilla CaDM &  0.227 (0.309)&0.201 (0.271)&0.131 (0.405)&0.101 (-0.369)&0.157 (-2.252)       \\
                           & PE-TS CaDM   &  0.35 (-2.571)&0.282 (-3.829)&0.216 (-1.706)&0.139 (0.746)&0.13 (0.892)     \\\midrule
\multirow{3}{*}{Pure-$\varepsilon$-40} & PerSim   &  {0.006 (0.999)} &  {0.005 (1.000)} &  {0.004 (1.000)} &  {0.006 (0.992)} &  {0.003 (0.999)} \\
                           & Vanilla CaDM & 0.199 (0.54)&0.176 (0.542)&0.106 (0.703)&0.119 (0.384)&0.187 (-9.005)       \\
                           & PE-TS CaDM   & 0.639 (-2.273)&0.283 (-2.299)&0.174 (-0.631)&0.192 (-6.056)&0.157 (0.546)   \\ \cmidrule[1.5pt]{1-7}
\end{tabular}
\end{table}

\begin{table}[h]
\caption{Prediction Error: CartPole}
\label{table:CP_pred_full}
\tabcolsep=0.15cm
\fontsize{8.5pt}{9.0pt}\selectfont
\centering
\begin{tabular}{@{}p{1.0cm}p{1.9cm}p{1.9cm}p{1.9cm}p{1.9cm}p{1.9cm}p{1.9cm}@{}}
\toprule
\multirow{2}{*}{Data}    & \multirow{2}{*}{Method}       & {Agent 1}              & {Agent 2}      & {{Agent 3}} & {Agent 4}               & {Agent 5}               \\
{}& {}& {(2/0.5) } & {(10.0/0.5) } & {(18.0/0.5)} & {(10/0.85)}   & {(10/0.15)}    \\ \toprule
\multirow{3}{*}{\rotatebox[origin=c]{0}{{{Pure}} }} 
&PerSim   &  {0.001 (1.000)} &  {0.001 (1.000)} &  {0.002 (1.000)} &  {0.001 (1.000)} &   {0.035 (0.995)} \\
&Vanilla CaDM& 0.403 (0.712)&0.152 (0.425)&0.148 (0.664)&0.039 (0.975)&2.531 (-0.532) \\
&PE-TS CaDM&0.031 (0.993)&0.011 (0.995)&0.016 (0.997)&0.006 (1.000)&0.319 (0.651) \\
\midrule
\multirow{3}{*}{\rotatebox[origin=c]{0}{{{Random}} }} 
&PerSim&    {0.014 (0.982)} &  {0.022 (0.970)} &  {0.030 (0.979)} &  {0.006 (0.999)} &  {0.152 (0.883)} \\
&Vanilla CaDM&0.172 (0.095)&0.282 (-0.483)&0.514 (-0.381)&0.098 (0.639)&3.307 (-0.734) \\
&PE-TS CaDM&0.564 (-1.357)&0.493 (-0.200)&0.891 (-0.458)&0.216 (0.450)&3.764 (-1.104) \\ \midrule
\multirow{3}{*}{\rotatebox[origin=c]{0}{{{Pure-$\varepsilon$-20}} }} 
&PerSim &  {0.000 (1.000)} &  {0.001 (1.000)} &  {0.008 (0.999)} &  {0.001 (1.000)} &  {0.048 (0.984)} \\
&Vanilla CaDM&0.270 (-0.982)&0.037 (0.973)&0.046 (0.991)&0.058 (0.943)&2.193 (-0.432) \\
&PE-TS CaDM&0.639 (-1.206)&0.252 (0.000)&0.434 (-0.170)&0.148 (0.384)&2.680 (-0.561) \\\midrule
\multirow{3}{*}{\rotatebox[origin=c]{0}{{{Pure-$\varepsilon$-40}} }} 
&PerSim &   {0.000 (1.000)} &  {0.002 (1.000)} &  {0.011 (0.998)} &  {0.000 (1.000)} &   {0.018 (0.998)} \\
&Vanilla CaDM&0.233 (-0.883)&0.055 (0.956)&0.032 (0.993)&0.035 (0.987)&3.286 (-0.618) \\
&PE-TS CaDM&0.051 (0.983)&0.017 (0.992)&0.013 (0.998)&0.010 (0.999)&0.411 (0.553) \\
\cmidrule[1.5pt]{1-7}
\end{tabular}
\end{table}

\begin{table}[h]
\caption{Prediction Error: HalfCheetah}
\label{table:HC_pred_full}
\tabcolsep=0.15cm
\fontsize{8.5pt}{9.0pt}\selectfont
\centering
\begin{tabular}{@{}p{1.0cm}p{1.9cm}p{1.9cm}p{1.9cm}p{1.9cm}p{1.9cm}p{1.9cm}@{}}
\toprule
\multirow{2}{*}{Data}    & \multirow{2}{*}{Method}       & {Agent 1}              & {Agent 2}      & {{Agent 3}} & {Agent 4}               & {Agent 5}               \\
& &    {(0.3/1.7)} & {(1.7/0.3)}   & {(0.3/0.3)}  & {(1.7/1.7)}  & {(1.0/1.0)} 
\\ \toprule
\multirow{3}{*}{\rotatebox[origin=c]{0}{{{Pure}} }} 
&PerSim&  {1.401 (0.890)} &  4.766 {(0.574)} &  {4.385 (0.791)}  &  {1.409 (0.840)}   &  {2.505 (0.793)}  \\
&Vanilla CaDM& 3.902 (0.472) & 3.851(0.490) & 6.308 (0.380) & 2.881 (0.336) & {1.804 (0.829)} \\
&PE-TS CaDM& 3.147 (0.614) & {3.080 (0.682)} & 5.270 (0.572) & 1.833 (0.647) & {1.915 (0.784)}\\
\midrule
\multirow{3}{*}{\rotatebox[origin=c]{0}{{{Random}} }} 
&PerSim &  {1.194 (0.916)} &  4.064 (0.670) & {4.070 (0.812)} &  {1.291 (0.855)} &  2.325 (0.812) \\
&Vanilla CaDM& 4.030 (0.435) & 4.121(0.430) & 4.446 (0.672) & 3.840 (-0.432) & 4.270 (0.255) \\
&PE-TS CaDM& 2.735 (0.731) & {2.756 (0.688)} & 4.141 (0.767) & 2.031 (0.601) & {1.957 (0.773)}\\ \midrule
\multirow{3}{*}{\rotatebox[origin=c]{0}{{{Pure-$\varepsilon$-20}} }} 
&PerSim&    {1.172  (0.922)} &  4.283 {(0.660)} &  {3.832 (0.844)} &  {1.123 (0.880)} &  {2.057 (0.840)} \\
&Vanilla CaDM&3.613 (0.534) & 3.455 (0.538) & 6.046 (0.477) & 2.256 (0.575) & 2.070 (0.753) \\
&PE-TS CaDM&2.913 (0.699) & {2.959} (0.652) & 4.818 (0.647) & 1.970 (0.633) & 2.073 (0.744) \\ \midrule
\multirow{3}{*}{\rotatebox[origin=c]{0}{{{Pure-$\varepsilon$-40}} }} 
&PerSim &  {1.016 (0.940)} &  4.021 {(0.664)} &  {3.742 (0.853)} &  {1.287 (0.868)} & {1.887 (0.836)} \\
&Vanilla CaDM& 3.685 (0.493) & 3.612 (0.480) & 6.000 (0.392) & 2.561 (0.521) & 2.136 (0.738) \\
&PE-TS CaDM&3.021 (0.672) & {3.025} (0.614) & 5.075 (0.615) & 1.792 (0.668) & 1.930 (0.779) \\
\cmidrule[1.5pt]{1-7}
\end{tabular}
\vspace{-3mm}
\end{table}

\FloatBarrier
\subsection{Detailed Average Reward Results} \label{appendix:sec:average_reward}
In this section, we show the full results for the experiments for the reward achieved in each environment. 
Specifically, we report the average reward achieved by PerSim and several state-of-the-art model-based and model-free offline RL methods on the three benchmark environments across 5 trials.
We  evaluate  the  performance  of  each  method  using  the average  reward  over  20  episodes  for  the  model-based methods and the average reward over 100 episodes for the model-free methods. 
We repeat each experiment five times and report the mean and standard deviation.

\noindent
The results are summarized in Tables \ref{table:avg_rew_MC_appendix}, \ref{table:ave_rew_CP_appendix}, and \ref{table:ave_rew_HC_appendix} for MountainCar, CartPole, and HalfCheetah, respectively.

\begin{table}[ht]
\vspace{-4mm}
\caption{Average Reward: MountainCar}
\label{table:avg_rew_MC_appendix}
\tabcolsep=0.15cm
\fontsize{9.0pt}{11.0pt}\selectfont
\centering
\begin{tabular}{@{}p{0.4cm}p{1.9cm}p{1.9cm}p{1.9cm}p{1.9cm}p{1.9cm}p{1.9cm}@{}}
\toprule
\multirow{2}{*}{Data}    & \multirow{2}{*}{Method}       & {Agent 1}              & {Agent 2}      & {{Agent 3}} & {Agent 4}               & {Agent 5}               \\
&  & {0.0001 }  & {0.0005}   & {0.001} & {0.0025}   & {0.0035}    \\ \toprule
\multirow{5}{*}{\rotatebox[origin=c]{90}{{{Pure}} }} 
&PerSim& -56.80$\pm$\scriptsize{1.83} & -74.30$\pm$\scriptsize{6.59} & -114.1$\pm$\scriptsize{16.1} & {-189.4}$\pm$\scriptsize{6.44} & {-210.6}$\pm$\scriptsize{4.27} \\
&Vanilla CaDM& -106.3$\pm$\scriptsize{44.1} & -289.8$\pm$\scriptsize{195} &  -332.8$\pm$\scriptsize{193}  &  -432.3$\pm$\scriptsize{117}  &  -471.8$\pm$\scriptsize{43.0}   \\
&PE-TS  CaDM&  -74.23$\pm$\scriptsize{16.5}  &  -119.1$\pm$\scriptsize{44.4} & -361.3$\pm$\scriptsize{240}  & -492.3$\pm$\scriptsize{13.3} &  -500.0$\pm$\scriptsize{0.0}  \\
&BCQ-P&  -67.60$\pm$\scriptsize{22.3}  & -68.60$\pm$\scriptsize{19.6} & -79.20$\pm$\scriptsize{14.7} & -267.8$\pm$\scriptsize{202} & -295.1$\pm$\scriptsize{180} \\
&BCQ-A&	{-44.79}$\pm$\scriptsize{0.08}  & 	{-50.50}$\pm$\scriptsize{0.40} &   	{-63.52}$\pm$\scriptsize{0.19}     &-380.7$\pm$\scriptsize{170}  & -500.0$\pm$\scriptsize{0.0}   \\
&CQL-P& -176.1$\pm$\scriptsize{45.2} &-161.8$\pm$\scriptsize{40.6} &-166.1$\pm$\scriptsize{33.9} &-316.4$\pm$\scriptsize{26.4} &-362.9$\pm$\scriptsize{17.9} \\ 
&CQL-A&  -44.60$\pm$\scriptsize{0.0} &-49.70$\pm$\scriptsize{0.0} &-63.30$\pm$\scriptsize{0.3} &-500.0$\pm$\scriptsize{0.0} &-499.3$\pm$\scriptsize{0.7}\\
&MOReL-P& -46.00$\pm$\scriptsize{1.1} &-53.20$\pm$\scriptsize{3.3} &-220.3$\pm$\scriptsize{162.6} &-500.0$\pm$\scriptsize{0.0} &-500.0$\pm$\scriptsize{0.0}\\
&MOReL-A &-373.0$\pm$\scriptsize{33.5}	&-488.8$\pm$\scriptsize{4.9}	&-499.4$\pm$\scriptsize{0.2} 	&-500.0$\pm$\scriptsize{0.0}	&-500.0$\pm$\scriptsize{0.0}\\
\midrule
\multirow{5}{*}{\rotatebox[origin=c]{90}{{{Random}}}} &PerSim & {-57.70}$\pm$\scriptsize{5.63} & {-74.30}$\pm$\scriptsize{6.39} & {-120.4}$\pm$\scriptsize{2.17} & {-186.6}$\pm$\scriptsize{4.25} & {-210.1}$\pm$\scriptsize{4.48}  \\
&Vanilla  CaDM & -62.57$\pm$\scriptsize{5.11} & -75.27$\pm$\scriptsize{4.59} &  -274.9$\pm$\scriptsize{74.2}  &  -479.3$\pm$\scriptsize{21.7}  &  -497.5$\pm$\scriptsize{4.39}   \\
&PE-TS  CaDM&  -82.00$\pm$\scriptsize{3.47}  &  -115.7$\pm$\scriptsize{7.63} & -472.1$\pm$\scriptsize{48.3}  & -500.0$\pm$\scriptsize{0.0} &  -500.0$\pm$\scriptsize{0.0}  \\
&BCQ-P& -500.0$\pm$\scriptsize{0.0}                 &   -500.0$\pm$\scriptsize{0.0}                   &    -500.0$\pm$\scriptsize{0.0}                   & -500.0$\pm$\scriptsize{0.0}                    &   -500.0$\pm$\scriptsize{0.0}                     \\
&BCQ-A&                 -500.0$\pm$\scriptsize{0.0} & -500.0$\pm$\scriptsize{0.0} &           -500.0$\pm$\scriptsize{0.0} &   -500.0$\pm$\scriptsize{0.0} &  -500.0$\pm$\scriptsize{0.0}  \\ 
&CQL-P&  -500.0$\pm$\scriptsize{0.0} &-500.0$\pm$\scriptsize{0.0} &-500.0$\pm$\scriptsize{0.0} &-500.0$\pm$\scriptsize{0.0} &-500.0$\pm$\scriptsize{0.0} \\ 
&CQL-A&  -500.0$\pm$\scriptsize{0.0} &-500.0$\pm$\scriptsize{0.0} &-500.0$\pm$\scriptsize{0.0} &-500.0$\pm$\scriptsize{0.0} &-500.0$\pm$\scriptsize{0.0} \\
&MOReL-P& -45.00$\pm$\scriptsize{0.3} &-50.40$\pm$\scriptsize{0.5} &-68.20$\pm$\scriptsize{5.4} &-500.0$\pm$\scriptsize{0.0} &-500.0$\pm$\scriptsize{0.0} \\
&MOReL-A &-495.0$\pm$\scriptsize{3.9}	&-480.6$\pm$\scriptsize{3.1}	&-500.0$\pm$\scriptsize{0.0} 	&-500.0$\pm$\scriptsize{0.0}	&-500.0$\pm$\scriptsize{0.0}\\
\midrule

\multirow{5}{*}{\rotatebox[origin=c]{90}{{{Pure-$\varepsilon$-20}} }} 
&PerSim&   {-54.20}$\pm$\scriptsize{0.56} & {-67.80}$\pm$\scriptsize{0.48} & {-111.7}$\pm$\scriptsize{6.20} & {-191.2}$\pm$\scriptsize{6.70} & {-199.7}$\pm$\scriptsize{3.99} \\
&Vanilla  CaDM & -56.73$\pm$\scriptsize{4.20} & -70.70$\pm$\scriptsize{1.54} &  -148.7$\pm$\scriptsize{11.6}  &  -463.2$\pm$\scriptsize{57.5}  &  -478.9$\pm$\scriptsize{35.8}   \\
&PE-TS  CaDM &  -107.6$\pm$\scriptsize{36.3}  &  -158.5$\pm$\scriptsize{84.3} & -464.6$\pm$\scriptsize{37.3}  & -500.0$\pm$\scriptsize{0.0} &  -500.0$\pm$\scriptsize{0.0}  \\
&BCQ-P&      -71.21$\pm$\scriptsize{24.4}  &  -72.10$\pm$\scriptsize{20.5}      &    	{-78.41}$\pm$\scriptsize{14.3}     &      -286.6$\pm$\scriptsize{196}   &     -328.3$\pm$\scriptsize{158}       \\
&BCQ-A&  -364.5$\pm$\scriptsize{180}   & -435.7$\pm$\scriptsize{63.7}    &    -282.7$\pm$\scriptsize{308}      &  -260.6$\pm$\scriptsize{51.4}  &   -204.5$\pm$\scriptsize{68.9}         \\
&CQL-P& -79.40$\pm$\scriptsize{16.1} &-78.80$\pm$\scriptsize{13.9} &-86.50$\pm$\scriptsize{10.8} &-357.9$\pm$\scriptsize{12.5} &-407.6$\pm$\scriptsize{15.3}  \\ 
&CQL-A& -44.70$\pm$\scriptsize{0.1} &-49.70$\pm$\scriptsize{0.0} &-63.30$\pm$\scriptsize{0.2} &-500.0$\pm$\scriptsize{0.0} &-500.0$\pm$\scriptsize{0.0}  \\
&MOReL-P&  -83.50$\pm$\scriptsize{15.6} &-81.70$\pm$\scriptsize{14.2} &-88.60$\pm$\scriptsize{11.1} &-357.0$\pm$\scriptsize{13.8} &-407.4$\pm$\scriptsize{17.1}  \\
&MOReL-A & -44.60$\pm$\scriptsize{0.1} &-49.70$\pm$\scriptsize{0.0} &-63.30$\pm$\scriptsize{0.2} &-500.0$\pm$\scriptsize{0.0} &-500.0$\pm$\scriptsize{0.0}  \\
\midrule
\multirow{5}{*}{\rotatebox[origin=c]{90}{{{Pure-$\varepsilon$-40}} }} 
&PerSim&  -54.60$\pm$\scriptsize{0.55} & -71.10$\pm$\scriptsize{1.89} & -115.7$\pm$\scriptsize{4.80} & {-189.7}$\pm$\scriptsize{7.14} & {-200.3}$\pm$\scriptsize{2.26} \\
&Vanilla CaDM & -55.23$\pm$\scriptsize{0.76} & -67.90$\pm$\scriptsize{7.30} &  -163.7$\pm$\scriptsize{30.8}  &  -481.7$\pm$\scriptsize{25.3}  &  -496.2$\pm$\scriptsize{4.31}   \\
&PE-TS  CaDM&  -102.3$\pm$\scriptsize{20.3}  &  -120.7$\pm$\scriptsize{18.8} & -476.0$\pm$\scriptsize{41.5}  & -500.0$\pm$\scriptsize{0.0} &  -500.0$\pm$\scriptsize{0.0}  \\
&BCQ-P& {-50.01}$\pm$\scriptsize{7.50}              &{-57.10}$\pm$\scriptsize{10.3}             & {-66.11}$\pm$\scriptsize{3.91}              &   -373.6$\pm$\scriptsize{180}            &     -352.0$\pm$\scriptsize{211}                \\
&BCQ-A&  -94.87$\pm$\scriptsize{0.88}              &   -80.03$\pm$\scriptsize{38.5}           &    -329.6$\pm$\scriptsize{242}                &    -358.7$\pm$\scriptsize{201}                &     -486.5$\pm$\scriptsize{20.6}      \\
&CQL-P&-61.40$\pm$\scriptsize{2.1} &-64.90$\pm$\scriptsize{2.1} &-75.10$\pm$\scriptsize{1.2} &-366.9$\pm$\scriptsize{21.3} &-429.1$\pm$\scriptsize{18.8}  \\ 
&CQL-A& -44.70$\pm$\scriptsize{0.0} &-49.70$\pm$\scriptsize{0.0} &-63.30$\pm$\scriptsize{0.4} &-500.0$\pm$\scriptsize{0.0} &-490.1$\pm$\scriptsize{9.6} \\
&MOReL-P& -61.30$\pm$\scriptsize{2.3} &-65.80$\pm$\scriptsize{1.2} &-75.00$\pm$\scriptsize{1.4} &-373.2$\pm$\scriptsize{19.3} &-428.5$\pm$\scriptsize{21.0}  \\
&MOReL-A & -44.70$\pm$\scriptsize{0.0} &-49.70$\pm$\scriptsize{0.0} &-63.30$\pm$\scriptsize{0.4} &-500.0$\pm$\scriptsize{0.0} &-492.1$\pm$\scriptsize{9.3} \\
\cmidrule[1.5pt]{1-7}
&True env+MPC& {-53.95$\pm$\scriptsize{4.10}}   & {-72.43$\pm$\scriptsize{7.80}}  & {-110.8$\pm$\scriptsize{23.8}}  & {-182.9$\pm$\scriptsize{22.9}} & {-197.5$\pm$\scriptsize{20.7}}   \\
\cmidrule[1.5pt]{1-7}
\end{tabular}
\vspace{-4mm}
\end{table}

\begin{table}[h]
\vspace{-4mm}
\caption{Average Reward: CartPole}
\label{table:ave_rew_CP_appendix}
\tabcolsep=0.15cm
\fontsize{9.0pt}{11.0pt}\selectfont
\centering
\begin{tabular}{@{}p{0.4cm}p{1.9cm}p{1.9cm}p{1.9cm}p{1.9cm}p{1.9cm}p{1.9cm}@{}}
\toprule
\multirow{2}{*}{Data}    & \multirow{2}{*}{Method}       & {Agent 1}              & {Agent 2}      & {{Agent 3}} & {Agent 4}               & {Agent 5}               \\
& & {(2/0.5) } & {(10.0/0.5) } & {(18.0/0.5)} & {(10/0.85)}   & {(10/0.15)}     \\ \toprule
\multirow{5}{*}{\rotatebox[origin=c]{90}{{{Pure}} }} 
&PerSim &  {199.7}$\pm$\scriptsize{0.58} & 198.7$\pm$\scriptsize{0.86} & 198.5$\pm$\scriptsize{1.16} & {193.8}$\pm$\scriptsize{4.28} & {192.0}$\pm$\scriptsize{2.28}  \\
&Vanilla CaDM&     168.0$\pm$\scriptsize{19.7} &    197.7$\pm$\scriptsize{1.50}  &  173.6$\pm$\scriptsize{6.10} &    190.8$\pm$\scriptsize{6.80}  &   58.10$\pm$\scriptsize{10.7}                \\
&PE-TS CaDM&  92.30$\pm$\scriptsize{44.8}    &    {200.0}$\pm$\scriptsize{0.0}    &  {200.0}$\pm$\scriptsize{0.0}      &  193.6$\pm$\scriptsize{8.30}            &  127.5$\pm$\scriptsize{9.50}      \\
&BCQ-P&   166.2$\pm$\scriptsize{39.3}  & 187.4$\pm$\scriptsize{14.7} & 187.2$\pm$\scriptsize{15.0}   &  181.2$\pm$\scriptsize{13.5}   &   182.8$\pm$\scriptsize{15.0}  \\
&BCQ-A&  65.40$\pm$\scriptsize{67.5}   & 138.0$\pm$\scriptsize{80.3}   &   79.20$\pm$\scriptsize{79.9}  & 79.20$\pm$\scriptsize{69.5}    &   132.1$\pm$\scriptsize{85.0} \\
&CQL-P& 154.4$\pm$\scriptsize{17.7} &190.8$\pm$\scriptsize{1.2} &193.9$\pm$\scriptsize{1.7} &190.9$\pm$\scriptsize{0.7} &170.2$\pm$\scriptsize{0.6}  \\ 
&CQL-A&  122.8$\pm$\scriptsize{63.4} &189.4$\pm$\scriptsize{6.6} &199.3$\pm$\scriptsize{0.7} &179.4$\pm$\scriptsize{20.6} &193.6$\pm$\scriptsize{5.4} \\
&MOReL-P& 35.80$\pm$\scriptsize{1.4} &90.40$\pm$\scriptsize{26.2} &94.20$\pm$\scriptsize{25.1} &96.60$\pm$\scriptsize{24.1} &66.40$\pm$\scriptsize{24.5}\\
&MOReL-A& 33.70$\pm$\scriptsize{3.8} &16.00$\pm$\scriptsize{6.5} &16.20$\pm$\scriptsize{0.6} &27.50$\pm$\scriptsize{0.1} &10.10$\pm$\scriptsize{0.7} \\ \midrule

\multirow{5}{*}{\rotatebox[origin=c]{90}{{{Random}} }} 
&PerSim& {197.7}$\pm$\scriptsize{7.82} & 189.5$\pm$\scriptsize{4.28} & 190.3$\pm$\scriptsize{4.74} & 193.0$\pm$\scriptsize{6.60} & {185.7}$\pm$\scriptsize{3.49}  \\
&Vanilla CaDM&   150.5$\pm$\scriptsize{15.7} &   158.7$\pm$\scriptsize{17.1} &   161.4$\pm$\scriptsize{15.8} &     175.6$\pm$\scriptsize{5.80}  &    65.10$\pm$\scriptsize{16.3} \\
&PE-TS CaDM&     88.60$\pm$\scriptsize{18.5}  &      {194.0}$\pm$\scriptsize{2.00}   &    {197.6}$\pm$\scriptsize{2.20}   &     {196.1}$\pm$\scriptsize{3.00}    &     171.0$\pm$\scriptsize{21.7}              \\
&BCQ-P&   44.80$\pm$\scriptsize{34.0}    &    58.21$\pm$\scriptsize{58.0}   &    56.92$\pm$\scriptsize{56.0}    &    57.91$\pm$\scriptsize{53.0}    &    36.40$\pm$\scriptsize{36.0}     \\
&BCQ-A&   43.90$\pm$\scriptsize{16.4} &    18.70$\pm$\scriptsize{13.1} &    7.200$\pm$\scriptsize{0.84}   &    21.10$\pm$\scriptsize{5.81}  &    39.50$\pm$\scriptsize{12.1}       \\ 
&CQL-P&  39.90$\pm$\scriptsize{29.9} &72.30$\pm$\scriptsize{41.7} &67.80$\pm$\scriptsize{44.4} &77.90$\pm$\scriptsize{26.2} &148.7$\pm$\scriptsize{17.1} \\ 
&CQL-A&  67.40$\pm$\scriptsize{49.1} &9.300$\pm$\scriptsize{0.1} &16.30$\pm$\scriptsize{9.9} &30.60$\pm$\scriptsize{26.4} &7.000$\pm$\scriptsize{1.9} \\
&MOReL-P& 50.20$\pm$\scriptsize{7.9} &59.10$\pm$\scriptsize{17.8} &57.80$\pm$\scriptsize{17.8} &68.90$\pm$\scriptsize{17.4} &40.40$\pm$\scriptsize{12.4}\\
&MOReL-A& 35.80$\pm$\scriptsize{0.4} &20.70$\pm$\scriptsize{1.0} &14.30$\pm$\scriptsize{1.4} &27.50$\pm$\scriptsize{0.7} &10.60$\pm$\scriptsize{0.2}\\
\midrule
\multirow{5}{*}{\rotatebox[origin=c]{90}{{{Pure-$\varepsilon$-20}} }} 
&PerSim& {199.8}$\pm$\scriptsize{0.24} & {200.0}$\pm$\scriptsize{0.0} & {200.0}$\pm$\scriptsize{0.0} & {199.1}$\pm$\scriptsize{1.3} & {197.8}$\pm$\scriptsize{1.68}  \\
&Vanilla CaDM&  171.1$\pm$\scriptsize{38.1} &   195.3$\pm$\scriptsize{3.00} &   180.7$\pm$\scriptsize{4.30} &     193.4$\pm$\scriptsize{2.10}  &    64.20$\pm$\scriptsize{10.0}                \\
&PE-TS CaDM&   98.30$\pm$\scriptsize{42.9} &   {199.6}$\pm$\scriptsize{0.50}  &   199.0$\pm$\scriptsize{1.40}  &      198.6$\pm$\scriptsize{0.40}  &    141.1$\pm$\scriptsize{12.0}  \\
&BCQ-P&   98.90$\pm$\scriptsize{30.2} &   170.9$\pm$\scriptsize{19.0}      &  163.0$\pm$\scriptsize{36.5}      &   162.1$\pm$\scriptsize{15.5}   &     86.10$\pm$\scriptsize{72.1}       \\
&BCQ-A& 67.30$\pm$\scriptsize{62.2}   &  130.0$\pm$\scriptsize{76.1}    &     33.40$\pm$\scriptsize{0.62}     &   65.60$\pm$\scriptsize{51.8}        & 140.0$\pm$\scriptsize{80.6}         \\
&CQL-P& 163.7$\pm$\scriptsize{13.6} &197.4$\pm$\scriptsize{2.9} &198.6$\pm$\scriptsize{2.4} &198.1$\pm$\scriptsize{2.6} &190.4$\pm$\scriptsize{6.6}  \\ 
&CQL-A&  42.40$\pm$\scriptsize{11.9} &189.8$\pm$\scriptsize{10.2} &22.20$\pm$\scriptsize{20.0} &199.0$\pm$\scriptsize{1.9} &199.8$\pm$\scriptsize{0.2}  \\
&MOReL-P& 166.7$\pm$\scriptsize{13.6} &197.0$\pm$\scriptsize{3.1} &198.3$\pm$\scriptsize{2.6} &197.7$\pm$\scriptsize{2.7} &189.4$\pm$\scriptsize{7.1}  \\
&MOReL-A& 41.40$\pm$\scriptsize{13.1} &188.3$\pm$\scriptsize{11.0} &20.00$\pm$\scriptsize{21.9} &198.8$\pm$\scriptsize{2.1} &199.9$\pm$\scriptsize{0.2}  \\
\midrule
\multirow{5}{*}{\rotatebox[origin=c]{90}{{{Pure-$\varepsilon$-40}} }} 
&PerSim&   {199.9}$\pm$\scriptsize{0.18} & {199.8}$\pm$\scriptsize{0.20} & 199.3$\pm$\scriptsize{1.34} & {198.0}$\pm$\scriptsize{1.21} & {197.4}$\pm$\scriptsize{1.72} \\
&Vanilla CaDM&    160.6$\pm$\scriptsize{46.6} &     197.3$\pm$\scriptsize{1.50} &     194.9$\pm$\scriptsize{3.70}  &   191.9$\pm$\scriptsize{6.40}&   79.60$\pm$\scriptsize{31.6}                    \\
&PE-TS CaDM&    91.90$\pm$\scriptsize{67.6}   &     {199.8}$\pm$\scriptsize{0.20}  &     {200.0}$\pm$\scriptsize{0.0}  &      197.0$\pm$\scriptsize{1.40}   &   143.5$\pm$\scriptsize{17.5}             \\
&BCQ-P&   28.90$\pm$\scriptsize{6.80} &   24.97$\pm$\scriptsize{12.8}                 &      27.90$\pm$\scriptsize{25.9}              &       31.80$\pm$\scriptsize{25.9}             &       18.50$\pm$\scriptsize{11.1}              \\
&BCQ-A&  34.60$\pm$\scriptsize{1.55}              &  23.20$\pm$\scriptsize{17.8}            &     7.180$\pm$\scriptsize{0.76}               &    47.71$\pm$\scriptsize{48.7}                &     23.20$\pm$\scriptsize{9.44}       \\
&CQL-P& 182.2$\pm$\scriptsize{18.0} &197.4$\pm$\scriptsize{4.6} &198.9$\pm$\scriptsize{2.1} &198.3$\pm$\scriptsize{2.9} &191.9$\pm$\scriptsize{8.7} \\ 
&CQL-A&   20.70$\pm$\scriptsize{0.7} &25.70$\pm$\scriptsize{14.3} &15.30$\pm$\scriptsize{7.3} &134.2$\pm$\scriptsize{10.2} &9.700$\pm$\scriptsize{8.1} \\
&MOReL-P& 178.9$\pm$\scriptsize{18.7} &196.8$\pm$\scriptsize{5.0} &198.7$\pm$\scriptsize{2.3} &197.9$\pm$\scriptsize{3.1} &190.6$\pm$\scriptsize{9.3}  \\
&MOReL-A& 20.70$\pm$\scriptsize{0.8} &23.90$\pm$\scriptsize{15.5} &16.60$\pm$\scriptsize{7.6} &135.4$\pm$\scriptsize{11.2} &10.80$\pm$\scriptsize{8.8} \\
\cmidrule[1.5pt]{1-7}
&True env+MPC& {200.0$\pm$\scriptsize{0.0}}  & {200.0$\pm$\scriptsize{0.0}} & {200.0$\pm$\scriptsize{0.0}}  & {198.4$\pm$\scriptsize{7.20}}& {200.0$\pm$\scriptsize{0.0}}  \\
\cmidrule[1.5pt]{1-7}
\end{tabular}
\vspace{-4mm}
\end{table}

\begin{table}[h]
\caption{Average Reward:  HalfCheetah}
\label{table:ave_rew_HC_appendix}
\tabcolsep=0.15cm
\fontsize{9.0pt}{11.0pt}\selectfont
\centering
\begin{tabular}{@{}p{0.4cm}p{1.9cm}p{1.9cm}p{1.9cm}p{1.9cm}p{1.9cm}p{1.9cm}@{}}
\toprule
\multirow{2}{*}{Data}    & \multirow{2}{*}{Method}       & {Agent 1}              & {Agent 2}      & {{Agent 3}} & {Agent 4}               & {Agent 5}               \\
&& {(0.3/1.7)} & {(1.7/0.3)}   & {(0.3/0.3)}  & {(1.7/1.7)}  & {(1.0/1.0)}    \\ \toprule
\multirow{5}{*}{\rotatebox[origin=c]{90}{{{Pure}} }} 
&{PerSim}&   ~{1984}~$\pm$\scriptsize{763} &	~997.0$\pm$\scriptsize{403} &	~{714.7}$\pm$\scriptsize{314} &	113.5$\pm$\scriptsize{289} &	~1459~$\pm$\scriptsize{398}\\
&Vanilla CaDM& ~50.31$\pm$\scriptsize{71.7}  & -134.0$\pm$\scriptsize{81.1}  & ~11.39$\pm$\scriptsize{171}  &-169.8$\pm$\scriptsize{67.5}  & ~331.3$\pm$\scriptsize{201}   \\
&PE-TS CaDM& ~481.1$\pm$\scriptsize{252} & ~503.7$\pm$\scriptsize{181} & ~553.0$\pm$\scriptsize{127} & ~246.0$\pm$\scriptsize{261} & ~840.1$\pm$\scriptsize{383}  \\
&BCQ-P& 	{~549.8}$\pm$\scriptsize{322} & 	{~2006~}$\pm$\scriptsize{153} & -65.18$\pm$\scriptsize{92.8} & 	{~2564~}$\pm$\scriptsize{70.2} & 	{~2469~}$\pm$\scriptsize{67.2}  \\
&BCQ-A& -262.7$\pm$\scriptsize{96.6} & -139.0$\pm$\scriptsize{236} & ~165.6$\pm$\scriptsize{83.1} & ~1649~$\pm$\scriptsize{622} & ~937.2$\pm$\scriptsize{221}    \\
&CQL-P& -353.5$\pm$\scriptsize{78.4} &-453.6$\pm$\scriptsize{71.9} &-476.7$\pm$\scriptsize{129} &~2037~$\pm$\scriptsize{294} &-145.1$\pm$\scriptsize{189} \\ 
&CQL-A& -65.00$\pm$\scriptsize{105} &-257.9$\pm$\scriptsize{35.9} &-279.6$\pm$\scriptsize{34.4} &~689.3$\pm$\scriptsize{52.9} &~301.9$\pm$\scriptsize{98.9} \\
&MOReL-P& -1297$\pm$\scriptsize{519} &-1256~$\pm$\scriptsize{627} &-1470~$\pm$\scriptsize{727} &-1175~$\pm$\scriptsize{592} &-1256~$\pm$\scriptsize{608}  \\
&MOReL-A& -726.2$\pm$\scriptsize{4.9} &-666.1$\pm$\scriptsize{42.6} &-841.7$\pm$\scriptsize{39.5} &-599.6$\pm$\scriptsize{28.2} &-688.6$\pm$\scriptsize{12.4} \\\midrule

\multirow{5}{*}{\rotatebox[origin=c]{90}{{{Random}} }} 
&{PerSim}&  ~{2124}~$\pm$\scriptsize{518}&	~{2060}~$\pm$\scriptsize{900}&	~472.0$\pm$\scriptsize{56.9}&	~{565.2}$\pm$\scriptsize{377}&	~474.8$\pm$\scriptsize{344}\\
&Vanilla CaDM& ~288.4$\pm$\scriptsize{32.4}  & ~362.9$\pm$\scriptsize{55.4}  & ~351.8$\pm$\scriptsize{34.9}  & ~358.4$\pm$\scriptsize{205}  & ~475.0$\pm$\scriptsize{102}   \\
&PE-TS CaDM& ~754.6$\pm$\scriptsize{242} & ~744.5$\pm$\scriptsize{281} & {~767.4}$\pm$\scriptsize{214} & {~555.4}$\pm$\scriptsize{73.1} & {~2486~}$\pm$\scriptsize{1488}  \\
&BCQ-P& -1.460$\pm$\scriptsize{0.16} & -1.750$\pm$\scriptsize{0.22} & -1.690$\pm$\scriptsize{0.19} & -1.790$\pm$\scriptsize{0.21} & -1.720$\pm$\scriptsize{0.20} \\
& BCQ-A& -498.9$\pm$\scriptsize{108} & -113.3$\pm$\scriptsize{13.0} & -159.5$\pm$\scriptsize{51.7} & -35.73$\pm$\scriptsize{7.22} & -171.9$\pm$\scriptsize{41.5}  \\ 
&CQL-P& -481.5$\pm$\scriptsize{25.9} &-442.2$\pm$\scriptsize{56.4} &-672.4$\pm$\scriptsize{17.6} &-254.5$\pm$\scriptsize{39.6} &-418.2$\pm$\scriptsize{23.4}  \\ 
&CQL-A& -0.700$\pm$\scriptsize{0.4} &-2.800$\pm$\scriptsize{0.8} &-0.600$\pm$\scriptsize{0.3} &-2.000$\pm$\scriptsize{0.6} &-5.500$\pm$\scriptsize{2.0} \\
&MOReL-P& -102.3$\pm$\scriptsize{45.7} &-188.7$\pm$\scriptsize{37.1} &-181.0$\pm$\scriptsize{67.7} &-142.5$\pm$\scriptsize{26.8} &-141.4$\pm$\scriptsize{7.3} \\
&MOReL-A& -430.8$\pm$\scriptsize{195} &-673.7$\pm$\scriptsize{39.3} &-365.5$\pm$\scriptsize{97.4} &-645.5$\pm$\scriptsize{53.0} &-674.9$\pm$\scriptsize{28.9} \\\midrule
\multirow{5}{*}{\rotatebox[origin=c]{90}{{{Pure-$\varepsilon$-20}} }} 
&{PerSim}&   ~{3186}~$\pm$\scriptsize{604}&
~1032~$\pm$\scriptsize{232}&
~1120~$\pm$\scriptsize{243}&
~971.2$\pm$\scriptsize{916}&
~1666~$\pm$\scriptsize{930}\\
&Vanilla CaDM& ~412.0$\pm$\scriptsize{152}  & ~31.92$\pm$\scriptsize{109}  & ~460.2$\pm$\scriptsize{159}  & ~60.33$\pm$\scriptsize{139}  & ~166.6$\pm$\scriptsize{71.8}   \\
&PE-TS CaDM& {~1082~}$\pm$\scriptsize{126} & ~{1125}~$\pm$\scriptsize{132} & {~1067~}$\pm$\scriptsize{64.3} & ~1098~$\pm$\scriptsize{344} & {~2843~}$\pm$\scriptsize{204}  \\
&BCQ-P& ~254.6$\pm$\scriptsize{352} & ~406.7$\pm$\scriptsize{71.1} & ~385.9$\pm$\scriptsize{57.1} & -95.34$\pm$\scriptsize{65.8} & 	{~738.0}$\pm$\scriptsize{512}  \\
&BCQ-A& ~376.8$\pm$\scriptsize{102} & ~84.66$\pm$\scriptsize{53.3} & ~230.1$\pm$\scriptsize{10.0} & 	{~1180~}$\pm$\scriptsize{87.3} & ~617.5$\pm$\scriptsize{32.6}   \\ 
&CQL-P& ~838.7$\pm$\scriptsize{24.5} &~3155~$\pm$\scriptsize{125} &~539.9$\pm$\scriptsize{313} &~1479~$\pm$\scriptsize{51.0} &~3561~$\pm$\scriptsize{170}  \\ 
&CQL-A& -15.50$\pm$\scriptsize{9.0} &-73.00$\pm$\scriptsize{26.3} &-108.0$\pm$\scriptsize{64.6} &~656.9$\pm$\scriptsize{181.6} &~357.6$\pm$\scriptsize{110} \\
&MOReL-P& ~0.600$\pm$\scriptsize{210} &-171.2$\pm$\scriptsize{125} &-219.9$\pm$\scriptsize{76.7} &-106.3$\pm$\scriptsize{18.6} &-83.60$\pm$\scriptsize{110} \\
&MOReL-A& -781.0$\pm$\scriptsize{37.9} &-613.1$\pm$\scriptsize{49.9} &-847.1$\pm$\scriptsize{64.7} &-599.2$\pm$\scriptsize{4.6} &-702.9$\pm$\scriptsize{26.8}  \\\midrule
\multirow{5}{*}{\rotatebox[origin=c]{90}{{{Pure-$\varepsilon$-40}} }} 
&{PerSim}&~{2590}~$\pm$\scriptsize{813}&	
~1016~$\pm$\scriptsize{283}&	
~{1365}~$\pm$\scriptsize{582}&
~803.9$\pm$\scriptsize{912}&
~724.9$\pm$\scriptsize{236}\\
&Vanilla CaDM& ~465.6$\pm$\scriptsize{49.2}   & ~452.7$\pm$\scriptsize{130}  & ~720.0$\pm$\scriptsize{74.9}  & ~176.7$\pm$\scriptsize{359}  & ~952.8$\pm$\scriptsize{591}    \\
&PE-TS CaDM&  {~1500~}$\pm$\scriptsize{246} & ~{1218}~$\pm$\scriptsize{221} & {~1339~}$\pm$\scriptsize{54.8} & {~1569~}$\pm$\scriptsize{306} & {~3094~}$\pm$\scriptsize{825}  \\
&BCQ-P& ~78.25$\pm$\scriptsize{200}   & ~173.8$\pm$\scriptsize{189}& ~417.1$\pm$\scriptsize{155}& -56.12$\pm$\scriptsize{64.4}& ~55.46$\pm$\scriptsize{128}  \\
&BCQ-A& ~269.2$\pm$\scriptsize{60.7}  & -181.5$\pm$\scriptsize{57.4}& ~193.0$\pm$\scriptsize{31.8}& 	{~636.4}$\pm$\scriptsize{137}& ~207.0$\pm$\scriptsize{106}     \\ 
&CQL-P& ~808.5$\pm$\scriptsize{240} &~1662~$\pm$\scriptsize{220} &-156.3$\pm$\scriptsize{119} &~1416~$\pm$\scriptsize{71.8} &~1908~$\pm$\scriptsize{461}  \\ 
&CQL-A& -6.200$\pm$\scriptsize{2.9} &-386.0$\pm$\scriptsize{42.4} &~37.10$\pm$\scriptsize{154} &~1121~$\pm$\scriptsize{95.6} &~1184~$\pm$\scriptsize{604} \\
&MOReL-P& ~8.500$\pm$\scriptsize{61.6} &-114.2$\pm$\scriptsize{72.2} &-195.9$\pm$\scriptsize{77.3} &-66.60$\pm$\scriptsize{7.5} &~22.60$\pm$\scriptsize{28.7}  \\
&MOReL-A& -325.9$\pm$\scriptsize{17.1} &-644.5$\pm$\scriptsize{18.8} &-798.8$\pm$\scriptsize{130} &-609.0$\pm$\scriptsize{16.8} &-711.7$\pm$\scriptsize{19.1} \\
\cmidrule[1.5pt]{1-7}
&True env+MPC& {~7459~$\pm$\scriptsize{171}}  & {42893$\pm$\scriptsize{6959}}  & {66675$\pm$\scriptsize{9364}}   & {1746$\pm$\scriptsize{624}} &  {36344$\pm$\scriptsize{7924}}    \\
\cmidrule[1.5pt]{1-7}
\end{tabular}
\vspace{-5mm}
\end{table}

\FloatBarrier
\subsection{Visualization of Agent Latent Factors} \label{appendix:sec:latent_factors}

\begin{figure}[h!tb]
\centering
\begin{subfigure}[t]{0.33\textwidth}
        \centering
    \includegraphics[width = 1.0 \linewidth]{figures/MC_LF_.png}
    \caption{MountainCar: gravity}
    \label{fig:MC_LF}
\end{subfigure}
\begin{subfigure}[t]{0.33\textwidth}
        \centering
    \includegraphics[width =1.0 \linewidth]{figures/CP_force_.png}
    \caption{CartPole: push force}
    \label{fig:CP_LF_force}
\end{subfigure}
\begin{subfigure}[t]{0.33\textwidth}
        \centering
    \includegraphics[width =1.0 \linewidth]{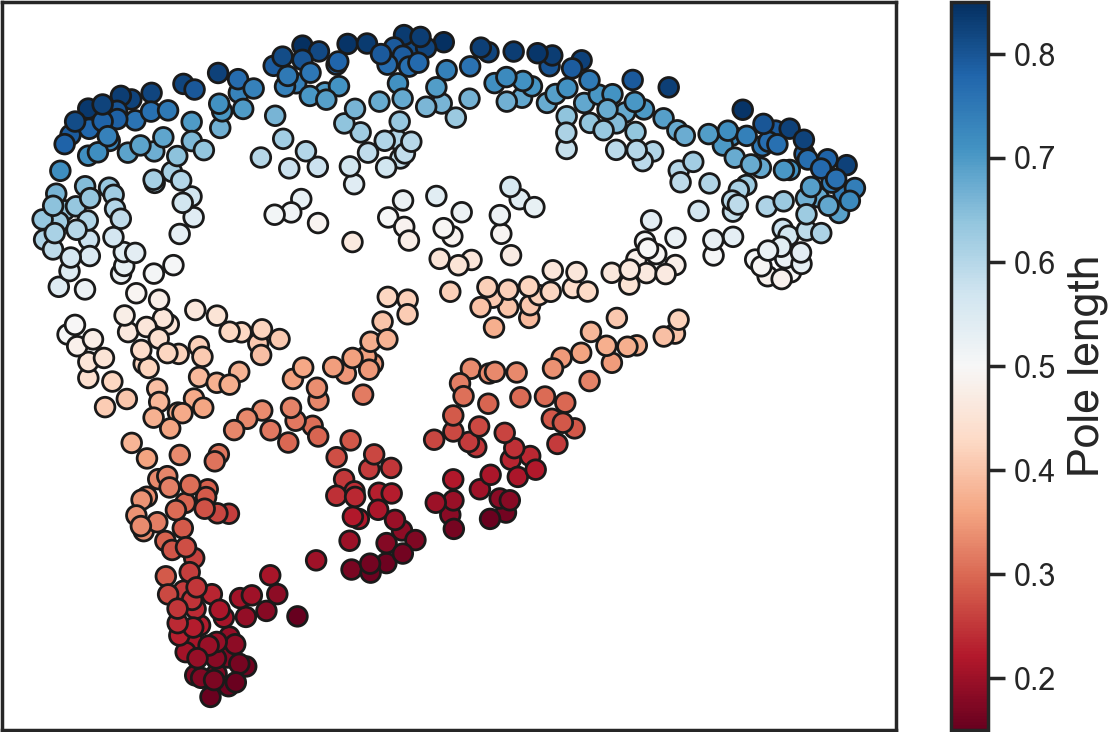}
    \caption{CartPole: pole length}
    \label{fig:CP_LF_length}
\end{subfigure}
\begin{subfigure}[t]{0.33\textwidth}
        \centering
    \includegraphics[width = 1.0 \linewidth]{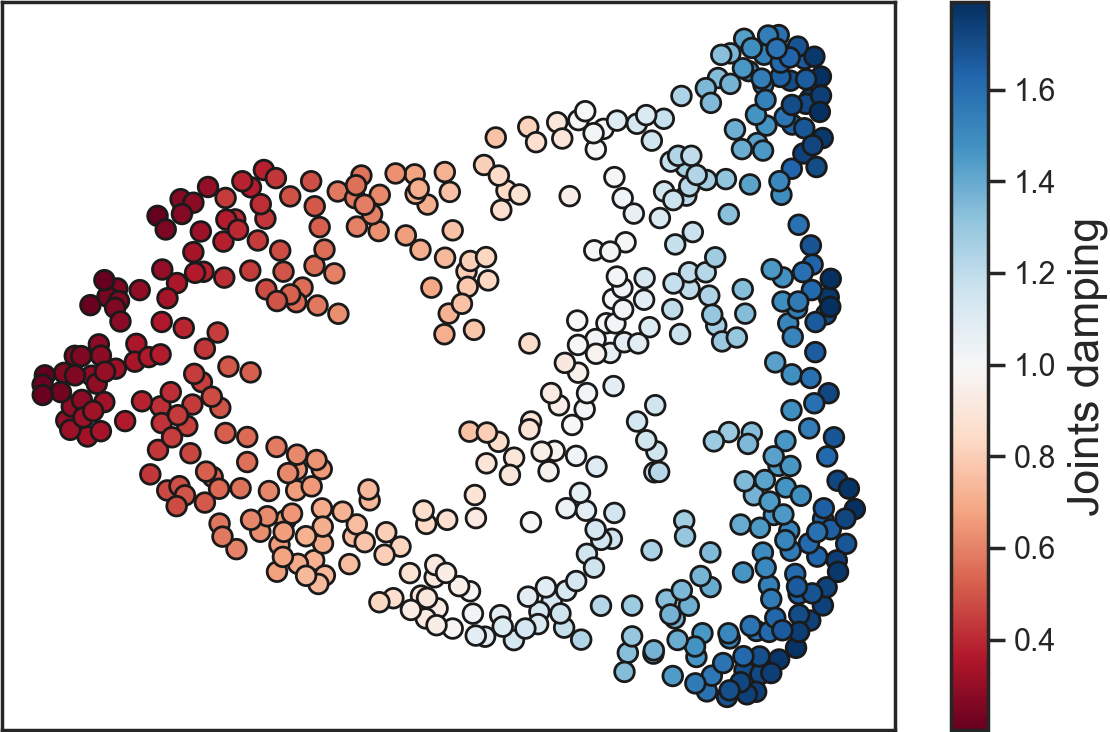}
    \caption{HalfCheetah: joints' damping}
    \label{fig:HC_LF_damp}
\end{subfigure}
\begin{subfigure}[t]{0.33\textwidth}
        \centering
    \includegraphics[width =  \linewidth]{figures/HC_mass_.png}
    \caption{HalfCheetah: links' mass}
    \label{fig:HC_LF_mass}
\end{subfigure}
\caption{t-SNE \cite{van2008visualizing} visualization of the agent latent factors for the 500 heterogeneous agents in MountainCar, CartPole, and HalfCheetah. Colors indicate the value of the modified parameter in each environment (e.g., gravity in MountainCar). These figures demonstrate that the learned latent factors indeed capture the relevant information about the agents heterogeneity in all environments.}
    \label{fig:LF}
\end{figure}

In this section, we visualize the learned agent latent factors associated with the 500 heterogeneous agents in each of the three benchmark environments. 
Specifically, we visualize the agent latent factors in
MountainCar, as we vary the gravity (Figure \ref{fig:MC_LF}); 
CartPole, as we vary the pole's length and the push force (Figures \ref{fig:CP_LF_length} and \ref{fig:CP_LF_force}, respectively); 
HalfCheetah, as we vary the cheetah's mass and the joints' damping (Figures \ref{fig:HC_LF_mass} and \ref{fig:HC_LF_damp}, respectively).

\colorred{
\subsection{Persim+BCQ/CQL: Experimental details}\label{appendix:bcq_persim}

We evaluate PerSim's simulation efficacy by quantifying how much the simulated trajectories improve the performance of model-free RL methods such as BCQ and CQL.
In particular, we use PerSim to generate synthetic trajectories for each agent of interest to augment the training data available for BCQ/CQL. 

\noindent We carry out these experiments for the three environments: MountainCar, CartPole, and HalfCheetah. In all environments, as is done in previous experiments (see Section \ref{sec:setup}), we train PerSim using a single observed trajectory from each of the 500 training agents. 
}
For these trajectories,  the actions are selected randomly, and the covariates of the training agents are selected as described in Section \ref{sec:setup}.
Then, we use the trained simulators to produce $5$ synthetic trajectories for each test agent. See Table \ref{table:env} for information about the test agents in each environment and the range of covariates used for the training agents. 

\noindent When generating these synthetic trajectories, we use MPC to choose the sequence of actions that maximizes the reward estimated by the simulator, as described in Appendix \ref{appendix:sec:implementation_our}. We use  a horizon $h$ of 50 for MountainCar and CartPole, and a horizon $h$ of 30 for HalfCheetah. 
The only difference is that instead of choosing the first element from the sequence of actions with the best {average} reward, we choose the full sequence of actions, and repeat the sampling process until we have a full trajectory.  

\vspace{2mm}

\subsection{Generalizing to Unseen Agents} \label{appendix:unseen_agents}

\noindent \textbf{Setup.} In this experiment, we evaluate PerSim's ability to generalize to unseen agents. 
An advantage of the factorized approach of PerSim is that the heterogeneity of an agent is  captured by its latent agent factors (see Table~\ref{fig:LF}). 
Hence, the problem of generalizing to unseen agents boils down to accurately estimating the latent agent factors. 
To estimate these latent agent factors, we assume access to the covariates of the unseen agents, as well as a fraction $1 \ge p > 0$ of the covariates of the training agents. 
With access to this information, we propose the following natural two-step procedure to estimate the latent agent factor:
(1) Use a supervised learning method to learn a mapping between the (available) training agent covariates and the learned agent-specific latent factor in PerSim (see Figure \ref{fig:LF_main_paper});
(2) Apply this mapping on the covariate data of an unseen test agent to estimate its latent agent factor, which is sufficient to build a personalized simulator for it.
\vspace{2mm}

\noindent We conduct this experiment for MountainCar, where we train PerSim using 500 agents, each with a gravity value selected uniformly at random from the range $[0.0001, 0.0035]$. 
We then evaluate PerSim's performance on 5 unseen agents selected as follow:
\begin{enumerate}
    \item One unseen agent from the training range $[0.0001, 0.0035]$. Specifically, the one with gravity $0.002$.
    \item Two agents  outside the lower end of the range with gravity of $0.00008$ and $0.00005$.
    \item Two agents  outside the upper end of the range with gravity of $0.0037$, and $0.004$.
\end{enumerate}
\noindent We first train PerSim on trajectories generated from the aforementioned 500 training agents, and carry out the experiments for trajectories generated via the random and pure policies. 
Then, we learn a mapping between the learned agent factors and covariates through an MLP with 2 hidden-layers each with 64 units.
We assume access to a fraction $p\in \{1.0,0.5,0.25,0.1\}$ of the covariates to train this function.
We report PerSim's efficacy through the same two metrics we used before: prediction error and  average reward for the five unseen agents.

\vspace{2mm}

\noindent \textbf{Results.}  As Table \ref{table:unseen_agent_results}  shows, in terms of prediction error, PerSim outperforms both Vanilla CaDM and PE-TS CaDM in both the random and pure datasets for all unseen test agents.
Further,  Table \ref{table:unseen_agents_rew} shows that PerSim achieves the best reward among the three baselines (BCQ-P, Vanilla CaDM and PE-TS CaDM) for most unseen test agents.
Pleasingly, these results are consistent as we vary $p \in \{1.0,0.5,0.25,0.1\}$.
That is,  even with access to only $10\%$ of the covariates of the training agents (i.e., $p = 0.1$), PerSim is able to simulate the unseen agents well. 
Note that PerSim is trained without access to any of the latent covariates (e.g., gravity in MountainCar), but to generalize to unseen agents, it requires access to {some} of the covariates to learn the mapping between these covariates and the latent agent factors. 
It is important to note that PerSim utilizes explicit knowledge of the covariates of the unseen test agents (and a subset of the training agents), which these other methods do not do in their current implementations.
Indeed, it is our latent factor representation (in particular, the agent-specific latent factors) which seamlessly allows us to utilize these covariates to build simulators for the unseen agents.
%
%

\begin{table}[h]
\caption{Prediction Error: MountainCar, Unseen Agents}
\label{table:unseen_agent_results}
\tabcolsep=0.02cm
\fontsize{7.5pt}{9.0pt}\selectfont
\centering
\begin{tabular}{@{}p{0.8cm} p{2.2cm}p{2.2cm}p{2.2cm}p{2.2cm}p{2.2cm}p{2.2cm}  @{}}
\toprule
\multirow{2}{*}{Data}     & \multirow{2}{*}{Method}       & {Agent 1}              &  {Agent 2}      & {{Agent 3}} & {Agent 4}               & {Agent 5}      \\     
& & {0.00005 }  & {0.00008}   & {0.002} & {0.0037}   & {0.004}     \\ \toprule
\multirow{6}{*}{\rotatebox[origin=c]{90}{{{Random}} }} &
PerSim(p=1.0)& {0.004 (1.00)} & {0.004 (1.00)} &{0.000 (1.00)} &{0.001 (1.00)} &{0.002 (1.00)}  \\&
PerSim(p=0.5)& 0.007 (0.99) &0.006 (0.99) &0.000 (1.00) &0.001 (1.00) &0.001 (1.00)\\&
PerSim(p=0.25)& 0.005 (0.99) &0.004 (1.00) &0.000 (1.00) &0.001 (1.00) &0.001 (1.00) \\&
PerSim(p=0.1)& 0.005 (0.99) &0.005 (0.99) &0.000 (1.00) &0.000 (1.00) &0.000 (1.00)  \\
&Vanilla CaDM&  0.378 (0.13) & 0.373 (0.12) & 0.159 (0.13) & 0.329 (0.16) & 0.339 (0.15)  \\
&PE-TS CaDM& 0.399 (0.06) & 0.351 (0.08) & 0.176 (0.07) & 0.193 (0.04) & 0.214 (0.05)  \\ \midrule
\multirow{5}{*}{\rotatebox[origin=c]{90}{{{Pure}} }} &
PerSim(p=1.0)& {0.192 (0.99)} &{0.183 (0.98)} &{0.029 (0.86)} &{0.029 (0.90)} &{0.034 (0.88)}  \\
&PerSim(p=0.5)& 0.039 (0.98) &0.052 (0.99) &0.029 (0.84) &0.030 (0.89) &0.036 (0.86)  \\
&PerSim(p=0.25)& 0.061 (0.98) &0.479 (0.98) &0.029 (0.88) &0.034 (0.87) &0.039 (0.84)  \\
&PerSim(p=0.1)& 0.043 (0.98) &0.046 (0.98) &0.033 (0.83) &0.029 (0.90) &0.034 (0.88) \\
&Vanilla CaDM& 0.213 (0.05) & 0.204 (0.04) & 0.199 (0.04) & 0.357 (0.03) & 0.362 (0.03) \\
&PE-TS CaDM&  0.432 (0.07) & 0.450 (0.09) & 0.187 (0.04) & 0.230 (0.06) & 0.232 (0.07)   \\
\cmidrule[1.5pt]{1-7}
\end{tabular}
\end{table}

\begin{table}[h]
\caption{Average Reward: MountainCar, Unseen Agents}
\label{table:unseen_agents_rew}
\tabcolsep=0.02cm
\fontsize{7.5pt}{9.0pt}\selectfont
\centering
\begin{tabular}{@{}p{0.8cm} p{2.2cm}p{2.2cm}p{2.2cm}p{2.2cm}p{2.2cm}p{2.2cm} @{}}
\toprule
\multirow{2}{*}{Data}     & \multirow{2}{*}{Method}       & {Agent 1}              &  {Agent 2}      & {{Agent 3}} & {Agent 4}               & {Agent 5}      \\     
& & {0.00005 }  & {0.00008}   & {0.002} & {0.0037}   & {0.004}     \\ \toprule
\multirow{6}{*}{\rotatebox[origin=c]{90}{{{Random}} }} &
PerSim(p=1.0)& {-53.82}$\pm$\scriptsize{0.41} &{-53.97}$\pm$\scriptsize{0.35} &{-188.7}$\pm$\scriptsize{6.46} &{-202.6}$\pm$\scriptsize{2.69} &{-200.2}$\pm$\scriptsize{0.74} \\& 
PerSim(p=0.5)& -53.98$\pm$\scriptsize{0.93} &-54.57$\pm$\scriptsize{0.39} &-192.3$\pm$\scriptsize{3.61} &-207.8$\pm$\scriptsize{2.53} &-204.0$\pm$\scriptsize{1.15} \\&
PerSim(p=0.25)& -53.77$\pm$\scriptsize{0.33} &-54.22$\pm$\scriptsize{0.88} &-196.4$\pm$\scriptsize{4.48} &-206.4$\pm$\scriptsize{5.45} &-209.9$\pm$\scriptsize{5.33}\\&
PerSim(p=0.1)& -53.25$\pm$\scriptsize{0.56} &-54.23$\pm$\scriptsize{1.05} &-198.2$\pm$\scriptsize{1.38} &-207.3$\pm$\scriptsize{0.86} &-208.6$\pm$\scriptsize{3.40} \\ 
&Vanilla CaDM & -102.4$\pm$\scriptsize{15.6} & -97.17$\pm$\scriptsize{20.6} &-500.0$\pm$\scriptsize{0.0} &-500.0$\pm$\scriptsize{0.0} &-500.0$\pm$\scriptsize{0.0}  \\
&PE-TS CaDM  & -82.60$\pm$\scriptsize{20.2} & -96.07$\pm$\scriptsize{5.10} & -500.0$\pm$\scriptsize{0.0} & -500.0$\pm$\scriptsize{0.0} &-500.0$\pm$\scriptsize{0.0} \\
&BCQ-P & -500.0$\pm$\scriptsize{0.0} &-500.0$\pm$\scriptsize{0.0} &-500.0$\pm$\scriptsize{0.0} &-500.0$\pm$\scriptsize{0.0} &-500.0$\pm$\scriptsize{0.0} \\\midrule
\multirow{5}{*}{\rotatebox[origin=c]{90}{{{Pure}} }} 
&PerSim(p=1.0)& -75.17$\pm$\scriptsize{12.8} &-74.73$\pm$\scriptsize{11.4} &{-179.3}$\pm$\scriptsize{4.74} &{-203.7}$\pm$\scriptsize{4.13} &{-201.4}$\pm$\scriptsize{5.72}  \\ &
PerSim(p=0.5)& -79.83$\pm$\scriptsize{14.9} &-79.40$\pm$\scriptsize{13.9} &-181.9$\pm$\scriptsize{2.02} &-206.7$\pm$\scriptsize{6.51} &-201.6$\pm$\scriptsize{4.67}  \\ &
PerSim(p=0.25)& -77.28$\pm$\scriptsize{13.4} &-77.73$\pm$\scriptsize{19.7} &-178.4$\pm$\scriptsize{2.77} &-199.7$\pm$\scriptsize{3.41} &-209.1$\pm$\scriptsize{4.83}  \\ &
PerSim(p=0.1)& -85.08$\pm$\scriptsize{22.7} &-90.42$\pm$\scriptsize{19.6} &-183.0$\pm$\scriptsize{1.40} &-204.2$\pm$\scriptsize{6.48} &-203.0$\pm$\scriptsize{5.12}  \\
&Vanilla CaDM  & {-52.63}$\pm$\scriptsize{1.05} & {-54.90}$\pm$\scriptsize{2.60} &-446.8$\pm$\scriptsize{51.2} &-494.1$\pm$\scriptsize{9.30} &-494.1$\pm$\scriptsize{10.3}  \\
&PE-TS CaDM & -60.17$\pm$\scriptsize{2.66} & -63.43$\pm$\scriptsize{4.80} & -374.4$\pm$\scriptsize{77.3} & -500.0$\pm$\scriptsize{0.0} &-500.0$\pm$\scriptsize{0.0}\\
&BCQ-P  & -184.9$\pm$\scriptsize{170} &-183.1$\pm$\scriptsize{170} &-277.6$\pm$\scriptsize{183} &-322.7$\pm$\scriptsize{146} &-333.7$\pm$\scriptsize{139}\\
\cmidrule[1.5pt]{1-7}
\end{tabular}
\end{table}

\subsection{Robustness to Data Scarcity} \label{appendix:num_agents}
\noindent \textbf{Setup.} In this experiment, we address the robustness of PerSim to data scarcity. In particular, we decrease the number of observed agents  from $N=250$ to $N=25$ in all three benchmarking environments. As is done in previous experiments, we compare with the two variants of CaDM and BCQ, and evaluate the performance on five test agents.
We use trajectories generated by a random policy in MountainCar and HalfCheetah, and pure policy for CartPole. 
We use a pure policy for CartPole to ensure that each trajectory is not too short (see Table~\ref{table:datasets_details} for the average length of a trajectory in each environment under different policies).
We perform these experiments five times, where in each time, the agent covariates are re-sampled from the covariates range. We report the average reward across the five trials and the corresponding standard deviations.

\vspace{2mm}
\noindent \textbf{Results.}  We report the average reward achieved by PerSim and baselines in Tables \ref{table:mc_data}, \ref{table:cp_data} and \ref{table:hc_data} for MountainCar, CartPole, and HalfCheetah, respectively.
As demonstrated in the tables, even when we vary the number of trajectories, PerSim consistently achieves a higher reward than the other baselines across all agents in MountainCar and CartPole.
In HalfCheetah, PerSim and PE-TS CaDM perform the best among the baselines. 
%
%
One thing to note is the high variance in HalfCheetah experiments, across all baselines, indicating the fundamental challenge faced when dealing with environments with both high-dimensional state space and limited data.
Addressing such a challenge remains an interesting direction for future work.

\begin{table}[!htb]
\caption{Average Reward: MountainCar with different number of training agents.}
\label{table:mc_data}
\tabcolsep=0.1cm
\fontsize{8.0pt}{11.0pt}\selectfont
\centering
\begin{tabular}{@{}p{0.6cm}p{0.6cm}p{1.95cm}p{1.95cm}p{1.95cm}p{1.95cm}p{1.95cm}p{1.95cm}@{}}
\toprule
\multirow{2}{*}{Data}   & \multirow{2}{*}{N}    & \multirow{2}{*}{Method}       & {Agent 1}              &  {Agent 2}      & {{Agent 3}} & {Agent 4}               & {Agent 5}               \\
{}    &                 &        & {0.0001 }  & {0.0005}   & {0.001} & {0.0025}   & {0.0035}     \\ \toprule
\multirow{4}{*}{\rotatebox[origin=c]{90}{{{Random}} }}& \multirow{4}{*}{\rotatebox[origin=c]{0}{{{250}} }} & PerSim 
& {-53.70}$\pm$\scriptsize{0.41} 
& {-66.50 }$\pm$\scriptsize{1.21} 
& {-116.6}$\pm$\scriptsize{3.18} 
& {{-192.3}}$\pm$\scriptsize{1.23} 
& {-199.6}$\pm$\scriptsize{3.40}  \\
&&Vanilla CaDM& -59.90$\pm$\scriptsize{1.61} & -78.13$\pm$\scriptsize{7.11} & -332.6$\pm$\scriptsize{41.3}  & -467.8$\pm$\scriptsize{16.2} & -500.0$\pm$\scriptsize{0.0} \\
&&PE-TS CaDM& -73.66$\pm$\scriptsize{3.15} & -106.8$\pm$\scriptsize{7.15} & -473.8$\pm$\scriptsize{37.0} & -500.0$\pm$\scriptsize{0.0}  & -500.0$\pm$\scriptsize{0.0} \\
&&BCQ-P& -500.0$\pm$\scriptsize{0.0} & -500.0$\pm$\scriptsize{0.0} & -500.0$\pm$\scriptsize{0.0} & -500.0$\pm$\scriptsize{0.0}  & -500.0$\pm$\scriptsize{0.0}   \\ \midrule
\multirow{4}{*}{\rotatebox[origin=c]{90}{{{Random}} }}& \multirow{4}{*}{\rotatebox[origin=c]{0}{{{100}} }} & PerSim &  {-55.00}$\pm$\scriptsize{0.70} &
{-67.02}$\pm$\scriptsize{1.76} & 
{-110.3}$\pm$\scriptsize{3.46} 
& {-193.1}$\pm$\scriptsize{5.21} & 
{-197.4}$\pm$\scriptsize{3.05} \\
&&Vanilla CaDM& -66.00$\pm$\scriptsize{5.06}  & -83.10$\pm$\scriptsize{8.64} & -307.1$\pm$\scriptsize{96.2} & -486.3$\pm$\scriptsize{23.7}  & -500.0$\pm$\scriptsize{0.0}    \\
&&PE-TS CaDM&     -79.33$\pm$\scriptsize{4.36} & -106.5$\pm$\scriptsize{19.3} & -418.4$\pm$\scriptsize{27.3} & -492.7$\pm$\scriptsize{10.3}  & -499.9$\pm$\scriptsize{0.14} \\
&&BCQ-P& -500.0$\pm$\scriptsize{0.0} & -500.0$\pm$\scriptsize{0.0} & -500.0$\pm$\scriptsize{0.0} & -500.0$\pm$\scriptsize{0.0}  & -500.0$\pm$\scriptsize{0.0}   \\ \midrule
\multirow{4}{*}{\rotatebox[origin=c]{90}{{{Random}} }}& \multirow{4}{*}{\rotatebox[origin=c]{0}{{{50}} }} & PerSim &  {-54.20}$\pm$\scriptsize{0.90} &{ -66.40}$\pm$\scriptsize{0.17} & {-110.2}$\pm$\scriptsize{8.65} & {-188.7}$\pm$\scriptsize{5.25} & {-199.6}$\pm$\scriptsize{3.23}  \\
&&Vanilla CaDM& -58.80$\pm$\scriptsize{1.40}   & {-66.37}$\pm$\scriptsize{0.35} & -131.8$\pm$\scriptsize{17.6} & -497.2$\pm$\scriptsize{4.85}  & -500.0$\pm$\scriptsize{0.0} \\
&&PE-TS CaDM& -67.86$\pm$\scriptsize{7.15} & -79.73$\pm$\scriptsize{6.37} & -290.5$\pm$\scriptsize{76.9} & -458.4$\pm$\scriptsize{26.8}  & -498.5$\pm$\scriptsize{2.12} \\
&&BCQ-P& -214.3$\pm$\scriptsize{202} & -348.6$\pm$\scriptsize{186} & -428.1$\pm$\scriptsize{103} & -500.0$\pm$\scriptsize{0.0}  & -500.0$\pm$\scriptsize{0.0} \\ \midrule
\multirow{4}{*}{\rotatebox[origin=c]{90}{{{Random}} }}& \multirow{4}{*}{\rotatebox[origin=c]{0}{{{25}} }} & PerSim    & {-56.80}$\pm$\scriptsize{1.81} & {-67.50}$\pm$\scriptsize{2.81} & {-139.9}$\pm$\scriptsize{29.5} & {-246.8}$\pm$\scriptsize{39.1} & {-243.8}$\pm$\scriptsize{47.1} \\
&&Vanilla CaDM& -62.33$\pm$\scriptsize{3.34} & -76.80$\pm$\scriptsize{10.9} & -275.7$\pm$\scriptsize{63.9} & -473.0$\pm$\scriptsize{24.1} & -496.5$\pm$\scriptsize{4.90}    \\
&&PE-TS CaDM& -67.26$\pm$\scriptsize{6.95} & -86.96$\pm$\scriptsize{10.9} & -331.0$\pm$\scriptsize{121} & -497.9$\pm$\scriptsize{2.92} & -500.0$\pm$\scriptsize{0.0}   \\
&&BCQ-P& -500.0$\pm$\scriptsize{0.0} & -500.0$\pm$\scriptsize{0.0} & -500.0$\pm$\scriptsize{0.0} & -500.0$\pm$\scriptsize{0.0}  & -500.0$\pm$\scriptsize{0.0}   \\ \midrule
\cmidrule[1.5pt]{1-8}
\end{tabular}
\vspace{-4mm}
\end{table}

\begin{table}[!tbh]
\caption{Average Reward: CartPole with different number of training agents. }
\label{table:cp_data}
\tabcolsep=0.1cm
\fontsize{8.0pt}{11.0pt}\selectfont
\centering
\begin{tabular}{@{}p{0.6cm}p{0.6cm}p{1.95cm}p{1.95cm}p{1.95cm}p{1.95cm}p{1.95cm}p{1.95cm}@{}}
\toprule
\multirow{2}{*}{Data}   & \multirow{2}{*}{N}    & \multirow{2}{*}{Method}       & {Agent 1}              &  {Agent 2}      & {{Agent 3}} & {Agent 4}               & {Agent 5}               \\
{}    &                 &        &     {(2/0.5) } & {(10.0/0.5) } & {(18.0/0.5)} & {(10/0.85)}   & {(10/0.15)}      \\ \toprule
\multirow{4}{*}{\rotatebox[origin=c]{90}{{{Pure}} }}& \multirow{4}{*}{\rotatebox[origin=c]{0}{{{250}} }} & PerSim & {200.0}$\pm$\scriptsize{0.0} & {198.6}$\pm$\scriptsize{1.96} & {197.3}$\pm$\scriptsize{3.82} & {198.3}$\pm$\scriptsize{1.45} & {196.4}$\pm$\scriptsize{5.11}    \\
&&Vanilla & 132.7$\pm$\scriptsize{15.1}  & 192.8$\pm$\scriptsize{1.69} & 191.9$\pm$\scriptsize{2.59}   &  186.4$\pm$\scriptsize{0.99}   &   65.21$\pm$\scriptsize{12.1}  \\
&&PE-TS CaDM&   65.65$\pm$\scriptsize{17.0}  & 200.0$\pm$\scriptsize{0.0} & 199.4$\pm$\scriptsize{0.90}   &  185.8$\pm$\scriptsize{11.8}   &   167.5$\pm$\scriptsize{15.7}  \\
&&BCQ-P&   130.2$\pm$\scriptsize{1.31} &169.6$\pm$\scriptsize{21.6} &173.3$\pm$\scriptsize{7.94} &167.3$\pm$\scriptsize{8.81} &179.6$\pm$\scriptsize{2.71}  \\ \midrule
\multirow{4}{*}{\rotatebox[origin=c]{90}{{{Pure}} }}& \multirow{4}{*}{\rotatebox[origin=c]{0}{{{100}} }} & PerSim   & {200.0}$\pm$\scriptsize{0.0} & {199.6}$\pm$\scriptsize{0.39} & {199.5}$\pm$\scriptsize{0.64} & {197.8}$\pm$\scriptsize{1.67} & {197.7}$\pm$\scriptsize{2.15}  \\
&&Vanilla CaDM&  150.1$\pm$\scriptsize{29.3}  & 187.8$\pm$\scriptsize{2.67} &  180.4$\pm$\scriptsize{8.08}   &  182.1$\pm$\scriptsize{13.5}   &  80.10$\pm$\scriptsize{12.8}   \\
&&PE-TS CaDM&   65.66$\pm$\scriptsize{23.6}  & 200.0$\pm$\scriptsize{0.0} &  199.9$\pm$\scriptsize{0.14}   &  200.0$\pm$\scriptsize{0.0}   &  170.0$\pm$\scriptsize{21.0}   \\
&&BCQ-P&     90.67$\pm$\scriptsize{52.1} &49.18$\pm$\scriptsize{32.0} &119.0$\pm$\scriptsize{78.6} &108.9$\pm$\scriptsize{70.1} &81.30$\pm$\scriptsize{66.6}  \\
\midrule
\multirow{4}{*}{\rotatebox[origin=c]{90}{{{Pure}} }}& \multirow{4}{*}{\rotatebox[origin=c]{0}{{{50}} }} & PerSim  & {200.0}$\pm$\scriptsize{0.0} & {199.4}$\pm$\scriptsize{0.87} & {199.0}$\pm$\scriptsize{1.34} & {191.2}$\pm$\scriptsize{2.87} & {182.7}$\pm$\scriptsize{15.1}  \\
&&Vanilla CaDM&    107.8$\pm$\scriptsize{6.89}  & 194.68$\pm$\scriptsize{3.87} & 184.3$\pm$\scriptsize{2.90}   &  187.3$\pm$\scriptsize{8.70}   &   76.46$\pm$\scriptsize{14.1}  \\
&&PE-TS CaDM&   96.47$\pm$\scriptsize{61.2}  & 200.00$\pm$\scriptsize{0.0} & 196.0$\pm$\scriptsize{4.02}   &  192.6$\pm$\scriptsize{7.40}   &   152.9$\pm$\scriptsize{20.2}  \\
&&BCQ-P&  121.0$\pm$\scriptsize{30.0} &70.96$\pm$\scriptsize{15.8} &122.4$\pm$\scriptsize{18.6} &145.6$\pm$\scriptsize{29.0} &100.1$\pm$\scriptsize{44.2}   \\ \midrule
\multirow{4}{*}{\rotatebox[origin=c]{90}{{{Pure}} }}& \multirow{4}{*}{\rotatebox[origin=c]{0}{{{25}} }} & PerSim   & {200.0}$\pm$\scriptsize{0.0} & {197.3}$\pm$\scriptsize{3.75} & {200.0}$\pm$\scriptsize{0.0} & {200.0}$\pm$\scriptsize{0.05} & {183.5}$\pm$\scriptsize{6.20}    \\
&&Vanilla CaDM&    108.6$\pm$\scriptsize{16.1} & 190.3$\pm$\scriptsize{3.65} & 187.1$\pm$\scriptsize{1.22} & 185.9$\pm$\scriptsize{6.89} & 56.76$\pm$\scriptsize{11.0}   \\
&&PE-TS CaDM&  56.93$\pm$\scriptsize{19.9} & 185.2$\pm$\scriptsize{5.02} & 175.5$\pm$\scriptsize{13.9} & 149.0$\pm$\scriptsize{14.9} & 155.3$\pm$\scriptsize{15.1}   \\
&&BCQ-P&  135.9$\pm$\scriptsize{4.07} &55.27$\pm$\scriptsize{47.4} &154.7$\pm$\scriptsize{9.99} &152.5$\pm$\scriptsize{16.7} &144.9$\pm$\scriptsize{25.3} \\
\cmidrule[1.5pt]{1-8}
\end{tabular}
\end{table}

\begin{table}[!tbh]
\caption{Average Reward: HalfCheetah with different number of training agents.}
\label{table:hc_data}
\tabcolsep=0.1cm
\fontsize{8.0pt}{11.0pt}\selectfont
\centering
\begin{tabular}{@{}p{0.6cm}p{0.6cm}p{1.95cm}p{1.95cm}p{1.95cm}p{1.95cm}p{1.95cm}p{1.95cm}@{}}
\toprule
\multirow{2}{*}{Data}   & \multirow{2}{*}{N}    & \multirow{2}{*}{Method}       & {Agent 1}              &  {Agent 2}      & {{Agent 3}} & {Agent 4}               & {Agent 5}               \\
{}    &                 &        &    {(0.3/1.7)} & {(1.7/0.3)}   & {(0.3/0.3)}  & {(1.7/1.7)}  & {(1.0/1.0)} \\ \toprule
\multirow{4}{*}{\rotatebox[origin=c]{90}{{{Random}} }}& \multirow{4}{*}{\rotatebox[origin=c]{0}{{{250}} }} &
 PerSim 
& ~{1688}$\pm$\scriptsize{1093} 
& ~1415$\pm$\scriptsize{1311} 
& ~703.1$\pm$\scriptsize{531} 
& ~{281.0}$\pm$\scriptsize{309} 
&  ~510.6$\pm$\scriptsize{510}  \\

&&Vanilla CaDM& ~277.6$\pm$\scriptsize{62.6} & ~335.0$\pm$\scriptsize{278} & ~240.4$\pm$\scriptsize{530} & ~278.9$\pm$\scriptsize{98.0} & ~362.2$\pm$\scriptsize{124} \\
&&PE-TS CaDM& ~1006$\pm$\scriptsize{556} & {~1833}$\pm$\scriptsize{666} & {~986.2}$\pm$\scriptsize{211} & {~659.2}$\pm$\scriptsize{76.3} & {~2303}$\pm$\scriptsize{571} \\
&&BCQ-P& -1.780$\pm$\scriptsize{0.36} &-1.410$\pm$\scriptsize{0.19} &-1.830$\pm$\scriptsize{0.21} &-1.880$\pm$\scriptsize{0.25} &-1.830$\pm$\scriptsize{0.26}   \\ \midrule

\multirow{4}{*}{\rotatebox[origin=c]{90}{{{Random}} }}& \multirow{4}{*}{\rotatebox[origin=c]{0}{{{100}} }} 
& PerSim &  ~{2072}$\pm$\scriptsize{284} &
~{1164}$\pm$\scriptsize{102} & 
~{1115}$\pm$\scriptsize{493} 
& ~{903.3}$\pm$\scriptsize{398} & 
~{1058}$\pm$\scriptsize{219} \\
&&Vanilla CaDM&  ~168.9$\pm$\scriptsize{154} & ~131.1$\pm$\scriptsize{110} & ~93.20$\pm$\scriptsize{182} & ~176.5$\pm$\scriptsize{131} & ~415.9$\pm$\scriptsize{151}   \\
&&PE-TS CaDM&   ~803.0$\pm$\scriptsize{335} & ~657.5$\pm$\scriptsize{125} & ~676.0$\pm$\scriptsize{244} & ~586.8$\pm$\scriptsize{43.2} & ~1484$\pm$\scriptsize{934}   \\
&&BCQ-P& -1.630$\pm$\scriptsize{0.08} &-1.430$\pm$\scriptsize{0.08} &-1.770$\pm$\scriptsize{0.12} &-1.830$\pm$\scriptsize{0.10} &-1.860$\pm$\scriptsize{0.10}   \\ \midrule

\multirow{4}{*}{\rotatebox[origin=c]{90}{{{Random}} }}& \multirow{4}{*}{\rotatebox[origin=c]{0}{{{50}} }} 
& PerSim &  ~{821.9}$\pm$\scriptsize{529} &
~{1984}$\pm$\scriptsize{58.2} & 
~{103.0}$\pm$\scriptsize{122} & 
~{66.49}$\pm$\scriptsize{194} & 
~{701.2}$\pm$\scriptsize{112}  \\
&&Vanilla CaDM& ~236.0$\pm$\scriptsize{234} & ~670.0$\pm$\scriptsize{167} & ~68.55$\pm$\scriptsize{220} & ~248.7$\pm$\scriptsize{44.4} & ~802.8$\pm$\scriptsize{355} \\
&& PE-TS CaDM& ~496.4$\pm$\scriptsize{166} & ~1002$\pm$\scriptsize{423} & 
~{119.6}$\pm$\scriptsize{62.0} & 
~{541.4}$\pm$\scriptsize{180} & 
~{1895}$\pm$\scriptsize{989} \\
&&BCQ-P& 
-1.710$\pm$\scriptsize{0.28} &-1.510$\pm$\scriptsize{0.22} &-1.800$\pm$\scriptsize{0.21} &-1.770$\pm$\scriptsize{0.26} &-1.910$\pm$\scriptsize{0.17} \\ \midrule

\multirow{4}{*}{\rotatebox[origin=c]{90}{{{Random}} }}& \multirow{4}{*}{\rotatebox[origin=c]{0}{{{25}} }}
& PerSim    & ~{1110}$\pm$\scriptsize{328} & ~{1125}$\pm$\scriptsize{195} & ~{686.4}$\pm$\scriptsize{352} & ~106.4$\pm$\scriptsize{81.1} &  ~801.2$\pm$\scriptsize{349}\\
&&Vanilla CaDM& ~229.8$\pm$\scriptsize{370} & ~187.5$\pm$\scriptsize{215} & -72.3$\pm$\scriptsize{148} & ~206.0$\pm$\scriptsize{132} & ~877.0$\pm$\scriptsize{449}    \\
&&PE-TS CaDM& ~619.9$\pm$\scriptsize{271} & ~878.6$\pm$\scriptsize{470} & ~84.30$\pm$\scriptsize{221} & ~{291.3}$\pm$\scriptsize{258} & ~{1464}$\pm$\scriptsize{913} \\
&&BCQ-P& -134.8$\pm$\scriptsize{8.40} &-210.5$\pm$\scriptsize{53.4} &-170.0$\pm$\scriptsize{14.9} &-178.5$\pm$\scriptsize{22.9} &-87.75$\pm$\scriptsize{19.0}   \\ \midrule
\cmidrule[1.5pt]{1-8}
\end{tabular}
\end{table}

\end{document}